\documentclass[10pt,journal,compsoc]{IEEEtran}

\ifCLASSOPTIONcompsoc
  \usepackage[nocompress]{cite}
\else
  \usepackage{cite}
\fi

\usepackage{amsmath,amssymb,graphicx}
\usepackage{amsfonts}
\usepackage{url}
\usepackage{algorithm}
\usepackage{algorithmicx}
\usepackage{algpseudocode}
\usepackage{booktabs}
\usepackage{threeparttable}
\usepackage{multirow}
\usepackage{subcaption}
\usepackage{caption}
\usepackage{array}
\usepackage{color}
\usepackage{soul}
\DeclareMathOperator*{\argmax}{argmax}

\graphicspath{{figure/}}

\begin{document}

\title{Advanced Dropout: A Model-free Methodology for Bayesian Dropout Optimization}

\author{Jiyang~Xie,~\IEEEmembership{Student Member,~IEEE,}
        Zhanyu~Ma,~\IEEEmembership{Senior Member,~IEEE,}\\
        Jianjun~Lei,~\IEEEmembership{Senior Member,~IEEE,}
        Guoqiang~Zhang,~\IEEEmembership{Member,~IEEE,}\\
        Jing-Hao~Xue,~\IEEEmembership{Member,~IEEE,}
        Zheng-Hua~Tan,~\IEEEmembership{Senior Member,~IEEE,}
        and~Jun~Guo
\IEEEcompsocitemizethanks{
\vspace{-1mm}
\IEEEcompsocthanksitem J. Xie, Z. Ma, and J. Guo are with the Pattern Recognition and Intelligent Systems Lab., Beijing University of Posts and Telecommunications, China. E-mail: $\{$xiejiyang$2013$, mazhanyu, guojun$\}$@bupt.edu.cn\protect\\
\vspace{-3mm}
\IEEEcompsocthanksitem J. Lei is with the School of Electrical and Information Engineering, Tianjin University, Tianjin, China. E-mail: jjlei@tju.edu.cn\protect\\
\vspace{-3mm}
\IEEEcompsocthanksitem G. Zhang is with the School of Electrical and Data Engineering, University of Technology Sydney, Australia. E-mail: guoqiang.zhang@uts.edu.au\protect\\
\vspace{-3mm}
\IEEEcompsocthanksitem J. -H. Xue is with the Department of Statistical Science, University College London, United Kingdom. E-mail: jinghao.xue@ucl.ac.uk\protect\\
\vspace{-3mm}
\IEEEcompsocthanksitem Z. -H. Tan is with the Department of Electronic Systems, Aalborg University, Denmark. E-mail: zt@es.aau.dk\protect\\
\vspace{-5mm}
}
\thanks{
(Corresponding author: Zhanyu Ma)}}

\markboth{Journal of \LaTeX\ Class Files,~Vol.~xx, No.~x,~20xx}%
{Xie \MakeLowercase{\textit{et al.}}: Bare Demo of IEEEtran.cls for Computer Society Journals}

\IEEEtitleabstractindextext{%
\begin{abstract}

Due to lack of data, overfitting ubiquitously exists in real-world applications of deep neural networks (DNNs). We propose advanced dropout, a model-free methodology, to mitigate overfitting and improve the performance of DNNs. The advanced dropout technique applies a model-free and easily implemented distribution with parametric prior, and adaptively adjusts dropout rate. Specifically, the distribution parameters are optimized by stochastic gradient variational Bayes in order to carry out an end-to-end training. We evaluate the effectiveness of the advanced dropout against nine dropout techniques on {seven} computer vision datasets (five small-scale datasets and {two} large-scale datasets) with various base models. The advanced dropout outperforms all the referred techniques on all the datasets.We further compare the effectiveness ratios and find that advanced dropout achieves the highest one on most cases. Next, we conduct a set of analysis of dropout rate characteristics, including convergence of the adaptive dropout rate, the learned distributions of dropout masks, and a comparison with dropout rate generation without  an explicit distribution. In addition, the ability of overfitting prevention is evaluated and confirmed. Finally, we extend the application of the advanced dropout to uncertainty inference, network pruning, {text classification}, and regression. The proposed advanced dropout is also superior to the corresponding referred methods. Codes are available at~\url{https://github.com/PRIS-CV/AdvancedDropout}.
\end{abstract}

}

\maketitle

\IEEEdisplaynontitleabstractindextext

\IEEEpeerreviewmaketitle

\IEEEraisesectionheading{\section{Introduction}\label{sec:introduction}}

\IEEEPARstart{S}{ignificant} progress has been made in deep learning in recent years, especially for computer vision tasks such as image classification~\cite{simonyan2015very,he2016deep,huang2017densely,li2019dual,li2019large} and image retrieval~\cite{xu2016instance,xu2018sketchmate,xu2018cross,ma2019shoe,bai2021unsupervised}. Given that deep neural networks (DNNs) have millions or even billions of parameters, they require a large amount of data for training. However, limited labeled data can be obtained in many applications~\cite{xu2018webly-supervised,adeli2019semi-supervised,ma2018group,zhu2019image-text,li2019dual}. Thus, overfitting ubiquitously exists in real-world data analysis, which negatively affects the performance of DNNs.

To address this issue, a dropout technique~\cite{hinton12} was proposed to regularize the model parameters by randomly dropping the hidden nodes of DNNs in the training steps to avoid co-adaptations of these nodes. The standard dropout technique and its variants have played an important role in preventing overfitting and popularizing deep learning~\cite{xie2019soft,achille2018information}.

\begin{figure*}[!t]
  \centering
  \includegraphics[width=0.9\textwidth]{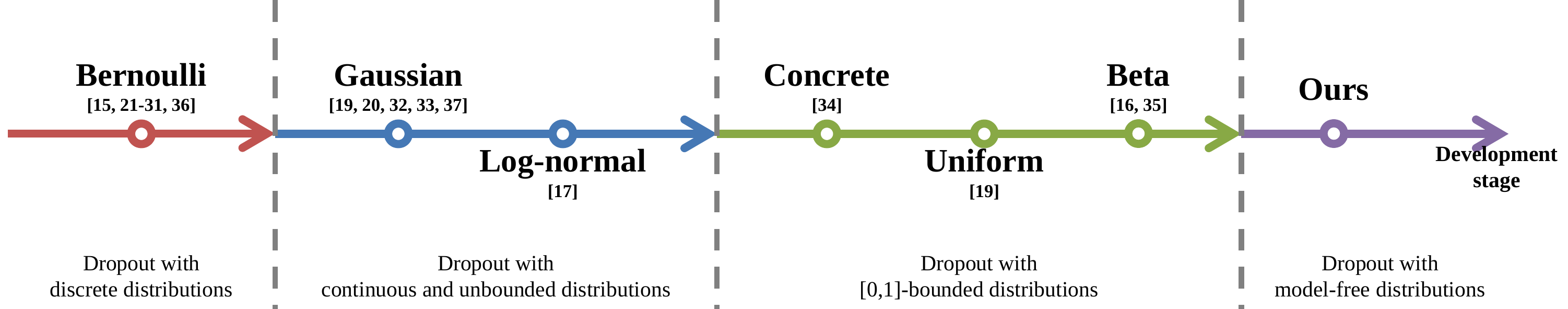}
  \caption{\footnotesize Development of the distributions applied in various dropout techniques.}\label{fig:distribdevelopment}
  \vspace{-6mm}
\end{figure*}

Various dropout techniques have been widely utilized in deep neural network training and inference as surveyed in~\cite{labach2019survey}. Existing works~\cite{hinton12,shen2018continuous,kingma2015variational,wan2013regularization,wang2019jumpout,ba2013adaptive,maeda2015a,gal16mcdropout,gal2016a,khan2019regularization,ko2017controlled,li2017dropout,salehinejad2019ising,chen2019mutual,wang2013fast,Srivastava14,achille2018information,gal17concrete,xie2019soft,liu2019beta,bulo2016dropout,ma2017dropout,gao2019demystifying,li2016improved} differentiate from each other in their use of distributions as shown in Figure~\ref{fig:distribdevelopment}. Most of the works utilized the Bernoulli distribution for their dropout masks to perform the ``dropping'' and ``holding'' in DNNs~\cite{hinton12,wan2013regularization,wang2019jumpout,ba2013adaptive,maeda2015a,gal16mcdropout,gal2016a,khan2019regularization}. In order to regularize model parameters, they randomly drop hidden nodes of DNNs by multiplying Bernoulli distributed masks during training. Meanwhile, some other works applied binary dropout masks, which can be considered as a special Bernoulli distribution in the viewpoint of Bayesian learning~\cite{ko2017controlled,li2017dropout,salehinejad2019ising,chen2019mutual}. Gaussian distribution is another popular choice for modeling the dropout masks~\cite{shen2018continuous,kingma2015variational,wang2013fast,Srivastava14}. They considered the Gaussian distribution as a fast approximation of the Bernoulli distribution~\cite{wang2013fast}, which was applicable in the local reparameterization trick and better than the Bernoulli distribution~\cite{kingma2015variational}. In addition, other distributions including log-normal~\cite{achille2018information}, uniform~\cite{shen2018continuous}, concrete~\cite{gal17concrete}, or beta~\cite{xie2019soft,liu2019beta} distributions, were also used in dropout variants.

One issue in common among all the aforementioned dropout techniques is that they all model the dropout masks via model-specific distributions, mainly with Bernoulli distribution~\cite{hinton12,wan2013regularization,wang2019jumpout,ba2013adaptive,maeda2015a,gal16mcdropout,gal2016a,khan2019regularization,wang2019rademacher}, and also with Gaussian~\cite{shen2018continuous,kingma2015variational,wang2013fast,Srivastava14,liu2019variational},  log-normal~\cite{achille2018information}, uniform~\cite{shen2018continuous}, concrete~\cite{gal17concrete}, or beta~\cite{xie2019soft,liu2019beta} distributions. The assumption of a specific distribution introduces a bias that can limit their ability of modeling the dropout masks, as the specific distribution may undesirably restrict the possibility of changes. Few works exceptionally consider whether the distributions are suitable for modelling dropout or not. Soft dropout~\cite{xie2019soft} and $\beta$-dropout~\cite{liu2019beta} applied the beta distribution which is able to approximate all the other aforementioned distributions, but it is difficult to optimize the parameters of the beta distribution for dropout masks. One solution to this problem is Bayesian approximation using tractable distributions~\cite{xie2019soft} and manual selection of beta distribution parameters~\cite{liu2019beta}, which both have limitations for dropout training.

In this paper, we propose a model-free methodology for dropout, named advanced dropout, to further improve the capability of overfitting prevention and boost the performance of DNNs. The dropout technique applies a model-free and easily implemented distribution with a parametric prior to adaptively adjust the dropout rate. Furthermore, the prior parameters are optimized by the stochastic gradient variational Bayes (SGVB) inference~\cite{kingma2014auto-encoding} to perform an end-to-end training procedure of DNNs. Our major contributions can be summarized as follows:

\begin{itemize}
  \item We introduce a novel~\emph{model-free} methodology that is more flexible than the existing techniques with specifically explicit distributions in dropouts. By choosing proper parameters, the proposed methodology is able to replace all the aforementioned distributions to generalize the other dropout variants under the model-free framework (Section~\ref{ssec:distrib}). 
  
  \item We apply a parametric prior form for the model-free distribution's parameters to~\emph{adaptively adjust the dropout rate} based on input features. The prior can be integrated into the end-to-end training such that the calculation is facilitated (Section~\ref{ssec:prior}). 
  
  \item We propose~\emph{advanced dropout} consisting of the two key components,~\emph{i.e.}, a model-free distribution for dropout masks and a parametric prior for the parameters of the distribution. We evaluate the effectiveness of the advanced dropout with image classification task. Experimental results demonstrate that the advanced dropout outperforms all the nine recently proposed techniques on {seven} widely used datasets with various base models. We further compare training time and effectiveness ratios and find that the advanced dropout achieves highest effectiveness ratios on most of the datasets (Section~\ref{ssec:classification}).
  
  \item We extensively study the advanced dropout technique on the following aspects: the effectiveness of each component, dropout rate characteristics, and the ability of overfitting prevention. We clarify that the key components are essential for the technique (Section~\ref{ssec:ablation}). Meanwhile, the analysis of dropout rate characteristics demonstrates that the technique is able to achieve a~\emph{stable convergence} of dropout rate (Section~\ref{sssec:convergence}), illustrates the learned distributions of dropout masks (Section~\ref{sssec:learneddistribution}), and shows its better performance than the one using dropout rate generation without an explicit distribution (Section~\ref{sssec:compdrwoexpldis}). Finally,~\emph{superior ability of preventing overfitting} of the advanced dropout technique is shown (Section~\ref{ssec:overfitting}).
  
  \item We extend the application of the advanced dropout technique to~\emph{uncertainty inference},~\emph{network pruning},{~\emph{text classification}}, and~\emph{regression}. We conduct several experiments on the model uncertainty inference and show the improvement of the advanced dropout technique (Section~\ref{ssec:uncertainty}). We also employ the advanced dropout technique as a network pruning technique and compare it with the state-of-the-art methods to show the performance improvement (Section~\ref{ssec:pruing}). In addition, we apply the advanced dropout technique on text classification (Section~\ref{ssec:textcls}) and regression (Section~\ref{ssec:regression}), and find its superiority among all the referred dropout techniques.

\end{itemize}

\vspace{-6mm}
\section{Related Work}\label{sec:relatedworks}

After Hinton~\emph{et al}.~\cite{hinton12} introduced the standard dropout in $2012$, many variants of dropout have been proposed in recent years, as the significant effectiveness of dropout on preventing overfitting has been discovered when applied on deep and wide DNN structures. Distinct distributions were applied by the dropout variants to their own design strategies of DNN regularization. In particular, six distributions were introduced, including Bernoulli~\cite{hinton12,wan2013regularization,wang2019jumpout,ba2013adaptive,maeda2015a,gal16mcdropout,gal2016a,khan2019regularization,wang2019rademacher}, Gaussian~\cite{shen2018continuous,kingma2015variational,wang2013fast,Srivastava14,liu2019variational}, log-normal~\cite{achille2018information}, uniform~\cite{shen2018continuous}, concrete~\cite{gal17concrete}, and beta~\cite{xie2019soft,liu2019beta} distributions. As shown in Figure~\ref{fig:distribdevelopment}, the dropout techniques can be divided into four development stages with their own kinds of distributions. The following parts of this section will review the works with different distributions in the stages, respectively.

\vspace{-4mm}
\subsection{On Discrete Distributions}\label{ssec:worksdiscrete}

In the research of dropouts, the discrete distributions commonly refer to the Bernoulli distribution and the binary distribution, while the latter can be considered as a special Bernoulli distribution in the viewpoint of Bayesian learning.

Most of works utilized the Bernoulli distribution for their dropout masks to perform ``dropping'' and ``holding'' in DNNs. This standard dropout aims to regularize the model parameters and reduce overfitting by randomly dropping hidden nodes of an fully connected (FC) neural network with Bernoulli distributed masks during training~\cite{hinton12}. DropConnect~\cite{wan2013regularization} randomly selected a subset of weights within the network to zero, rather than activations in each layer. Meanwhile, standout~\cite{ba2013adaptive} performed as a binary belief network, was trained jointly with the DNN using stochastic gradient descent (SGD), and computed the local expectations of binary dropout variables. Maeda~\cite{maeda2015a} introduced a Bayesian interpretation to optimize the dropout rate, which was beneficial for model training and prediction. Gal and Ghahramani~\cite{gal16mcdropout} utilized standard dropout to predict model uncertainty in DNNs in regression, classification, and reinforcement learning. All the aforementioned techniques rely on standard dropout with fixed parameters which are empirically set. Later, a dropout variant has been proposed for RNNs focusing on time dependence representation and demonstrated outstanding effectiveness~\cite{gal2016a}. Spectral dropout~\cite{khan2019regularization} instead implemented standard dropout on the spectrum dimension of the convolutional feature maps, preventing overfitting by eliminating the weak Fourier domain coefficients of activations. Jumpout~\cite{wang2019jumpout} sampled the dropout rate from a monotone decreasing distribution and adaptively normalized the rate at each layer to keep the effective rate. Wang~\emph{et al}.~\cite{wang2019rademacher} proposed a lightweight complexity algorithm called Rademacher Dropout (RadDropout) to achieve adaptive adjustment of dropout rates.

Furthermore, some works applied the binary distribution for the dropout masks, and selected or dropped nodes according to some specific rules. Ko~\emph{et al}.~\cite{ko2017controlled} introduced controlled dropout and intentionally chose the activations to drop non-randomly for improving the training speed and the memory efficiency. Alpha-divergence dropout~\cite{li2017dropout} applied the alpha divergence as a regularization term, replacing the conventional Kullback-Leibler (KL) divergence in approximate variational inference for dropout training. Ising-dropout~\cite{salehinejad2019ising} dropped the activations in a DNN using Ising energy of the network to eliminate the optimization of unnecessary parameters during training. In addition, Chen~\emph{et al}.~\cite{chen2019mutual} proposed mutual information-based dropout (DropMI), introducing mutual information dynamic analysis to the model and highlighting the important activations that are beneficial to the feature representation.

However, one issue with the discrete distributions is the difficulty of dropout rate optimization combined in DNN training, which does not exist with continuous distributions. Except the fixed dropout rate in~\cite{hinton12,wan2013regularization,gal16mcdropout,khan2019regularization}, some works addressed the issue by updating asynchronously~\cite{ba2013adaptive}, adding regularization terms into the loss functions~\cite{maeda2015a}, and randomly selecting from other distributions~\cite{wang2019jumpout}.

\vspace{-1mm}
\subsection{On Continuous and Unbounded Distributions}\label{ssec:worksconandunbound}

The works in~\cite{shen2018continuous,kingma2015variational,wang2013fast,Srivastava14,achille2018information} introduced the continuous and unbounded distributions to dropouts, mainly including the Gaussian and the log-normal distributions. They took advantages of the continuity and the differentiability of the distributions for end-to-end training of DNNs.

The Gaussian distribution is another popular choice for the dropout masks. It is considered a fast approximation of the Bernoulli distribution~\cite{wang2013fast} and is applicable in the local reparameterization trick, better than the Bernoulli distribution~\cite{kingma2015variational}. Wang and Manning~\cite{wang2013fast} proposed a Gaussian approximation of standard dropout under the variational Bayes framework, called fast dropout, with virtually identical regularization performance but much faster convergence. Another extension of dropout in~\cite{Srivastava14} applied Gaussian multiplicative noise with unit mean, replacing the Bernoulli noise. It can be interpreted as a variational method given a particular prior over the network weights to some extent~\cite{Srivastava14}. Kingma et.al.~\cite{kingma2015variational} introduced variational dropout where the dropout rate was optimized by the stochastic gradient variational Bayes (SGVB) inference~\cite{kingma2014auto-encoding}. Continuous dropout~\cite{shen2018continuous} replaced the Bernoulli distribution by a Gaussian distribution with mean $0.5$ or a uniform one as the prior of continuous masks in practice, even though the variance of the Gaussian distribution could not be optimized during training. Liu~\emph{et al}.~\cite{liu2019variational} proposed a variational Bayesian dropout with a hierarchical prior.

In addition, information dropout~\cite{achille2018information} improved dropout by information theory principles and adapted the dropout rate to the data automatically under the Bayesian theory. It applied the log-normal distribution, which is also an unbounded-domain distribution as the Gaussian distribution, and obtained positive dropout masks.

However, masks with large values approaching infinite can be sampled and negatively affect gradient backpropagation, resulting in gradient exploding, which is a huge problem in practice.

\vspace{-4mm}
\subsection{On $\boldsymbol{[0,1]}$-bounded Distributions}\label{ssec:worksbound}

In addition, other distributions with $[0,1]$ bound were also used in the dropout variants. The $[0,1]$-bounded distributions contain the concrete, the uniform, and the beta distributions, which are also continuous distributions.

Gal~\emph{et al}.~\cite{gal17concrete} proposed concrete dropout by introducing a concrete distribution for the dropout masks. The concrete distribution is the first $[0,1]$-bounded distribution applied to dropout and obtained remarkable improvement on overfitting prevention. Furthermore, the uniform distribution was also applied by the continuous dropout and compared with the Gaussian distribution with different variance settings~\cite{shen2018continuous}. The uniform distribution usually performed worse than the Gaussian distribution according to the experimental results in~\cite{shen2018continuous}, due to the low degree of freedom.

Given that beta distribution with different parameters can approximate the Bernoulli, the Gaussian, and the uniform distributions to some extent~\cite{liu2019beta}, it should be great potential than other distributions in dropout regularization. The $\beta$-dropout technique performed best over others and conducted finer control of its regularization; however, the parameters of the beta distributed masks have to be manually selected~\cite{liu2019beta}. In soft dropout~\cite{xie2019soft}, the soft dropout masks were expressed by beta distributed variables which have better continuity than the Bernoulli distribution and higher flexibility in shape than the Gaussian distribution. However, the beta distribution is unfeasible to be directly extended to the SGVB inference which is one of the most effective solutions combining variational inference with the SGD optimization~\cite{xie2019soft}. To utilize soft dropout in the SGVB inference, the beta distribution was approximated by half-Gaussian and half-Laplace distributions, respectively, and the technique could adaptively adjust its dropout rate in the training process~\cite{xie2019soft}. However, the half-Gaussian and half-Laplace approximations are complex and can only approximate part of shapes (U shape) well, but hardly represent other shapes (\emph{e.g.}, uniform shape). Furthermore, the parameters of these two approximations are optimized without following any prior distributions, which affects the performance of overfitting prevention.

\begin{figure}[!t]
  \centering
  \includegraphics[width=0.85\linewidth]{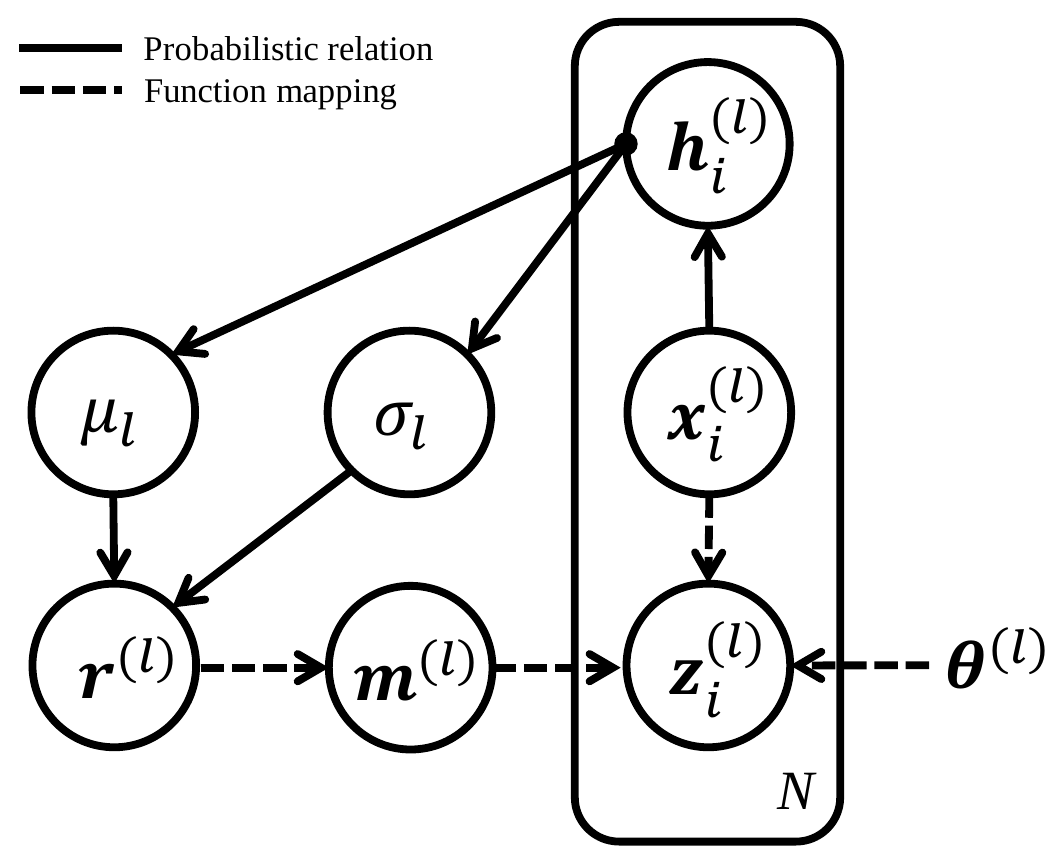}
  \caption{\footnotesize Probabilistic graphical model of the advanced dropout technique. Two key components in the architecture are $1$) the model-free distribution $g(\boldsymbol{m}^{(l)}|\mu_l,\sigma_l)$ and $2$) the parametric prior $p(\mu_l,\sigma_l|\boldsymbol{h}_1^{(l)},\cdots,\boldsymbol{h}_N^{(l)})\prod_{i=1}^N p(\boldsymbol{h}_i^{(l)}|\boldsymbol{x}_i^{(l)})$.}\label{fig:graphicmodel}
  \vspace{-5mm}
\end{figure}

In this context, we consider to design a more generalized and model-free distribution followed by the dropout masks, which is able to approximate the widely used Bernoulli, Gaussian, uniform, concrete, and beta distributions to generalize the fundamental principles of all the dropout variants and has an easy implementation in the SGVB inference for optimization. Although all the aforementioned techniques introduced exact distributions to explicitly explain the dropout mask, we utilize the model-free distributions and overcome the difficulties of them (\emph{i.e.}, shape limitations and distribution parameter optimization) discussed above. In addition, a suitable prior is essential in Bayesian learning. We introduce a parametric prior which is supported by features and integrate information to better optimize the parameters of the model-free distribution.

\vspace{-3mm}
\section{Preliminaries}\label{sec:dropout}

We start with an $L$-layer DNN, which has $K_l, l=1,\cdots,L,$ features in the $l^{th}$ layer. The parameter set of the DNN is defined as $\boldsymbol{\Theta}=\{\boldsymbol{\theta}^{(l)}\}_{l=1}^L$, where $\boldsymbol{\theta}^{(l)}\in R^{K_l\times K_{l-1}}$ is the network parameter matrix of the $l^{th}$ layer. For each layer, we define a layer model with standard dropout~\cite{hinton12} as

\begin{footnotesize}
\begin{equation}\label{eq:dropoutlayer}
  \boldsymbol{z}^{(l)}=a\left(\boldsymbol{m}^{(l)}\odot(\boldsymbol{\theta}^{(l)}\boldsymbol{x}^{(l)})\right),
\end{equation}
\vspace{-4mm}
\end{footnotesize}

\noindent where $\boldsymbol{x}^{(l)}$ and $\boldsymbol{z}^{(l)}$, with $\boldsymbol{x}^{(l+1)}=\boldsymbol{z}^{(l)}$ for $1\le l<L$, are the input and the output of the layer, respectively. $a(\cdot)$ is the activation function,~\emph{e.g.}, rectified linear unit (ReLU) function, $\odot$ is the Hadamard product operation, and $\boldsymbol{m}^{(l)}$ is the dropout mask vector.

In the standard dropout framework~\cite{hinton12}, the elements in $\boldsymbol{m}^{(l)}$ are random variables following~\emph{i.i.d.} Bernoulli distributions with the dropout rate $\rho_l=1-\text{E}[m_j^{(l)}]=1-p_l, j=1,\cdots,K_l$, as

\begin{footnotesize}
\vspace{-4mm}
\begin{equation}\label{eq:bernoulli}
  \boldsymbol{m}^{(l)}\sim\prod_{j=1}^{K_l}\text{Bernoulli}_j(p_l)=\prod_{j=1}^{K_l} p_l^{m_j^{(l)}}(1-p_l)^{1-m_j^{(l)}},
\end{equation}
\vspace{-4mm}
\end{footnotesize}

\noindent where $p_l$ is the parameter of the Bernoulli distribution. In the training step, the dropout mask vector $\boldsymbol{m}^{(l)}$ is sampled from its distribution, producing a binary mask vector. Meanwhile, $\boldsymbol{x}^{(l)}$ is scaled by the mean vector $[p_l]_{K_l}$ in the test step~\cite{hinton12}.

In order to better express the continuity of the dropout masks, the Bernoulli distribution is replaced by different distributions including Gaussian~\cite{Srivastava14}, log-normal~\cite{achille2018information}, uniform~\cite{shen2018continuous}, concrete~\cite{gal17concrete}, and beta~\cite{xie2019soft} distributions. Considering the recent soft dropout~\cite{xie2019soft} as an example, the prior distribution of $\boldsymbol{m}^{(l)}$ is modified into a multi-dimensional beta distribution, which can be considered as a product of~\emph{i.i.d.} beta distributions with the dropout rate $\rho_l=1-\frac{\alpha_l}{\alpha_l+\beta_l}$ as

\begin{footnotesize}
\vspace{-4mm}
\begin{align}\label{eq:betadistrib}
  \boldsymbol{m}^{(l)}&\sim\prod_{j=1}^{K_l}\text{Beta}_j(\alpha_l,\beta_l)\nonumber\\
  &=\prod_{j=1}^{K_l}\frac{(m_j^{(l)})^{\alpha_l-1}(1-m_j^{(l)})^{\beta_l-1}}{\int_0^1 u^{\alpha_l-1}(1-u)^{\beta_l-1}\,du},
\end{align}
\vspace{-3mm}
\end{footnotesize}

\noindent where $\alpha_l,\beta_l>0$ are the shape parameters. Similar to standard dropout,  $\boldsymbol{x}^{(l)}$ is scaled by the mean vector $[\frac{\alpha_l}{\alpha_l+\beta_l}]_{K_l}$ in the test step.

By replacing the binary dropout masks with the beta distributed variables, the soft dropout masks are continuously distributed in the interval $[0,1]$,  rather than only zero or one. Thus, it samples the masks from an infinite space for parameter selection, giving infinite states of each soft dropout mask. The binary mask optimization space of the standard dropout~\cite{hinton12} can be considered as a subset of the soft one.

\vspace{-4mm}
\section{Advanced Dropout}\label{sec:asdp}

In this section, an advanced dropout technique is proposed for DNNs. The probabilistic graphical model is shown in Figure~\ref{fig:graphicmodel}, including two key components in the architecture,~\emph{i.e.}, the model-free distribution and the parametric prior. After introducing the two parts in Section~\ref{ssec:distrib} and Section~\ref{ssec:prior}, respectively, we discuss how to optimize the whole advanced dropout technique by the SGVB inference~\cite{kingma2014auto-encoding} in DNN training within the SGD algorithm in Section~\ref{ssec:sgvb}.

\vspace{-4mm}
\subsection{Model-free Distribution for Advanced Dropout}\label{ssec:distrib}

The development of the distributions applied in various dropout techniques is illustrated in Figure~\ref{fig:distribdevelopment}. The dropout masks firstly followed the Bernoulli distribution, a discrete distribution, to perform the fully ``dropping'' and ``holding'' in DNNs~\cite{hinton12,wan2013regularization,wang2019jumpout,ba2013adaptive,maeda2015a,gal16mcdropout,gal2016a,khan2019regularization,wang2019rademacher}. Then, the distribution was replaced by the Gaussian~\cite{shen2018continuous,kingma2015variational,wang2013fast,Srivastava14,liu2019variational} or the log-normal~\cite{achille2018information} distributions, which are continuous and unbounded distributions. These distributions, as the approximations of the Bernoulli distribution, performed well. However, masks with large values approaching infinity can be sampled to some extent, leading to gradient exploding, which can be a huge problem in practice.

To address this problem, $[0,1]$-bounded distributions including the concrete~\cite{gal17concrete}, the uniform~\cite{shen2018continuous}, and the beta~\cite{xie2019soft,liu2019beta} distributions were introduced and achieved better performance. In addition, the asymmetry of a distribution can introduce more flexible shapes of the probability density function (PDF) for the distribution and adapt different dropout rates in DNN training, which are beneficial to the dropout techniques. As we know that a symmetric distribution can merely present dropout with its dropout rate at $0.5$ (\emph{e.g.}, the Gaussian distribution with mean $0.5$ and the uniform distribution in~\cite{shen2018continuous}), an asymmetric one can express not only the half-dropping case, but also all the other cases of the dropout rate, implemented by different values of the parameters.

\begin{figure}[!t]
  \centering
  \includegraphics[width=0.85\linewidth]{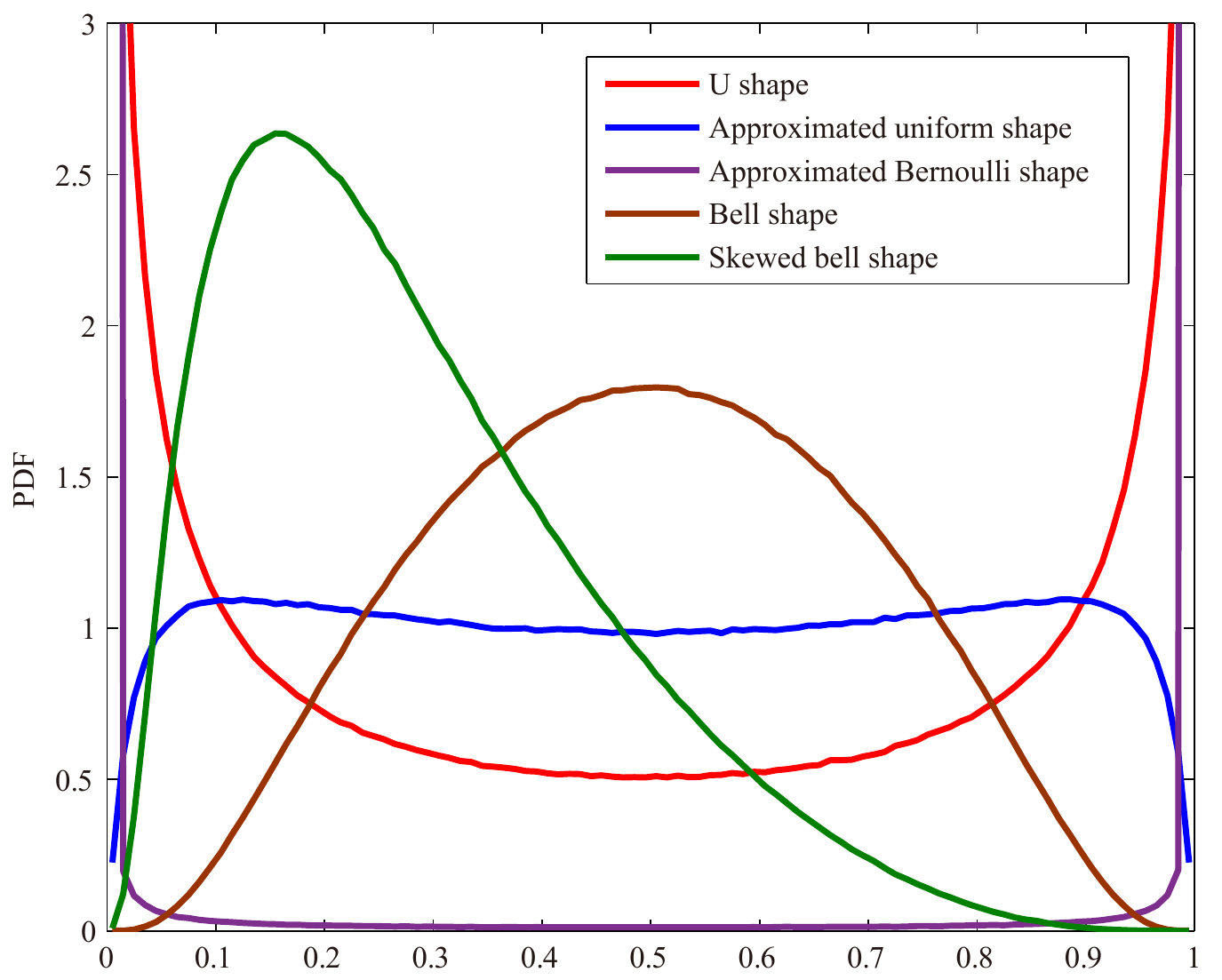}
  \vspace{-3mm}
  \caption{\footnotesize PDFs of the model-free distribution $g$ applied in experiments with different parameter settings. The parameter settings of $g$ with different shapes are $\mu=0, \sigma=3$ for the ``U'' shape, $\mu=0, \sigma=1.6$ for the approximated uniform shape, $\mu=0,\sigma=150$ for the approximated Bernoulli shape, $\mu=0,\sigma=1$ for the bell shape, and $\mu=-1,\sigma=1$ for the skewed bell shape. Note that we can approximate the Gaussian distribution by the bell-shaped $g$ and the log-normal distribution by the skewed bell-shaped $g$, and the beta distribution by the U-, bell-, skewed bell-shaped $g$, respectively.}\label{fig:gpdf}
  \vspace{-5mm}
\end{figure}

In addition, shape is significant for the dropout distribution, because different shapes (\emph{e.g.}, ``U'' and bell shapes) can represent distinct states of the masks, mediately reflecting the state of the model. The beta distribution has different shapes of PDF, including the ``U'', the bell, the skewed bell, and the uniform shapes. However, the concrete distribution has only the ``U'' shape in practice. Thus, the beta distribution is preferred to the other distributions mentioned above. Although all the aforementioned dropout techniques have their own distributions,~\emph{do these distributions suit the needs of dropout itself?}

To further address the issue, we introduce a model-free distribution which satisfies~\emph{all} the properties of the aforementioned distributions and~\emph{all} the requirements of soft dropout. The model-free distribution will not be restricted by any distribution forms and can prevent soft dropout from the infeasibility incurred by some distributions, such as the beta distribution, in the end-to-end training~\cite{xie2019soft}. It can be defined in arbitrary forms and we are able to produce the various shapes by adjusting the parameters.

Here, we propose a model-free distribution, which is denoted by $g$. For the $j^{th}$ element of the dropout mask $m_j^{(l)}\sim g(m_j^{(l)})$ in the $l^{th}$ layer of the DNN, another hidden variable $r_j^{(l)}$ is introduced as the seed variable, following a seed distribution $s(\cdot)$. Then, by introducing a monotonic and differentiable function $k(\cdot)$ as the mapping function, which has its value space in the interval of $[0,1]$, we obtain the $[0,1]$-bounded continuous variable $m_j^{(l)}=k(r_j^{(l)})$ directly. In this case, the seed distribution $s(\cdot)$ and the mapping function $k(\cdot)$ should satisfy two conditions:
\begin{enumerate}
  \item $s(\cdot)$ can be transformed into a differentiable function of its parameters and standard distribution, and should be easy to be sampled. For example, Gaussian, Laplace, exponential, or uniform distributions.
  \item $k(\cdot)$ should be monotonic and differentiable in its domain, and the domain of output values of $k(\cdot)$ should be in the interval of $[0,1]$. For example, Sigmoid function and some piecewise differentiable functions.
\end{enumerate}

\noindent In principle,~\emph{any} seed distribution and mapping function pair can be applied to construct the model-free distribution, as long as they satisfy the conditions $1$) and $2$), respectively. The model-free distribution $g$ can be~\emph{any} form satisfying $\int g(x)d\,x=1$. The relationship between $s(\cdot)$, $k(\cdot)$, and $g$ can be obtained via the formula for calculating the distribution of functions of random variables as

\begin{footnotesize}
\vspace{-4mm}
\begin{equation}\label{eq:goriginalpdf}
  g(m_j^{(l)})=s(k^{-1}(m_j^{(l)}))\frac{d\,k^{-1}(m_j^{(l)})}{d\,m_j^{(l)}},
\end{equation}
\vspace{-3mm}
\end{footnotesize}

\noindent where $k^{-1}(\cdot)$ is the inverse function of $k(\cdot)$. Note that $k(\cdot)$ is a monotonic function, so that $k^{-1}(\cdot)$ can be always obtained.

As we know in DNNs, the Gaussian distribution is a popular assumption and the Sigmoid function is widely used, which satisfy the conditions of $s(\cdot)$ and $k(\cdot)$, respectively. We involve them into the model-free distribution framework. The linear additivity of the Gaussian distribution and differentiability of the Sigmoid function make the model-free distribution $g$ feasible for the SGVB inference. In addition, $k(\cdot)$ allows $m_j^{(l)}$ falling in the bounded interval $[0,1]$. In this case, the variable $m_j^{(l)}$ can be explicitly defined by $r_j^{(l)}$ as

\begin{footnotesize}
\vspace{-4mm}
\begin{align}\label{eq:r}
  m_j^{(l)}&=k(r_j^{(l)})=\text{Sigmoid}(r_j^{(l)})=\frac{1}{1+e^{-r_j^{(l)}}},\\
  r_j^{(l)}&\sim\mathcal{N}(\mu_l,\sigma_l^2),\nonumber
\end{align}
\vspace{-4mm}
\end{footnotesize}

\noindent where $\mu_l$ and $\sigma_l$ are the mean and the standard deviation of the seed variable $r_j^{(l)}$. Furthermore, the PDF of $g$ in this case can be defined as

\begin{footnotesize}
\vspace{-4mm}
\begin{align}\label{eq:gpdf}
  g(m_j^{(l)})&=\mathcal{N}(k^{-1}(m_j^{(l)});\mu_l,\sigma_l^2)\frac{d\,k^{-1}(m_j^{(l)})}{d\,m_j^{(l)}}\nonumber\\
  &=\frac{1}{\sqrt{2\pi}\sigma_l}e^{-\frac{\left(\ln m_j^{(l)}-\ln(1-m_j^{(l)})-\mu_l\right)^2}{2\sigma_l^2}}\left(\frac{1}{m_j^{(l)}(1-m_j^{(l)})}\right),
\end{align}
\vspace{-4mm}
\end{footnotesize}

\noindent where $m_j^{(l)}$ is subject to the interval of $[0,1]$ by $k(\cdot)$. Here, although any seed distribution and mapping function pairs satisfying the conditions $1$) and $2$) can be introduced into the model-free distribution, we use the Gaussian distribution as the seed distribution for the purpose of easy implementation. It is worth to mention that the PDF of $g$ in~\eqref{eq:gpdf} has the same form with that of logit-normal distribution. The logit-normal distribution can be considered as a special case of the model-free distribution. We introduce the logit-normal distributed variables generated by Gaussian variables through the Sigmoid function to exhibit the effectiveness of the proposed methodology in experiments merely.

The expectation of the variable $m_j^{(l)}$ is then calculated as

\begin{footnotesize}
\vspace{-4mm}
\begin{align}\label{eq:rmean}
  \text{E}\left[m_j^{(l)}\right]&=\int_0^1 m_j^{(l)} g(m_j^{(l)})d\,m_j^{(l)}\nonumber\\
  &=\int_{-\infty}^{+\infty}\text{Sigmoid}(r_j^{(l)})\mathcal{N}(r_j^{(l)};\mu_l,\sigma_l^2)d\,r_j^{(l)}\nonumber\\
  &\approx\int_{-\infty}^{+\infty}\phi(\sqrt{\frac{\pi}{8}}r_j^{(l)})\mathcal{N}(r_j^{(l)};\mu_l,\sigma_l^2)d\,r_j^{(l)}\nonumber\\
  &=\phi\left(\frac{\mu_l}{\sqrt{\frac{8}{\pi}+\sigma_l^2}}\right)\approx\text{Sigmoid}\left(\frac{\mu_l}{\sqrt{1+\frac{\pi}{8}\sigma_l^2}}\right),
\end{align}
\vspace{-2mm}
\end{footnotesize}

\noindent where

\begin{footnotesize}
\vspace{-2mm}
\begin{equation}\label{eq:phi}
  \phi(x)=\int_{-\infty}^{x}\mathcal{N}(t;0,1)dt \approx\text{Sigmoid}\left(\sqrt{\frac{8}{\pi}}x\right)
\end{equation}
\vspace{-4mm}
\end{footnotesize}

\noindent is the cumulative distribution function of a standard normal distribution~\cite{bishop06}.

\begin{figure}[!t]
  \centering
  \includegraphics[width=0.85\linewidth]{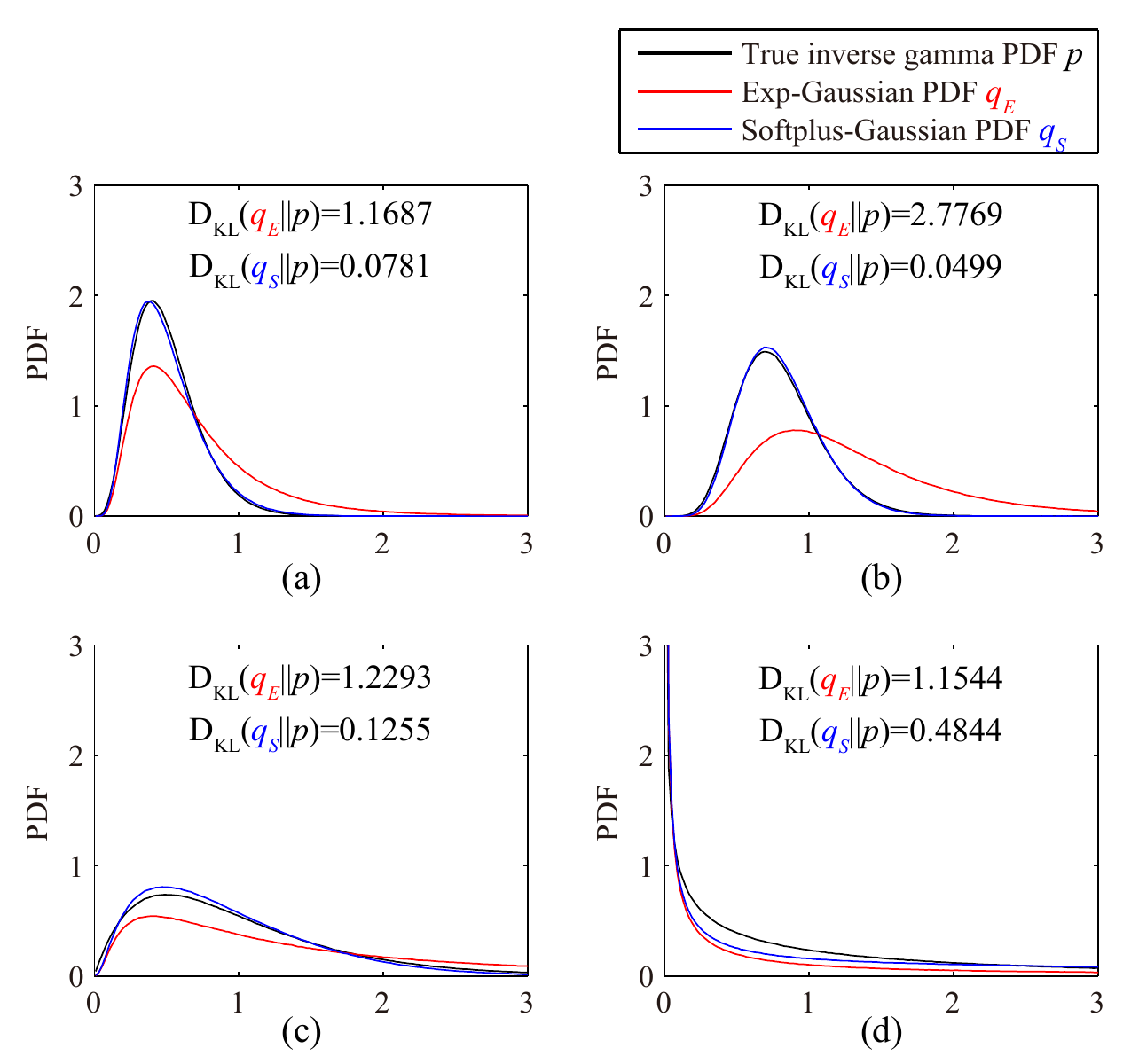}
  \vspace{-4mm}
  \caption{\footnotesize PDFs of the softplus-Gaussian distribution (in blue) and log-normal distribution~\cite{kingma2014auto-encoding} (in red) approximating the inverse gamma distribution (in black) with different parameters:
  (a) $k=5,\theta=0.1$;
  (b) $k=8,\theta=0.1$;
  (c) $k=2,\theta=0.5$;
  (d) $k=0.5,\theta=3$.}\label{fig:softplusapprox}
  \vspace{-6mm}
\end{figure}

Finally, we discuss the distinct shape advantages of the model-free distribution utilized in this paper. As shown in Figure~\ref{fig:gpdf}, the model-free distribution $g$ (with the Gaussian seed distribution and the Sigmoid function) has various shapes corresponding to different parameter settings, including ``U'', bell, skewed bell, and approximated uniform and Bernoulli shapes. It can approximate the Bernoulli, the uniform, the Gaussian, the log-normal, and the beta distributions by adjusting $\mu_l$ and $\sigma_l$. Meanwhile, the concrete distribution is a special case in the form of the model-free PDF $g$. Therefore, the model-free distribution can replace all the aforementioned distributions applied in dropout variants.

\vspace{-3mm}
\subsection{Parametric Prior Distribution for Parameters of Model-free Distribution}\label{ssec:prior}

After defining the model-free distribution form of the dropout masks $\boldsymbol{m}^{(l)}$ in Figure~\ref{fig:graphicmodel}, a prior distribution of the distribution parameters $\mu_l$ and $\sigma_l$ is required in the Bayesian inference. We know that the prior of $\mu_l$ and $\sigma_l$ given the input features $\boldsymbol{x}^{(l)}$ of the $l^{th}$ layer can integrate information and extract features for helping the dropout technique learning a better distribution~\cite{ba2013adaptive,achille2018information}, which means that we can adaptively adjust the dropout rate by optimizing distribution parameters with the prior. In this case, we introduce the prior distribution $p(\mu_l,\sigma_l|\boldsymbol{x}^{(l)})$ as

\begin{footnotesize}
\vspace{-4mm}
\begin{equation}\label{eq:prior}
  p(\mu_l,\sigma_l|\boldsymbol{x}^{(l)})=\int p_{\mu}(\mu_l|\boldsymbol{h}^{(l)})p_{\sigma}(\sigma_l|\boldsymbol{h}^{(l)}) p_{h}(\boldsymbol{h}^{(l)}|\boldsymbol{x}^{(l)}) d\,\boldsymbol{h}^{(l)},
\end{equation}
\vspace{-4mm}
\end{footnotesize}

\noindent where $\boldsymbol{h}^{(l)}$ is the multivariate Gaussian distributed hidden states of the $l^{th}$ layer.

To simplify Bayesian optimization later on, we assume $\mu_l$ and $\sigma_l$ follow Gaussian and inverse gamma distributions, respectively. That is, we set their prior distributions as

\begin{footnotesize}
\vspace{-4mm}
\begin{align}\label{eq:musigmaprior}
  \mu_l&\sim\mathcal{N}(\mu_l|\cdot)\approx p_{\mu}(\mu_l|\boldsymbol{h}^{(l)}),\\
  \sigma_l&\sim\mathcal{IG}(\sigma_l|\cdot)\approx p_{\sigma}(\sigma_l|\boldsymbol{h}^{(l)}),\\
  \boldsymbol{h}^{(l)}&\sim\mathcal{N}(\boldsymbol{h}^{(l)}|\cdot)\approx p_{h}(\boldsymbol{h}^{(l)}|\boldsymbol{x}^{(l)}),
\end{align}
\vspace{-4mm}
\end{footnotesize}

\noindent where $\mathcal{IG}$ is the inverse gamma distribution. Here, $p_{\mu}$ and $p_{\sigma}$ perform as encoders, and the maximization of the distributions $p_{\mu}(\mu_l|\boldsymbol{h}^{(l)})$ and $p_{\sigma}(\sigma_l|\boldsymbol{h}^{(l)})$ can be approximated by multi-layer perceptrons (MLPs) with Gaussian outputs~\cite{kingma2014auto-encoding} as

\begin{footnotesize}
\vspace{-4mm}
\begin{align}
  \hat{\mu}_l&=\argmax_{\mu_l}\prod_{i=1}^N p_{\mu}(\mu_l|\boldsymbol{h}_i^{(l)})\nonumber\\
  &\approx\frac{1}{N}\sum_{i=1}^N\boldsymbol{\Omega}_{\mu}^{(l)}\boldsymbol{h}_i^{(l)}+\boldsymbol{b}_{\mu}^{(l)},\label{eq:muencoder}\\
  \hat{\sigma}_l&=\argmax_{\sigma_l}\prod_{i=1}^N p_{\sigma}(\sigma_l|\boldsymbol{h}_i^{(l)})\nonumber\\
  &\approx\frac{1}{N}\sum_{i=1}^N\text{Softplus}\left(\boldsymbol{\Omega}_{\sigma}^{(l)}\boldsymbol{h}_i^{(l)}+\boldsymbol{b}_{\sigma}^{(l)}\right),\label{eq:sigmaencoder}
\end{align}
\vspace{-2mm}
\end{footnotesize}

\noindent where $\boldsymbol{\gamma}^{(l)}=\{\boldsymbol{\Omega}_{\mu}^{(l)},\boldsymbol{\Omega}_{\sigma}^{(l)},\boldsymbol{b}_{\mu}^{(l)},\boldsymbol{b}_{\sigma}^{(l)}\}$ are the weights and the biases of the MLPs, $N$ is sample number, and the softplus function $\text{Softplus}(\cdot)$ is defined as

\begin{footnotesize}
\begin{equation}\label{eq:softplus}
  \text{Softplus}(x)=\ln(1+e^x).
\end{equation}
\vspace{-4mm}
\end{footnotesize}

Here, we define the inverse gamma approximation by the softplus function with Gaussian input as softplus-Gaussian distribution, which is a normalized distribution. To evaluate the effectiveness of the softplus-Gaussian distribution, we conduct a group of experiments in comparison with the classical log-normal distribution used in~\cite{kingma2014auto-encoding}, as shown in Figure~\ref{fig:softplusapprox}. The distributions approximate the true inverse gamma approximation by moment matching. The subfigures in Figure~\ref{fig:softplusapprox} show that the proposed softplus-Gaussian distribution is better than the referred one, due to smaller KL divergences in all the different cases.

In addition, the optimal hidden state $\hat{\boldsymbol{h}}^{(l)}$ (by maximizing $p_{h}(\boldsymbol{h}^{(l)}|\boldsymbol{x}^{(l)})$) can be inferred as

\begin{footnotesize}
\vspace{-4mm}
\begin{align}\label{eq:hprior}
  \hat{\boldsymbol{h}}^{(l)}&=\argmax_{\boldsymbol{h}^{(l)}}p_{h}(\boldsymbol{h}^{(l)}|\boldsymbol{x}^{(l)})\nonumber\\
  &\approx\boldsymbol{\Omega}_{h}^{(l)}\boldsymbol{x}^{(l)}+\boldsymbol{b}_{h}^{(l)},
\end{align}
\vspace{-4mm}
\end{footnotesize}

\noindent where $\boldsymbol{\delta}^{(l)}=\{\boldsymbol{\Omega}_{h}^{(l)},\boldsymbol{b}_{h}^{(l)}\}$ are the weights and the biases of the MLP.

\vspace{-4mm}
\subsection{Stochastic Gradient Variational Bayes (SGVB) for Advanced Dropout}\label{ssec:sgvb}

After constructing the whole advanced dropout technique as shown via the probabilistic graphical model in Figure~\ref{fig:graphicmodel}, we consider how to optimize it using the SGVB inference~\cite{kingma2014auto-encoding}.

We first define the dataset $\boldsymbol{D}=\{\boldsymbol{X},\boldsymbol{Y}\}$ for DNN training where $\boldsymbol{X}=\{\boldsymbol{x}_i\}_{i=1}^N$ and $\boldsymbol{Y}=\{\boldsymbol{y}_i\}_{i=1}^N$ are sets of input and target with $N$ samples, respectively, and the dropout-masked parameter set $\boldsymbol{W}=\{\boldsymbol{w}^{(l)}\}_{l=1}^L$ in which

\begin{footnotesize}
\begin{equation}\label{eq:maskedparam}
  \boldsymbol{w}^{(l)}=\text{diag}(\boldsymbol{m}^{(l)})\boldsymbol{\theta}^{(l)},
\end{equation}
\vspace{-4mm}
\end{footnotesize}

\noindent where $\text{diag}(\cdot)$ is the matrix operation that transforms a vector into a squared diagonal matrix with the vector as the main diagonal. In addition, we define the parameter set $\boldsymbol{\Lambda}=\{\boldsymbol{\lambda}_l\}_{l=1}^L$ where $\boldsymbol{\lambda}_l=\{\mu_l,\sigma_l,\boldsymbol{\gamma}^{(l)},\boldsymbol{\delta}^{(l)}\}$. The joint parameter set including original parameters $\boldsymbol{\Theta}$ and $\boldsymbol{\Lambda}$ is defined as $\boldsymbol{\Phi}=\{\boldsymbol{\Theta},\boldsymbol{\Lambda}\}$.

We can then obtain the joint distribution of the dataset $\boldsymbol{D}$ and the dropout-masked parameter set $\boldsymbol{W}$ as

\begin{footnotesize}
\begin{equation}\label{eq:jointdistrib}
  p(\boldsymbol{D},\boldsymbol{W})=p(\boldsymbol{D}|\boldsymbol{W})p(\boldsymbol{W}).
\end{equation}
\vspace{-4mm}
\end{footnotesize}

\noindent Introducing the approximated distribution $q_{\boldsymbol{\Phi}}(\boldsymbol{W}|\boldsymbol{Z})$ of $\boldsymbol{W}$ given $\boldsymbol{Z}=\{\boldsymbol{z}^{(0)},\cdots,\boldsymbol{z}^{(L-1)}\}$ ($\boldsymbol{z}^{(l)}=\boldsymbol{x}^{(l+1)}$) as

\begin{footnotesize}
\begin{equation}\label{eq:q}
  q_{\boldsymbol{\Phi}}(\boldsymbol{W}|\boldsymbol{Z})=\prod_l q_{\boldsymbol{\Phi}}(\boldsymbol{w}^{(l)}|\boldsymbol{z}^{(l-1)}),
\end{equation}
\vspace{-4mm}
\end{footnotesize}

\noindent we divide the left-hand side (LHS) and the right-hand side (RHS) of~\eqref{eq:jointdistrib} by $q_{\boldsymbol{\Phi}}(\boldsymbol{W}|\boldsymbol{Z})$ and take the logarithms of each side which give

\begin{footnotesize}
\begin{equation}\label{eq:logjointdistribdivq}
  \log\frac{p(\boldsymbol{D},\boldsymbol{W})}{q_{\boldsymbol{\Phi}}(\boldsymbol{W}|\boldsymbol{Z})} =\log\frac{p(\boldsymbol{D}|\boldsymbol{W})p(\boldsymbol{W})}{q_{\boldsymbol{\Phi}}(\boldsymbol{W}|\boldsymbol{Z})}.
\end{equation}
\vspace{-4mm}
\end{footnotesize}

We consider the expectation of the LHS and the RHS of~\eqref{eq:logjointdistribdivq},~\emph{w.r.t.} $q_{\boldsymbol{\Phi}}(\boldsymbol{W}|\boldsymbol{Z})$ as

\begin{footnotesize}
\vspace{-2mm}
\begin{align}\label{eq:expectation}
  &\underbrace{\int q_{\boldsymbol{\Phi}}(\boldsymbol{W}|\boldsymbol{Z})\log \frac{p(\boldsymbol{D},\boldsymbol{W})}{q_{\boldsymbol{\Phi}}(\boldsymbol{W}|\boldsymbol{Z})}\, d\,\boldsymbol{W}}_{L(\boldsymbol{\Phi})}\nonumber\\
  =&\underbrace{\int q_{\boldsymbol{\Phi}}(\boldsymbol{W}|\boldsymbol{Z})\log p(\boldsymbol{D}|\boldsymbol{W})\, d\,\boldsymbol{W}\vphantom{\frac{p(W)}{q_{\Phi}(W)p(D)}}}_{L_D(\boldsymbol{\Phi})}
  -\underbrace{\int q_{\boldsymbol{\Phi}}(\boldsymbol{W}|\boldsymbol{Z})\log \frac{q_{\boldsymbol{\Phi}}(\boldsymbol{W}|\boldsymbol{Z})}{p(\boldsymbol{W})}\, d\,\boldsymbol{W}}_{\text{D}_{\text{KL}}(q_{\boldsymbol{\Phi}}(\boldsymbol{W}|\boldsymbol{Z})||p(\boldsymbol{W}))},
\end{align}
\vspace{-4mm}
\end{footnotesize}

\noindent where the LHS $L(\boldsymbol{\Phi})$ is the lower bound of the expectation of posterior distribution $\text{E}_{\boldsymbol{\Phi}}[p(\boldsymbol{W}|\boldsymbol{D})]$ of $\boldsymbol{W}$ given $\boldsymbol{D}$ in the approximated variational inference, $L_D(\boldsymbol{\Phi})$ in the RHS is the expected log-likelihood of DNN training, and $\text{D}_{\text{KL}}(q_{\boldsymbol{\Phi}}(\boldsymbol{W}|\boldsymbol{Z})||p(\boldsymbol{W}))$ in the RHS is the KL divergence term from the approximated distribution of $\boldsymbol{W}$ to the prior distribution of $\boldsymbol{W}$ as a regularization term.

In the optimization of the approximated variational inference, we commonly maximize the RHS of~\eqref{eq:expectation}, rather than maximizing the lower bound $L(\boldsymbol{\Phi})$ directly. Here, in SGVB inference, we approximate the expected log-likelihood $L_D(\boldsymbol{\Phi})$ by a mini-batch-form log-likelihood as we usually apply the mini-batch SGD algorithm for DNN training. The approximated log-likelihood $L_D^{\text{SGVB}}(\boldsymbol{\Phi})$ for a mini batch can be considered as

\begin{footnotesize}
\vspace{-4mm}
\begin{align}\label{eq:sgvbll}
  L_D^{\text{SGVB}}(\boldsymbol{\Phi})&=\sum_{(\boldsymbol{x}_i,\boldsymbol{y}_i)\in(\boldsymbol{X}_b,\boldsymbol{Y}_b)}
  \int q_{\boldsymbol{\Phi}}(\boldsymbol{W}|\boldsymbol{Z})\log p(\boldsymbol{y}_i|\boldsymbol{x}_i,\boldsymbol{W})d\,\boldsymbol{W}\nonumber\\
  &=\frac{N}{N_b}\sum_{i=1}^{N_b}\log p(\boldsymbol{y}_i|\boldsymbol{x}_i,\boldsymbol{W}=f(\boldsymbol{\epsilon};\boldsymbol{\Phi},\boldsymbol{Z}))\nonumber\\
  &\approx L_D(\boldsymbol{\Phi}),
\end{align}
\vspace{-4mm}
\end{footnotesize}

\noindent where $(\boldsymbol{X}_b,\boldsymbol{Y}_b)=\{(\boldsymbol{x}_i, \boldsymbol{y}_i)\}_{i=1}^{N_b}$ is the $b^{th}$ mini batch stochastically selected from $\boldsymbol{D}$, $N_b=|(\boldsymbol{X}_b,\boldsymbol{Y}_b)|$ is the mini batch size, and $f(\boldsymbol{\epsilon};\boldsymbol{\Phi},\boldsymbol{Z})$ is a differentiable function for reparameterizing the dropout mask in the set $\boldsymbol{M}=\{\boldsymbol{m}^{(l)}\}_{l=1}^L$ by random samples $\boldsymbol{\epsilon}$, and for generating $\boldsymbol{W}$. Parameters in $\boldsymbol{\Lambda}$ are estimated directly by~\eqref{eq:sgvbll}, respectively.

\begin{table*}[!t]
  \centering
  \caption{\footnotesize Test accuracies ($\%$) on MNIST, CIFAR-$10$, CIFAR-$100$,~\emph{mini}ImageNet, and Caltech-$256$ datasets. Note that the best results are marked in~\textbf{bold} and the second best results are marked by~\underline{underline}, respectively.}
  \vspace{-3mm}
  \scriptsize
  \centering
  \resizebox{0.8\linewidth}{!}{
    \begin{tabular}{@{}l@{}c@{}c@{}c@{}c@{}c@{}}
    \toprule
    \multicolumn{1}{c}{Dataset} & MNIST & CIFAR-$10$ & CIFAR-$100$ &  \emph{mini}ImageNet   & Caltech-$256$ \\
    \multicolumn{1}{c}{Base model} & $784$-$2\!\times\!800$-$10$ & VGG$16$/ResNet$18$ & VGG$16$/ResNet$18$ & VGG$16$/ResNet$18$ & VGG$16$/ResNet$18$ \\
    \midrule
    No dropout                 & $98.23\pm0.11$ & $93.86\pm0.10$/$94.83\pm0.08$ & $73.62\pm0.38$/$76.44\pm0.14$ & $76.35\pm0.08$/$71.80\pm0.02$ & $63.87\pm0.07$/$61.36\pm0.10$ \\
    Dropout, Bernoulli     & $98.46\pm0.06$ & $93.81\pm0.14$/$94.84\pm0.07$ & $73.77\pm0.14$/$76.67\pm0.22$ & $76.21\pm0.09$/$72.02\pm0.40$ & $64.76\pm0.14$/$61.64\pm0.10$ \\
    Dropout, Gaussian      & $98.45\pm0.05$ & $93.83\pm0.14$/$95.02\pm0.14$ & $73.78\pm0.19$/$76.62\pm0.14$ & $76.14\pm0.09$/$71.98\pm0.45$ & $64.47\pm0.22$/$61.97\pm0.29$ \\
    Dropout, uniform        & $98.50\pm0.12$ & $93.82\pm0.09$/$94.86\pm0.14$ & $73.76\pm0.17$/$76.76\pm0.11$ & $76.84\pm0.07$/$72.07\pm0.02$ & $64.43\pm0.10$/\underline{$62.14\pm0.08$} \\
    Concrete dropout        & $98.45\pm0.04$ & $93.79\pm0.14$/$94.99\pm0.11$ & $73.67\pm0.16$/$76.48\pm0.26$ & $76.35\pm0.08$/$71.56\pm0.11$ & $63.79\pm0.07$/$62.12\pm0.08$ \\
    Variational dropout    & $98.46\pm0.14$ & $93.81\pm0.10$/$95.05\pm0.10$ & $73.98\pm0.25$/$76.76\pm0.38$ & $76.56\pm0.07$/$72.06\pm0.04$ & $64.46\pm0.11$/$61.98\pm0.18$ \\
    $\beta$-dropout          & $98.62\pm0.09$ & $93.95\pm0.19$/$95.07\pm0.11$ & $74.03\pm0.10$/$76.79\pm0.28$ & \underline{$77.13\pm0.13$}/$72.24\pm0.06$ & \underline{$64.84\pm0.07$}/$62.14\pm0.08$ \\
    Continuous dropout     & $98.45\pm0.20$ & $93.86\pm0.08$/$94.92\pm0.10$ & $73.85\pm0.21$/$76.90\pm0.27$ & $76.71\pm0.13$/\underline{$72.33\pm0.04$} & $64.69\pm0.09$/$61.80\pm0.16$ \\
    Information dropout    & $98.22\pm0.25$ & $93.88\pm0.18$/$94.97\pm0.16$ & $73.70\pm0.49$/$76.47\pm0.28$ & $76.44\pm0.08$/$71.90\pm0.06$ & $64.11\pm0.13$/$61.76\pm0.09$ \\
    Gaussian soft dropout & $98.64\pm0.04$ &\underline{$93.97\pm0.24$}/\underline{$95.09\pm0.09$} &\underline{$74.07\pm0.38$}/\underline{$77.22\pm0.23$} & $76.56\pm0.05$/$71.74\pm0.03$ & $63.80\pm0.10$/$58.68\pm0.24$ \\
    Laplace soft dropout  &\underline{$98.70\pm0.10$} & $93.95\pm0.11$/$95.03\pm0.11$ & $74.05\pm0.23$/$77.13\pm0.27$ & $76.61\pm0.06$/$71.55\pm0.02$ & $64.60\pm0.05$/$57.22\pm1.15$ \\
    Advanced dropout      & $\boldsymbol{98.89\pm0.04}$ & \ \ \ $\boldsymbol{94.28\pm0.03/95.52\pm0.09}$ &\ \ \  $\boldsymbol{74.94\pm0.24/77.78\pm0.08}$ &\ \ \  $\boldsymbol{77.35\pm0.01/72.89\pm0.06}$ &\ \ \  $\boldsymbol{65.09\pm0.03/62.53\pm0.01}$ \\
    \bottomrule
    \end{tabular}}
  \label{tab:results}
    \vspace{-2mm}
\end{table*}

\begin{table*}[!t]
  \centering
  \caption{\footnotesize The \emph{p}-values of student's~\emph{t}-tests between the accuracies of the advanced dropout technique and all the referred techniques with different base models on MNIST, CIFAR-$10$, CIFAR-$100$,~\emph{mini}ImageNet, and Caltech-$256$ datasets, respectively. The significance level was $0.05$. The advanced dropout technique has statistically significant difference from a referred technique if the corresponding~\emph{p}-value is smaller than $0.05$.}
  \vspace{-3mm}
  \scriptsize
  \resizebox{0.8\linewidth}{!}{
  \begin{tabular}{lccccc}
    \toprule
    \multicolumn{1}{c}{Dataset} & MNIST & CIFAR-$10$ & CIFAR-$100$ & \emph{Mini}ImageNet & Caltech-$256$ \\
    \multicolumn{1}{c}{Base model} & $784$-$2\!\times\!800$-$10$ & VGG$16$/ResNet$18$ & VGG$16$/ResNet$18$ & VGG$16$/ResNet$18$ & VGG$16$/ResNet$18$ \\
    \midrule
    No dropout & $9.04\times10^{-5}$ & $2.92\times10^{-4}$/$1.31\times10^{-6}$ & $3.59\times10^{-4}$/$1.69\times10^{-6}$ & $6.16\times10^{-6}$/$2.81\times10^{-8}$ & $1.09\times10^{-7}$/$8.60\times10^{-6}$ \\
    Dropout, Bernoulli & $7.32\times10^{-6}$ & $1.06\times10^{-3}$/$8.69\times10^{-7}$ & $1.92\times10^{-5}$/$1.30\times10^{-4}$ & $7.44\times10^{-6}$/$5.62\times10^{-3}$ & $8.43\times10^{-3}$/$2.64\times10^{-5}$ \\
    Dropout, Gaussian & $1.23\times10^{-6}$ & $1.37\times10^{-3}$/$3.39\times10^{-4}$ & $1.93\times10^{-5}$/$2.56\times10^{-6}$ & $4.26\times10^{-6}$/$4.84\times10^{-3}$ & $1.91\times10^{-3}$/$6.67\times10^{-3}$ \\
    Dropout, uniform & $2.25\times10^{-3}$ & $1.08\times10^{-4}$/$6.39\times10^{-5}$ & $1.55\times10^{-5}$/$7.18\times10^{-7}$ & $6.62\times10^{-5}$/$2.94\times10^{-7}$ & $6.53\times10^{-5}$/$3.34\times10^{-4}$ \\
    Concrete dropout & $3.00\times10^{-7}$ & $8.43\times10^{-4}$/$5.18\times10^{-5}$ & $9.07\times10^{-6}$/$1.62\times10^{-4}$ & $6.07\times10^{-6}$/$6.74\times10^{-7}$ & $1.35\times10^{-7}$/$2.45\times10^{-4}$ \\
    Variational dropout & $2.32\times10^{-3}$ & $1.86\times10^{-4}$/$7.19\times10^{-5}$ & $2.14\times10^{-4}$/$2.60\times10^{-3}$ & $1.12\times10^{-5}$/$7.26\times10^{-9}$ & $1.24\times10^{-4}$/$1.76\times10^{-3}$ \\
    $\beta$-dropout & $4.10\times10^{-3}$ & $1.18\times10^{-2}$/$1.25\times10^{-4}$ & $1.51\times10^{-4}$/$7.66\times10^{-4}$ & $1.30\times10^{-2}$/$1.76\times10^{-7}$ & $1.77\times10^{-3}$/$3.34\times10^{-4}$ \\
    Continuous dropout & $9.67\times10^{-3}$ & $1.12\times10^{-4}$/$8.41\times10^{-6}$ & $4.11\times10^{-5}$/$1.03\times10^{-3}$ & $2.76\times10^{-4}$/$1.24\times10^{-7}$ & $5.47\times10^{-4}$/$4.30\times10^{-4}$ \\
    Information dropout & $2.73\times10^{-3}$ & $3.76\times10^{-3}$/$2.52\times10^{-4}$ & $1.24\times10^{-3}$/$1.16\times10^{-4}$ & $8.20\times10^{-6}$/$2.22\times10^{-9}$ & $2.28\times10^{-5}$/$2.23\times10^{-5}$ \\
    Gaussian soft dropout & $5.58\times10^{-5}$ & $1.21\times10^{-2}$/$4.35\times10^{-5}$ & $1.88\times10^{-3}$/$7.72\times10^{-3}$ & $1.28\times10^{-6}$/$3.12\times10^{-8}$ & $1.19\times10^{-6}$/$1.73\times10^{-6}$ \\
    Laplace soft dropout & $2.91\times10^{-2}$ & $9.80\times10^{-4}$/$3.46\times10^{-5}$ & $1.53\times10^{-4}$/$2.05\times10^{-3}$ & $4.59\times10^{-6}$/$1.09\times10^{-7}$ & $1.52\times10^{-7}$/$2.47\times10^{-4}$ \\
    \bottomrule
  \end{tabular}}
  \label{tab:pvalue}
  \vspace{-6mm}
\end{table*}

For the $l^{th}$ layer as an example, the dropout-masked parameter matrix $\boldsymbol{w}^{(l)}$ can be computed as

\begin{footnotesize}
\vspace{-4mm}
\begin{align}\label{eq:wreparam}
  \boldsymbol{w}^{(l)}&=f\left(\boldsymbol{\epsilon}^{(l)};\boldsymbol{\theta}^{(l)},\boldsymbol{\lambda}_l\right)\nonumber\\
  &=\text{diag}(\boldsymbol{m}^{(l)})\boldsymbol{\theta}^{(l)}\nonumber\\
  &=\text{diag}\left(\text{Sigmoid}(\mu_l+\sigma_l\times\boldsymbol{\epsilon}^{(l)})\right)\boldsymbol{\theta}^{(l)},
\end{align}
\vspace{-4mm}
\end{footnotesize}

\noindent where element $\epsilon_j^{(l)}\sim\mathcal{N}(0,1),j=1,\cdots,K_l$ in $\boldsymbol{\epsilon}^{(l)}$ is randomly sampled from the normal distribution with zero mean and unit variance.

\vspace{-3mm}
\section{Experimental Results And Discussions}\label{sec:experiments}

\subsection{Advanced Dropout for Classification}\label{ssec:classification}

\subsubsection{Datasets}\label{sssec:datasets}

We evaluated the advanced dropout technique on {seven} image classification datasets, including MNIST~\cite{mnist98}, CIFAR-$10$ and -$100$~\cite{krizhevsky09cifar},~\emph{mini}ImageNet~\cite{vinyals2016matching}, Caltech-$256$~\cite{griffin2006caltech}, ImageNet-$32\!\times\!32$~\cite{denoord2016pixel}, {and ImageNet~\cite{russakovsky2015imagenet}} datasets. Note that in the~\emph{mini}ImageNet dataset, $500$ and $100$ images per class were randomly selected from the training set of the full ImageNet dataset as the training set and the test set of this work, respectively. In the Caltech-$256$ dataset, $60$ samples were randomly selected from each class, gathering as the training set. The remaining samples were for the test set. The ImageNet-$32\!\times\!32$ dataset is more difficult than the full ImageNet dataset (with $224\!\times\!224$ images in general), since all the images are downsampled to $32\!\times\!32$ for both training and test~\cite{denoord2016pixel}. {The former five ones are small-scale datasets, while the ImageNet-$32\!\times\!32$ and the ImageNet datasets are large-scale ones.} We will separately discuss the two kinds.

\vspace{-2mm}
\subsubsection{Implementation Details}\label{sssec:implement}

For the MNIST dataset, a FC neural network with two hidden layers and $800$ hidden nodes each was constructed for the evaluation, while VGG$16$~\cite{simonyan2015very} and ResNet$18$~\cite{he2016deep} models were considered as the base models for all the other datasets.

In DNN training on the MNIST dataset, we trained the FC neural network with different dropout techniques for $200$ epochs with the fixed learning rate as $0.01$. For the ImageNet-$32\!\times\!32$ dataset, we trained the models under $100$ epochs with the initialized learning rate as $0.1$ and decayed by a factor of $10$ at the $50^{th}$ and the $75^{th}$ epochs, respectively. For all the other datasets, we trained the models $300$ epochs with the initialized learning rate as $0.1$ and decayed by a factor of $10$ at the $150^{th}$ and the $225^{th}$ epochs, respectively. The batch size on each dataset was set as $256$. In addition, we adopted the SGD optimizer, of which the momentum and the weight decay values were set as $0.9$ and $5\times10^{-4}$, respectively. All the models with the advanced dropout technique and the referred techniques have been experimented $5$ runs with random initialization, {while we compared the results of the advanced dropout technique on the ImageNet dataset (with standard experimental settings) with those reported in references~\cite{shen2018continuous,liu2019beta}}. The means and the standard deviations of the classification accuracies are presented in the following sections for comparison.

Meanwhile, nine referred techniques were selected based on their distributions of the dropout masks including standard dropout with Bernoulli noise (``dropout, Bernoulli'')~\cite{hinton12}, Gaussian noise (``dropout, Gaussian'')~\cite{Srivastava14}, and uniform noise (``dropout, uniform'')~\cite{shen2018continuous}, variational dropout~\cite{kingma2015variational}, concrete dropout~\cite{gal17concrete}, continuous dropout~\cite{shen2018continuous}, $\beta$-dropout~\cite{liu2019beta}, information dropout~\cite{achille2018information}, and soft dropout~\cite{xie2019soft}. We reimplemented and compared them with the advanced dropout technique. The two adaptive version of the soft dropout,~\emph{i.e.}, Gaussian soft dropout and Laplace soft dropout, were implemented on all the datasets for comparison, respectively. For the standard dropout with Bernoulli noise and Gaussian noise, we fixed the dropout rate as $0.5$ during training. For the continuous dropout, we selected the variance $\sigma^2$ of its Gaussian prior from the set $\{0.2,0.3,0.4\}$, following the strategy in~\cite{shen2018continuous}. Meanwhile, for the $\beta$-dropout, the shape parameter $\beta$ was set as values drawn from the set $\{0.001,0.1,0.2,0.5,1,3\}$, as suggested in~\cite{liu2019beta}. For the information dropout, the multiplier $\beta$ in its loss function was empirically selected as $0.1$.

In order to check whether the advanced dropout technique has statistically significant performance improvement compared with the referred techniques, we conducted Student's~\emph{t}-tests between accuracies of them with the null-hypothesis that the means of two populations are equal. The significance level was set as $0.05$.

\begin{table*}[!t]
  \centering
  \caption{\footnotesize Test accuracies ($\%$) and the $p$-values of Student's \emph{t}-tests between the accuracies of the advanced dropout technique and all the referred techniques on the ImageNet-$32\!\times\!32$ dataset. Note that the best results are marked in \textbf{bold} and the second best results are marked by \underline{underline}, respectively. The significance level was set as $0.05$.}
  \vspace{-3mm}
  \scriptsize
  \resizebox{0.8\linewidth}{!}{
  \begin{tabular}{lcccc}
    \toprule
    \multicolumn{1}{c}{ Dataset} & \multicolumn{2}{c}{ VGG$16$} & \multicolumn{2}{c}{ ResNet$18$} \\
    \multicolumn{1}{c}{ Base model} &  Top-$1$ acc./$p$-value &  Top-$5$ acc./$p$-value &  Top-$1$ acc./$p$-value &  Top-$5$ acc./$p$-value \\
    \midrule
     No dropout &  $40.58\pm0.15$/$3.57\times10^{-8}$ &  $64.50\pm0.22$/$1.56\times10^{-6}$ &  $45.46\pm0.39$/$5.44\times10^{-4}$ &  $70.47\pm0.38$/$9.05\times10^{-4}$ \\
     Dropout, Bernoulli &  $41.26\pm0.21$/$1.67\times10^{-5}$ &  $64.14\pm0.12$/$1.63\times10^{-8}$ &  $45.72\pm0.13$/$1.46\times10^{-6}$ &  $70.73\pm0.08$/$4.24\times10^{-7}$ \\
     Dropout, Gaussian &  $41.33\pm0.14$/$3.36\times10^{-7}$ &  $64.22\pm0.08$/$9.44\times10^{-8}$ &  $45.00\pm0.07$/$7.09\times10^{-1}$ &  $69.74\pm0.19$/$6.72\times10^{-7}$ \\
     Dropout, uniform &  $41.36\pm0.16$/$2.35\times10^{-6}$ &  $64.88\pm0.15$/$5.63\times10^{-7}$ &  $45.26\pm0.08$/$5.27\times10^{-0}$ &  $69.62\pm0.07$/$2.47\times10^{-9}$ \\
     Concrete dropout &  $41.37\pm0.05$/$1.73\times10^{-7}$ &  $64.33\pm0.19$/$2.90\times10^{-7}$ &  $45.75\pm0.12$/$7.26\times10^{-7}$ &  $70.55\pm0.17$/$5.68\times10^{-6}$ \\
     Variational dropout &  $41.45\pm0.06$/$2.29\times10^{-7}$ &  $64.35\pm0.10$/$5.48\times10^{-8}$ &  $45.53\pm0.09$/$4.26\times10^{-9}$ &  $69.63\pm0.14$/$1.09\times10^{-8}$ \\
     $\beta$-dropout &  $41.51\pm0.08$/$8.15\times10^{-8}$ &  $64.27\pm0.13$/$2.85\times10^{-8}$ &  $45.09\pm0.08$/$1.47\times10^{-10}$ &  $69.73\pm0.13$/$6.11\times10^{-9}$ \\
     Continuous dropout &  $41.23\pm0.11$/$2.63\times10^{-8}$ &  $64.07\pm0.08$/$8.76\times10^{-8}$ &  $44.60\pm0.12$/$6.45\times10^{-9}$ &  $69.39\pm0.09$/$5.56\times10^{-10}$ \\
     Information dropout &  $41.35\pm0.12$/$1.50\times10^{-7}$ &  $64.12\pm0.09$/$6.00\times10^{-8}$ &  $45.55\pm0.07$/$6.71\times10^{-10}$ &  $70.31\pm0.12$/$7.02\times10^{-8}$ \\
     Gaussian soft dropout &  $\underline{41.69\pm0.07}$/$4.30\times10^{-7}$ &  $64.59\pm0.10$/$2.47\times10^{-7}$ &  $\underline{46.33\pm0.13}$/$4.06\times10^{-5}$ &  $71.04\pm0.06$/$1.09\times10^{-5}$ \\
     Laplace soft dropout &  $41.57\pm0.14$/$1.33\times10^{-6}$ &  $\underline{65.45\pm0.10}$/$2.88\times10^{-5}$ &  $46.31\pm0.19$/$3.85\times10^{-4}$ &  $\underline{71.14\pm0.06}$/$3.14\times10^{-5}$ \\
     Advanced dropout &  $\boldsymbol{42.65\pm0.11}$/ N/A &  $\boldsymbol{66.23\pm0.16}$/ N/A &  $\boldsymbol{47.03\pm0.07}$/ N/A &  $\boldsymbol{71.75\pm0.12}$/ N/A \\
    \bottomrule
  \end{tabular}}
  \label{tab:imagenet32x32}
  \vspace{-6mm}
\end{table*}

\begin{table}[!t]
\vspace{1mm}
  \centering
  \caption{\footnotesize Test accuracies ($\%$) and the $p$-values of Student's \emph{t}-tests between the accuracies of the advanced dropout technique and all the referred techniques on the ImageNet dataset. Note that the best results are marked in \textbf{bold} and the second best results are marked by \underline{underline}, respectively. The significance level was set as $0.05$. Note that ``$\dagger$'' means the results in the row are reported in~\cite{shen2018continuous} and ``$\ddagger$'' means the results in the row are reported in~\cite{liu2019beta}.}
  \vspace{-3mm}
  \resizebox{0.95\linewidth}{!}{
    \begin{tabular}{lcc}
    \toprule
    \multicolumn{1}{c}{Method} & Top-$1$ acc./$p$-value & Top-$5$ acc./$p$-value \\
    \midrule
    No dropout & $73.42\pm0.059$/$2.43\times10^{-11}$ & $91.68\pm0.025$/$6.30\times10^{-10}$ \\
    Adaptive dropout$^\dagger$ & $73.73\pm0.046$/$2.65\times10^{-16}$ & $91.59\pm0.061$/$6.56\times10^{-12}$ \\
    DropConnect$^\dagger$ & $73.18\pm0.050$/$3.70\times10^{-18}$ & $91.44\pm0.037$/$2.04\times10^{-15}$ \\
    Dropout, Bernoulli$^\dagger$ & $73.01\pm0.065$/$2.71\times10^{-17}$ & $91.14\pm0.042$/$1.36\times10^{-16}$ \\
    Dropout, Gaussian$^\dagger$ & $74.21\pm0.045$/$3.60\times10^{-13}$ & $92.01\pm0.065$/$1.98\times10^{-6}$ \\
    Dropout, uniform$^\dagger$ & $74.09\pm0.046$/$4.67\times10^{-14}$ & $91.92\pm0.048$/$2.25\times10^{-9}$ \\
    $\beta$-dropout$^\ddagger$ & $\underline{74.25}\pm\underline{0.052}$/$4.92\times10^{-12}$ & $\underline{92.04}\pm\underline{0.046}$/$2.64\times10^{-7}$ \\
    Continuous dropout$^\dagger$ & $74.21\pm0.045$/$3.60\times10^{-13}$ & $92.01\pm0.065$/$1.98\times10^{-6}$ \\
    Advanced dropout & $\boldsymbol{74.82\pm0.017}$/N/A & $\boldsymbol{92.26\pm0.029}$/N/A \\
    \bottomrule
    \end{tabular}}
  \label{tab:imagenet}
  \vspace{-6mm}
\end{table}

\vspace{-1mm}
\subsubsection{Performance on MNIST Dataset}\label{sssec:mnist}

From Table~\ref{tab:results}, the advanced dropout technique achieves the best accuracy at $98.89\%$ among the other techniques with the two-hidden-layer FC neural network on the MNIST dataset. At the meantime, it obtains the smallest standard deviation at $0.04\%$ as well. While the second best technique, the Laplace soft dropout, achieves the classification accuracy at $98.70\%$, the proposed technique outperforms the second best one slightly at about $0.2\%$, but has an increase at about $0.7\%$ compared with the model without dropout.

\vspace{-1mm}
\subsubsection{Performance on CIFAR-$\textit{10}$ and -$\textit{100}$ Datasets}\label{sssec:cifars}

On the CIFAR-$10$ dataset, the advanced dropout technique performs best again with both VGG$16$ and ResNet$18$ models. It achieves the averaged classification accuracies of $94.28\%$ and $95.52\%$ for each base model, respectively, while the accuracies of all the others are less than $94\%$ and $95.1\%$ with the VGG$16$ and the ResNet$18$ base models, respectively. Meanwhile, the proposed technique outperforms the second best technique, by about $0.3\%$ and $0.5\%$, and the base model by about $0.4\%$ and $0.7\%$, respectively.

Moreover, the advanced dropout technique achieves the classification accuracies of $74.94\%$ and $77.78\%$ on the CIFAR-$100$ dataset with the VGG$16$ and the ResNet$18$ base models, respectively, which are the best results among all the referred techniques. The VGG$16$ model with the advanced dropout technique obtains an improvement of $0.9\%$ and $1.3\%$, compared with the Gaussian soft dropout and the base model, while the ResNet$18$ with it also achieves more than $0.5\%$ and $1.3\%$ higher performance compared with the Gaussian soft dropout and the base model.

\vspace{-1mm}
\subsubsection{Performance on MiniImageNet Dataset}\label{sssec:miniimagenet}

In Table~\ref{tab:results}, the advanced dropout technique achieves the classification accuracies at $77.35\%$ and $72.89\%$, respectively, which is the best results with the VGG$16$ and the ResNet$18$ models among the referred methods on the~\emph{mini}ImageNet dataset. The accuracies obtained by most of the referred techniques with the VGG$16$ model are lower than $77\%$. The proposed technique outperforms the base model by $1\%$ and the second best model, the $\beta$-dropout with the VGG$16$ model, by more than $0.2\%$.

Furthermore, the advanced dropout-based ResNet$18$ model also shows improvement in classification accuracies on the~\emph{mini}ImageNet dataset. Compared with the base model, the advanced dropout technique with the ResNet$18$ model increases its accuracy by about $1.1\%$, which is a promising performance improvement. Meanwhile, it surpasses continuous dropout with the ResNet$18$ ($72.33\%$) by more than $0.5\%$.

\vspace{-1mm}
\subsubsection{Performance on Caltech-$\textit{256}$ Dataset}\label{sssec:caltech}

In the last column of Table~\ref{tab:results}, the advanced dropout technique with the VGG$16$ and the ResNet$18$ models performs the best on the Caltech-$256$ dataset and achieve the averaged accuracies of $65.09\%$ and $62.53\%$ with the smallest standard deviations of each at $0.03\%$ and $0.01\%$, respectively. Compared with their corresponding base models, the models based on the advanced dropout technique improve the accuracies by a large margin of around $1.2\%$ for both. Applying the VGG$16$ model as the base model, the classification accuracy of the advanced dropout is higher than the second best model, the $\beta$-dropout, more than $0.2\%$. Meanwhile, with the ResNet$18$ base model, the advanced dropout technique shows considerable performance improvements compared with the base model and the second best model as well. It improves the accuracy by $0.4\%$ more than the second best technique, the dropout with uniform noise.

\vspace{-1mm}
\subsubsection{Performance on ImageNet-$32\!\times\!32$ Dataset}\label{sssec:imagenet32x32}

In this section, we evaluate the proposed advanced dropout on a large-scale dataset,~\emph{i.e.}, the ImageNet-$32\!\times\!32$ dataset. The accuracies and $p$-values are listed in Table~\ref{tab:imagenet32x32}. According to the table, the advanced dropout achieves statistically significant improvements over all the referred methods, improving top-$1$ accuracy by about $1\%$ and $0.7\%$ and top-$5$ accuracy by about $0.8\%$ and $0.6\%$ with VGG$16$ and ResNet$18$ models, respectively.

\begin{table*}[!t]
  \centering
  \caption{\footnotesize Comparison of training time per epoch (in second/epoch) on MNIST, CIFAR-$10$, CIFAR-$100$,~\emph{mini}ImageNet, Caltech-$256$, and ImageNet-$32\!\times\!32$ datasets by using one NVIDIA GTX$1080$Ti. All the models with different dropout variants are trained under equal conditions. The shortest time and the longest time on each dataset are highlighted in~\textbf{bold} and~\textit{italic}, respectively.}
  \vspace{-3mm}
  \scriptsize
  \resizebox{0.8\linewidth}{!}{
    \begin{tabular}{lcccccc}
    \toprule
    \multicolumn{1}{c}{ Dataset} &  MNIST &  CIFAR-$10$ &  CIFAR-$100$ &  \emph{Mini}ImageNet &  Caltech-$256$ &  ImageNet-$32\!\times\!32$ \\
    \multicolumn{1}{c}{ Base model} &  $784$-$2\times800$-$10$ &  VGG$16$/ResNet$18$ &  VGG$16$/ResNet$18$ &  VGG$16$/ResNet$18$ &  VGG$16$/ResNet$18$ &  VGG$16$/ResNet$18$ \\
    \midrule
     No dropout &  $\boldsymbol{3.95}$  &  $\boldsymbol{18.34}$/$\boldsymbol{38.07}$  &  $\boldsymbol{45.41}$/$\boldsymbol{20.62}$  &  $\boldsymbol{301.55}$/$\boldsymbol{174.75}$  &  $\boldsymbol{127.74}$/$\boldsymbol{60.20}$ &  $\boldsymbol{406.87}$/$\boldsymbol{556.13}$ \\
     Dropout, Bernoulli &  $4.00$  &  $23.32$/$44.65$  &  $48.81$/$26.95$  &  $302.47$/$175.18$  &  $127.77$/$62.50$ &  $411.62$/$572.08$ \\
     Dropout, Gaussian &  $9.45$  &  $\emph{24.69}$/$\emph{57.83}$  &  $49.77$/$53.54$  &  $365.86$/$183.41$  &  $148.12$/$70.09$ &  $438.49$/$579.57$ \\
     Dropout, uniform &  $4.22$  &  $20.87$/$41.87$  &  $52.55$/$46.56$  &  $307.60$/$177.50$  &  $136.59$/$65.71$ &  $416.47$/$572.70$ \\
     Concrete dropout &  $4.55$  &  $20.57$/$41.58$  &  $52.93$/$36.79$  &  $308.47$/$175.54$  &  $128.62$/$64.35$ &  $425.68$/$564.60$ \\
     Variational dropout &  $5.47$  &  $20.89$/$41.38$  &  $54.17$/$37.28$  &  $309.19$/$176.44$  &  $128.27$/$64.81$ &  $446.90$/$581.87$ \\
     $\beta$-dropout &  $\emph{15.15}$  &  $24.27$/$55.00$  &  $\emph{63.84}$/$\emph{57.75}$  &  $\emph{520.72}$/$\emph{210.97}$  &  $\emph{221.21}$/$\emph{90.25}$ &  $\emph{605.93}$/$\emph{696.83}$ \\
     Continuous dropout &  $4.55$  &  $19.81$/$52.66$  &  $53.93$/$49.33$  &  $319.40$/$178.57$  &  $132.48$/$66.20$ &  $420.53$/$559.86$ \\
     Information dropout &  $4.93$  &  $22.52$/$53.06$  &  $52.21$/$42.44$  &  $309.12$/$175.60$ &  $127.96$/$65.11$ &  $413.30$/$575.36$ \\
     Gaussian soft dropout &  $6.15$  &  $21.25$/$47.10$  &  $58.17$/$44.10$  &  $321.40$/$179.45$  &  $138.53$/$67.18$ &  $418.97$/$586.96$ \\
     Laplace soft dropout &  $5.52$  &  $22.01$/$45.19$  &  $51.49$/$42.39$  &  $321.01$/$179.04$  &  $135.81$/$66.35$ &  $425.97$/$582.75$ \\
     Advanced dropout &  $6.87$  &  $21.08$/$52.16$  &  $59.43$/$48.93$  &  $323.18$/$181.20$  &  $163.28$/$65.41$ &  $419.59$/$582.22$ \\
    \bottomrule
    \end{tabular}}
  \label{tab:trainingtime}
  \vspace{-4mm}
\end{table*}

\begin{figure*}[!t]
  \centering
  \begin{subfigure}[t]{0.38\linewidth}
    \centering
    \includegraphics[width=1\linewidth]{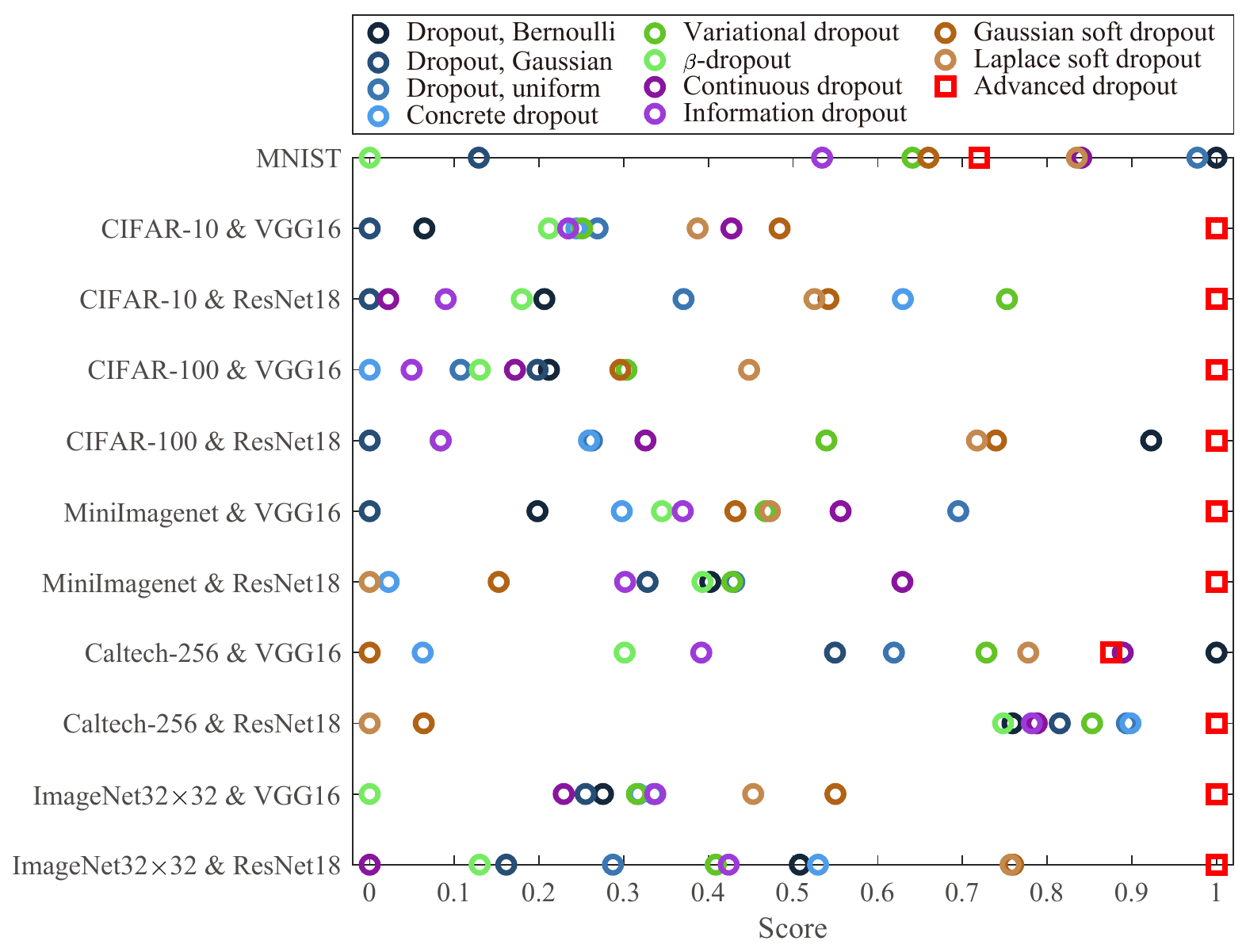}
    \vspace{-6mm}
    \subcaption{\footnotesize Normalized effectiveness ratio with ``metric $1$''}
  \end{subfigure}
  \begin{subfigure}[t]{0.38\linewidth}
    \centering
    \includegraphics[width=1\linewidth]{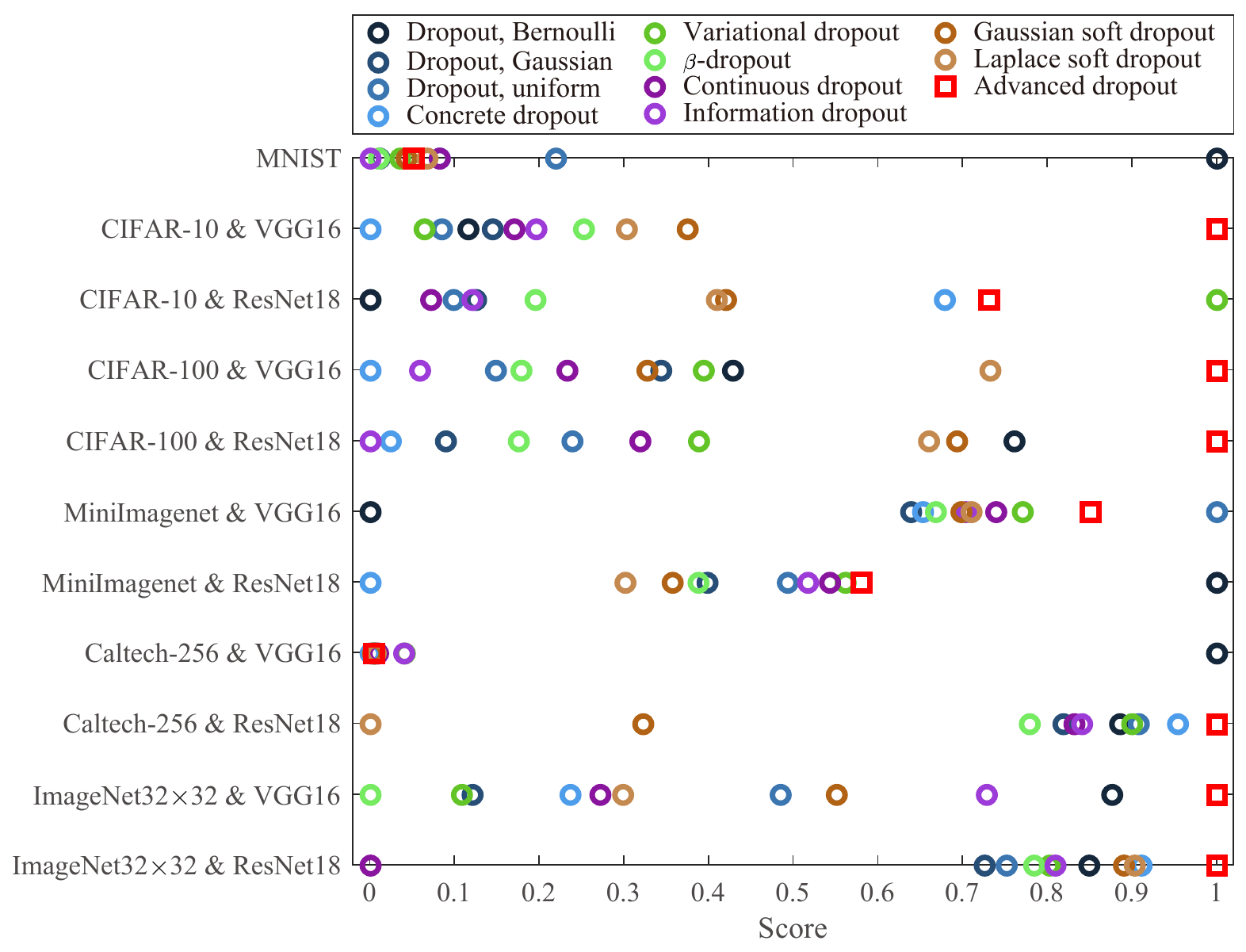}
    \vspace{-6mm}
    \subcaption{\footnotesize Normalized effectiveness ratio with ``metric $2$''}
  \end{subfigure}
  \vspace{-2mm}
  \caption{\footnotesize Normalized effectiveness ratios of the advanced dropout technique and various dropout techniques for quantitative comparison of accuracy improvement over training time. We show the results with the two metrics in (a) and (b), respectively. The red squares are the proposed advanced dropout technique, while other circles represent the referred dropout techniques. Note that for the ImageNet-$32\!\times\!32$ dataset, top-$1$ accuracies are used only.}\label{fig:accvstime}
  \vspace{-6mm}
\end{figure*}

\vspace{-1mm}
\subsubsection{{Performance on ImageNet Dataset}}\label{sssec:imagenet}

{In this section, the proposed advanced dropout was evaluated on the ImageNet dataset with VGG$16$ model as the base model. The accuracies and the corresponding $p$-values are listed in Table~\ref{tab:imagenet}. According to the table, the advanced dropout achieves statistically significant improvements over all the referred methods, improving top-$1$ accuracy by about $0.6\%$ and top-$5$ accuracy by about $0.2\%$, respectively.}

\vspace{-1mm}
\subsubsection{Trade-off between Performance and Time}\label{sssec:trainingtime}

In this section, we compare the training time (in second/epoch) of the proposed advanced dropout with those of the other dropout variants by using one NVIDIA GTX$1080$Ti. Note that the dropout techniques for regularization are only used when training the models. Their inference time is the same as the models without dropouts. Thus, we only compare the training time on MNIST, CIFAR-$10$, CIFAR-$100$,~\emph{Mini}ImageNet, Caltech-$256$, and ImageNet-$32\!\times\!32$ datasets, respectively. According to the results shown in Table~\ref{tab:trainingtime}, the models without dropouts always take the shortest time. Meanwhile, the proposed advanced dropout indeed required relatively longer time in each epoch, but it is not the slowest one. Moreover, the differences among the dropout techniques are not large. This indicates that time consumption for the dropout technique is mainly for sampling the dropout masks and the parametric prior does not take much training effort.

Furthermore, in order to quantitatively compare the accuracy improvement over the training time, we design a effectiveness ratio, which is similar to the cost effectiveness in~\cite{li2016efficient}, to rank all the dropout techniques based on their differences from the base model without dropout. A higher effectiveness ratio means more accuracy improvement can be achieved over the same unit of training time cost. We define the two metrics for the calculation of the effectiveness ratio and name them as ``metric $1$'' and ``metric $2$'', respectively. For the ``metric $1$'', the effectiveness ratio $s_1$ is 

\begin{footnotesize}
\vspace{-2mm}
\begin{equation}\label{eq:score}
  s_1=\frac{\text{Sigmoid}\left(a-a'\right)}{\text{Sigmoid}\left((t-t')/t'\right)},
\end{equation}
\vspace{-2mm}
\end{footnotesize}

\noindent where $a$ and $t$ are the accuracy (in $\%$) and training time (in second/epoch) of a dropout technique, and $a'$ and $t'$ are those of the base model with no dropout. The Sigmoid functions here are for scaling the accuracy improvement and the training time into the same range. For the ``metric $2$'', the Sigmoid functions are removed and the effectiveness ratio $s_2$ is

\begin{footnotesize}
\vspace{-2mm}
\begin{equation}\label{eq:score}
  s_2=\frac{(a-a')/a'}{(t-t')/t'}.
\end{equation}
\vspace{-2mm}
\end{footnotesize}

For better illustration, we normalize all the effectiveness ratios of different dropout techniques from one dataset by their maximum and minimum values. The normalized effectiveness ratio $s^i_{\text{norm}}$ of the ``metric $i$'' ($i=1,2$) for one specific dataset is defined as

\begin{footnotesize}
\vspace{-2mm}
\begin{equation}\label{eq:normalizedscore}
  s^i_{\text{norm}}=\frac{s-\min_{s'\in\mathcal{S}_i}s'}{\max_{s'\in\mathcal{S}_i}s'-\min_{s'\in\mathcal{S}_i}s'},s\in\mathcal{S}_i,
\end{equation}
\vspace{-2mm}
\end{footnotesize}

\noindent where $\mathcal{S}_i,i=1,2$ is the set of effectiveness ratios of the dropout techniques calculated by the two metrics, respectively. Here, ``normalized'' means mapping the effectiveness ratios from one dataset and in a base model (called a case) into the interval of $[0,1]$ for comparing their relative values.

Figure~\ref{fig:accvstime} demonstrates the normalized effectiveness ratios of different dropout techniques on each dataset. By using ``metric $1$'' (Figure~\ref{fig:accvstime}(a)), the advanced dropout can outperform all the other dropout techniques on nine cases out of all the eleven cases. Meanwhile, with ``metric $2$'' (Figure~\ref{fig:accvstime}(b)), the advanced dropout can outperform all the other dropout techniques on six cases in all eleven cases and perform the second best on three cases.
\vspace{-1.5mm}
\subsubsection{Performance with other Base Models}\label{sssec:otherbackbones}

We further evaluated the advanced dropout technique with three other base models, including DenseNet$40$~\cite{huang2017densely}, MobileNet~\cite{howard2017mobilenets}, wide residual network (WRN)~\cite{zagoruyko2016wide}, on both the CIFAR-$10$ and -$100$ datasets. For hyperparameter selection of the base models, the growth rate and the reduction rate were set as $12$ and $0.5$ for the DenseNet$40$, respectively. $\alpha$ was set as $1$ for the MobileNet, and $k$ was set as $10$ for the WRN$16$. The optimizer, the batch sizes, and the learning rates were set the same as those of the previous settings. All the models with the advanced dropout technique and the referred techniques were experimented $5$ runs with random initialization. In addition, Student's~\emph{t}-tests was conducted between the accuracies of the advanced dropout technique and all the referred techniques.

The accuracies and the $p$-values are listed in Table~\ref{tab:otherbackbones}. It can be observed that the advanced dropout achieves statistically significant improvement on all the base models.

\begin{table}[!t]
  \centering
  \caption{\footnotesize Test accuracies ($\%$) and $p$-values of Student's \emph{t}-tests between the accuracies of the advanced dropout technique and all the referred techniques with other base models on CIFAR-$10$ and -$100$ datasets, respectively. Note that the best results are marked in \textbf{bold} and the second best results are marked by \underline{underline}, respectively. The significance level was set as $0.05$.}
  \vspace{-3mm}
  \scriptsize
  \resizebox{1.1\linewidth}{!}{
  \begin{tabular}{lccc}
    \toprule
    \multicolumn{1}{c}{ Dataset} & \multicolumn{3}{c}{ CIFAR-$10$} \\
    \multicolumn{1}{c}{ Base model} &  DenseNet$40$/$p$-value &  MobileNet/$p$-value &  WRN$16$-$10$/$p$-value \\
    \midrule
     No dropout &  $92.54\pm0.33$/$1.47\times10^{-3}$ &  $90.82\pm0.12$/$8.99\times10^{-6}$ &  $94.29\pm0.37$/$9.36\times10^{-4}$ \\
     Dropout, Bernoulli &  $92.06\pm0.26$/$4.96\times10^{-5}$ &  $90.95\pm0.12$/$1.32\times10^{-5}$ &  $94.65\pm0.40$/$4.59\times10^{-3}$ \\
     Dropout, Gaussian &  $92.05\pm0.25$/$3.70\times10^{-5}$ &  $91.23\pm0.19$/$2.04\times10^{-3}$ &  $94.93\pm0.12$/$2.22\times10^{-4}$ \\
     Dropout, uniform &  $92.41\pm0.21$/$6.69\times10^{-5}$ &  $90.95\pm0.30$/$2.36\times10^{-3}$ &  $94.83\pm0.35$/$6.59\times10^{-3}$ \\
     Concrete dropout &  $92.50\pm0.26$/$3.95\times10^{-4}$ &  $91.04\pm0.17$/$3.09\times10^{-4}$ &  $94.94\pm0.05$/$1.25\times10^{-6}$ \\
     Variational dropout &  $92.71\pm0.15$/$2.87\times10^{-5}$ &  $91.17\pm0.27$/$5.71\times10^{-3}$ &  $\underline{94.98\pm0.11}$/$1.65\times10^{-4}$ \\
     $\beta$-dropout &  $92.41\pm0.21$/$5.18\times10^{-5}$ &  $90.94\pm0.25$/$9.98\times10^{-4}$ &  $94.83\pm0.12$/$1.02\times10^{-4}$ \\
     Continuous dropout &  $92.08\pm0.18$/$3.20\times10^{-6}$ &  $90.85\pm0.34$/$2.47\times10^{-3}$ &  $94.95\pm0.13$/$3.38\times10^{-4}$ \\
     Information dropout &  $92.68\pm0.15$/$2.14\times10^{-5}$ &  $91.29\pm0.20$/$4.82\times10^{-3}$ &  $94.94\pm0.13$/$3.19\times10^{-4}$ \\
     Gaussian soft dropout &  $\underline{92.73\pm0.17}$/$8.64\times10^{-5}$ &  $\underline{91.34\pm0.22}$/$1.17\times10^{-2}$ &  $94.97\pm0.07$/$1.94\times10^{-5}$ \\
     Laplace soft dropout &  $92.71\pm0.20$/$2.20\times10^{-4}$ &  $91.34\pm0.25$/$1.65\times10^{-2}$ &  $94.87\pm0.05$/$9.74\times10^{-7}$ \\
     Advanced dropout &  $\boldsymbol{93.41\pm0.09}$/ N/A &  $\boldsymbol{91.76\pm0.05}$/ N/A &  $\boldsymbol{95.50\pm0.02}$/ N/A \\
    \bottomrule
    \toprule
    \multicolumn{1}{c}{ Dataset} & \multicolumn{3}{c}{ CIFAR-$100$} \\
    \multicolumn{1}{c}{ Base model} &  DenseNet$40$/$p$-value &  MobileNet/$p$-value &  WRN$16$-$10$/$p$-value \\
    \midrule
     No dropout &  $68.36\pm0.27$/$6.75\times10^{-7}$ &  $67.12\pm0.25$/$1.89\times10^{-6}$ &  $77.67\pm0.19$/$2.83\times10^{-3}$ \\
     Dropout, Bernoulli &  $68.39\pm0.27$/$7.10\times10^{-7}$ &  $67.39\pm0.51$/$7.27\times10^{-4}$ &  $77.55\pm0.18$/$1.35\times10^{-4}$ \\
     Dropout, Gaussian &  $68.43\pm0.31$/$3.50\times10^{-6}$ &  $67.30\pm0.41$/$1.25\times10^{-4}$ &  $77.68\pm0.16$/$4.95\times10^{-4}$ \\
     Dropout, uniform &  $68.29\pm0.36$/$8.24\times10^{-6}$ &  $\underline{67.87\pm0.12}$/$8.34\times10^{-5}$ &  $77.63\pm0.14$/$3.80\times10^{-3}$ \\
     Concrete dropout &  $68.48\pm0.49$/$1.19\times10^{-4}$ &  $67.87\pm0.23$/$7.79\times10^{-5}$ &  $77.69\pm0.16$/$5.24\times10^{-5}$ \\
     Variational dropout &  $\underline{69.70\pm0.28}$/$2.38\times10^{-4}$ &  $67.81\pm0.27$/$1.04\times10^{-4}$ &  $77.73\pm0.17$/$1.26\times10^{-3}$ \\
     $\beta$-dropout &  $68.52\pm0.35$/$1.08\times10^{-5}$ &  $67.26\pm0.36$/$4.02\times10^{-5}$ &  $77.78\pm0.12$/$1.09\times10^{-4}$ \\
     Continuous dropout &  $68.43\pm0.33$/$5.09\times10^{-6}$ &  $67.12\pm0.36$/$2.62\times10^{-5}$ &  $77.56\pm0.14$/$1.28\times10^{-5}$ \\
     Information dropout &  $69.40\pm0.25$/$1.69\times10^{-5}$ &  $67.69\pm0.36$/$2.68\times10^{-4}$ &  $\underline{77.85\pm0.05}$/$1.74\times10^{-4}$ \\
     Gaussian soft dropout &  $69.20\pm0.16$/$3.61\times10^{-7}$ &  $67.75\pm0.32$/$1.83\times10^{-4}$ &  $77.83\pm0.11$/$6.03\times10^{-6}$ \\
     Laplace soft dropout &  $69.46\pm0.13$/$1.46\times10^{-6}$ &  $67.8\pm0.13$/$4.93\times10^{-5}$ &  $77.74\pm0.19$/$2.94\times10^{-5}$ \\
     Advanced dropout &  $\boldsymbol{70.65\pm0.17}$/ N/A &  $\boldsymbol{68.85\pm0.24}$/ N/A &  $\boldsymbol{78.22\pm0.15}$/ N/A \\
    \bottomrule
  \end{tabular}}
  \label{tab:otherbackbones}
  \vspace{-6mm}
\end{table}

\begin{table*}[!t]
  \centering
  \caption{\footnotesize Test accuracies ($\%$) of ablation studies on MNIST, CIFAR-$10$, CIFAR-$100$,~\emph{mini}ImageNet, and Caltech-$256$ datasets. ``Advanced dropout w/o optimization'' means the advanced dropout with fixed parameters; ``Advanced dropout w/o prior'' mean the advanced dropout minus the prior. Note that the best results are marked in~\textbf{bold}, respectively.}
  \vspace{-3mm}
  \resizebox{0.8\linewidth}{!}{
    \begin{tabular}{lccccc}
    \toprule
    \multicolumn{1}{c}{Dataset} & MNIST & CIFAR-$10$ & CIFAR-$100$ & \emph{Mini}ImageNet & Caltech-$256$ \\
    \multicolumn{1}{c}{Base Model} & $784$-$2\!\times\!800$-$10$ & VGG$16$ & VGG$16$ & VGG$16$ & VGG$16$ \\
    \midrule
    No dropout & $98.23\pm0.11$ & $93.86\pm0.10$ & $73.62\pm0.38$ & $76.35\pm0.08$ & $63.87\pm0.07$ \\
    Advanced dropout w/o optimization & $98.70\pm0.04$ & $93.98\pm0.09$ & $74.55\pm0.19$ & $77.20\pm0.08$ & $64.88\pm0.11$ \\
    Advanced dropout w/o prior & $98.81\pm0.02$ & $94.20\pm0.07$ & $74.77\pm0.10$ & $77.30\pm0.08$ & $65.03\pm0.01$ \\
    Advanced dropout & $\boldsymbol{98.89\pm0.04}$ & $\boldsymbol{94.28\pm0.03}$ & $\boldsymbol{74.94\pm0.24}$ & $\boldsymbol{77.35\pm0.01}$ & $\boldsymbol{65.09\pm0.03}$ \\
    \bottomrule
    \end{tabular}
  \label{tab:ablation}}
  \vspace{-2mm}
\end{table*}

\begin{figure*}[!t]
  \centering
  \begin{subfigure}[t]{0.245\linewidth}
    \centering
    \includegraphics[width=\linewidth]{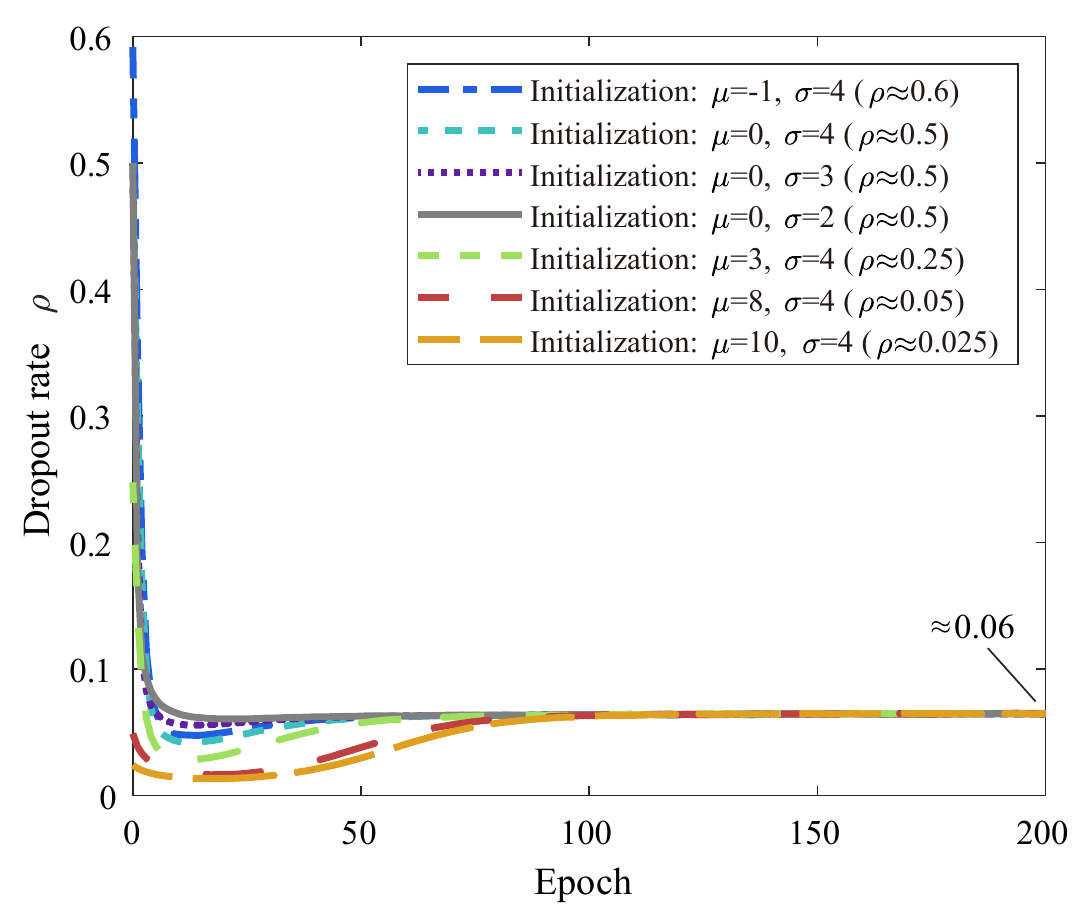}
    \vspace{-6mm}
    \subcaption{\footnotesize }
  \end{subfigure}
  \begin{subfigure}[t]{0.245\linewidth}
    \centering
    \includegraphics[width=\linewidth]{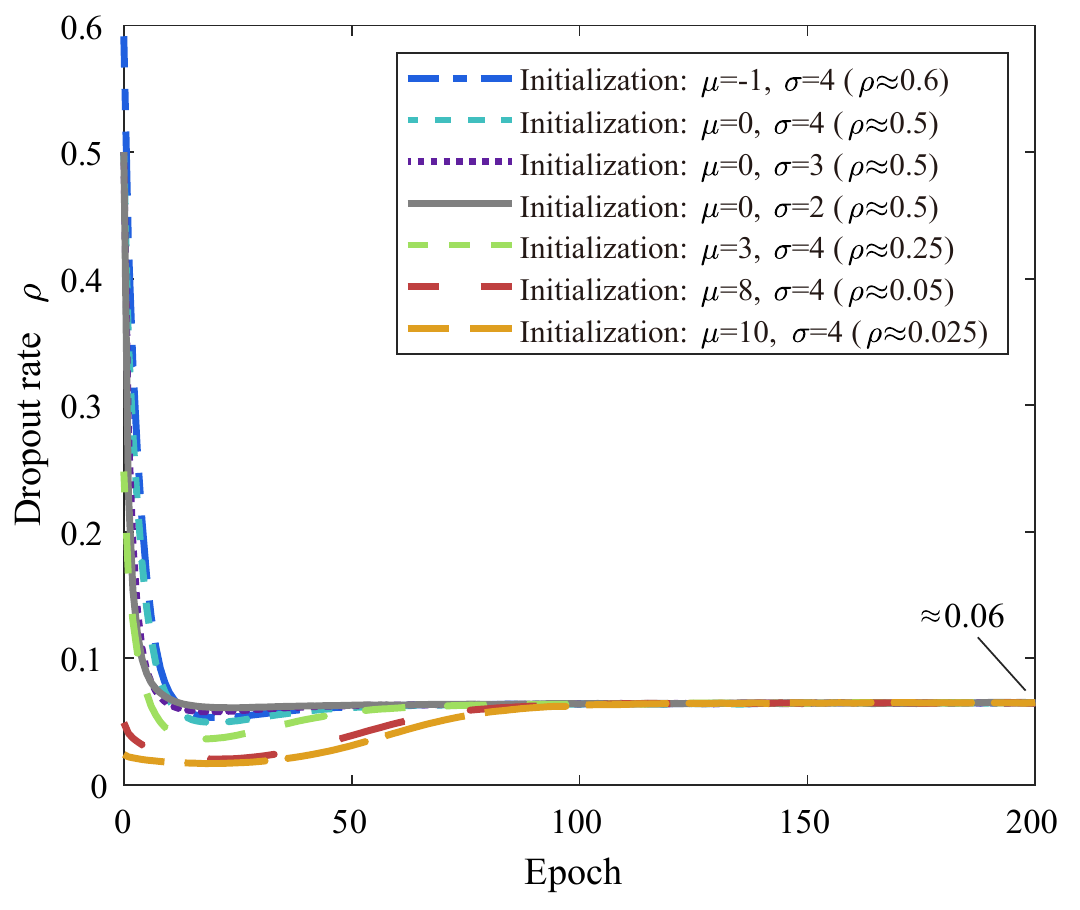}
    \vspace{-6mm}
    \subcaption{\footnotesize }
  \end{subfigure}
  \begin{subfigure}[t]{0.245\linewidth}
    \centering
    \includegraphics[width=\linewidth]{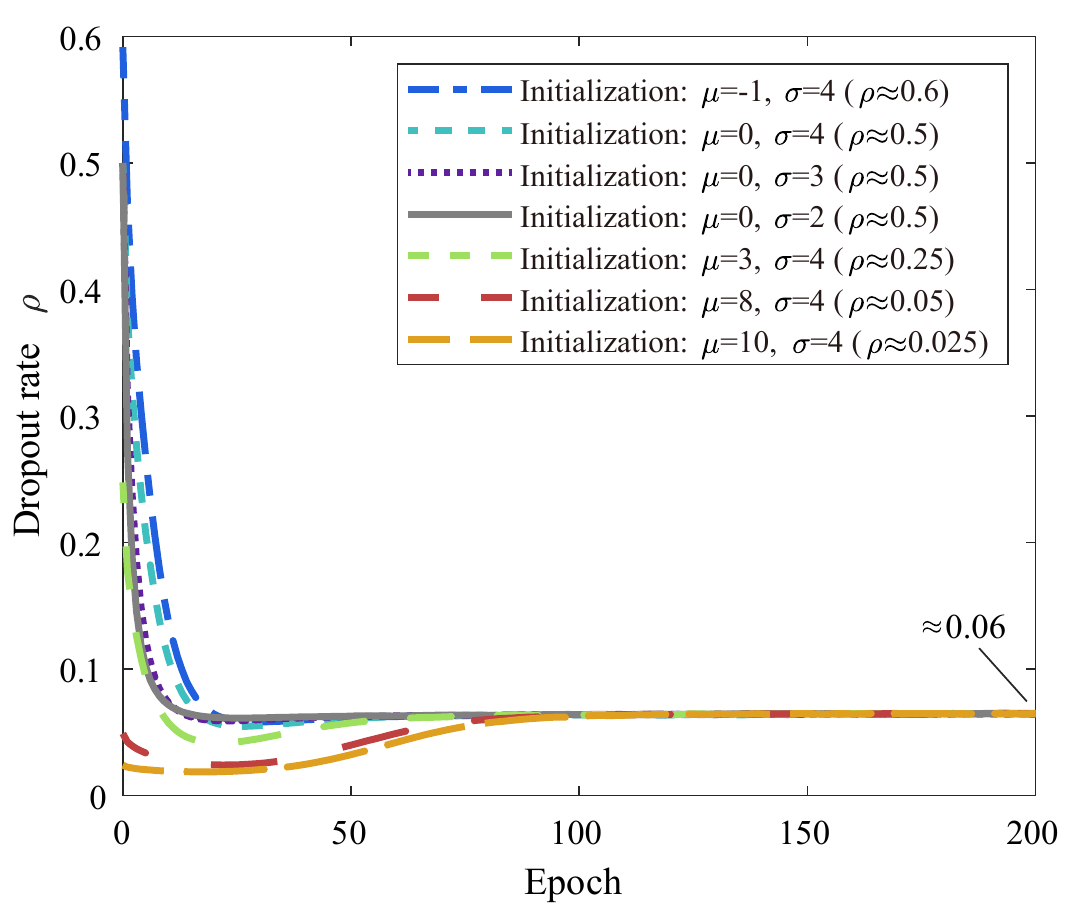}
    \vspace{-6mm}
    \subcaption{\footnotesize }
  \end{subfigure}
  \begin{subfigure}[t]{0.245\linewidth}
    \centering
    \vspace{-30mm}
    \resizebox{\linewidth}{!}{
        \begin{tabular}{lc}
            \hline
            \multicolumn{1}{c}{ Initialization}  &  Test acc. (\%) \\
            \hline
             $\mu=-1$, $\sigma=4$ ($\rho\approx0.6$) &  $98.85$ \\
             $\mu=0$, $\sigma=4$ ($\rho\approx0.5$) &  $98.89$ \\
             $\mu=0$, $\sigma=3$ ($\rho\approx0.5$) &  $98.94$ \\
             $\mu=0$, $\sigma=2$ ($\rho\approx0.5$) &  $98.87$ \\
             $\mu=3$, $\sigma=4$ ($\rho\approx0.25$) &  $98.91$ \\
             $\mu=8$, $\sigma=4$ ($\rho\approx0.05$) &  $98.90$ \\
             $\mu=10$, $\sigma=4$ ($\rho\approx0.025$) &  $98.90$ \\
            \hline
        \end{tabular}}
    \vspace{+6mm}
    \subcaption{\footnotesize }
  \end{subfigure}
  \vspace{-4mm}
  \caption{\footnotesize {Illustration of the adaptive dropout rates in the advanced dropout technique on the MNIST dataset. The subfigures illustrate the dropout rate curves of the advanced dropout of (a) the input features, (b) the first hidden layer, and (c) the second hidden layer. Note that differently initialized dropout rates $\rho$ (see legends) are set for initialization of the parameters $\mu_l$ and $\sigma_l$. Please refer to~\eqref{eq:dropoutrate} for the calculation formula of the dropout rate.} { Corresponding test accuracies (acc.) with different initializations are also reported in (d).}}\label{fig:parammnist}
  \vspace{-2mm}
\end{figure*}

\begin{figure*}[!t]
  \centering
  \begin{subfigure}[t]{0.245\linewidth}
    \centering
    \includegraphics[width=\linewidth]{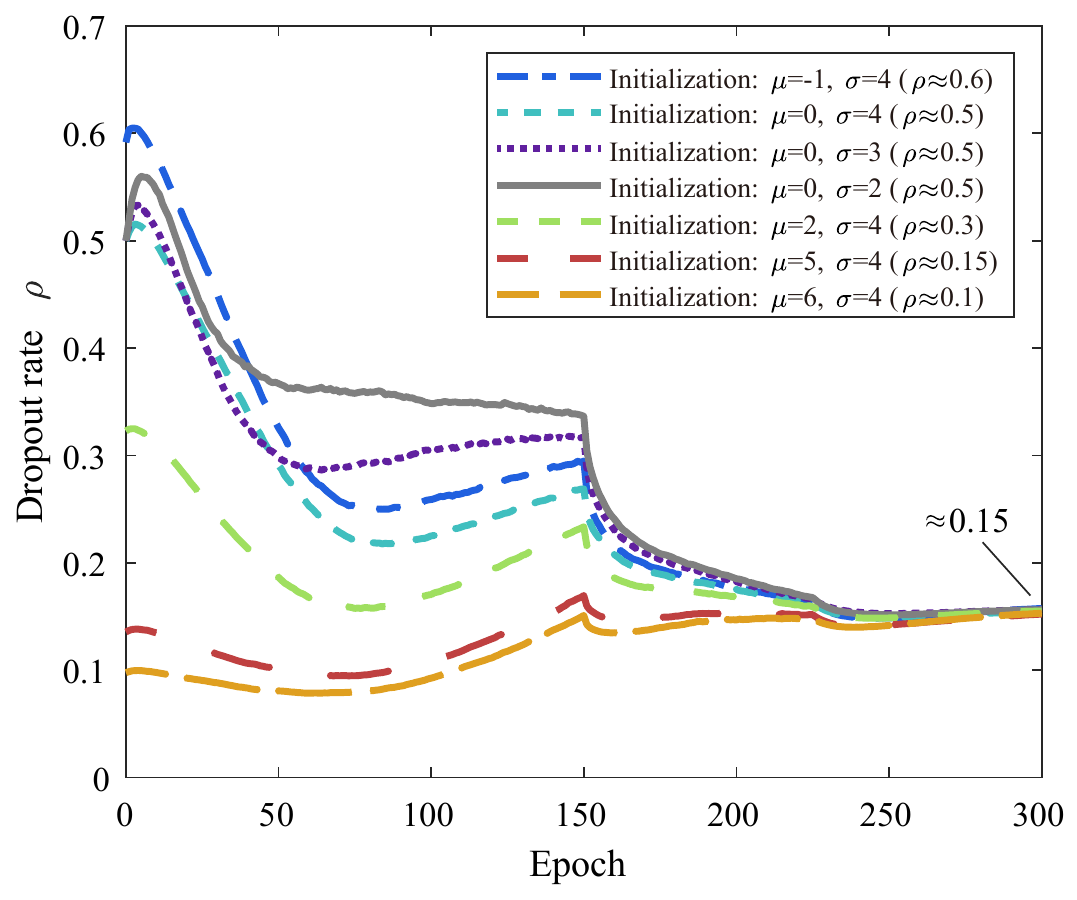}
    \vspace{-6mm}
    \subcaption{\footnotesize }
  \end{subfigure}
  \begin{subfigure}[t]{0.245\linewidth}
    \centering
    \includegraphics[width=\linewidth]{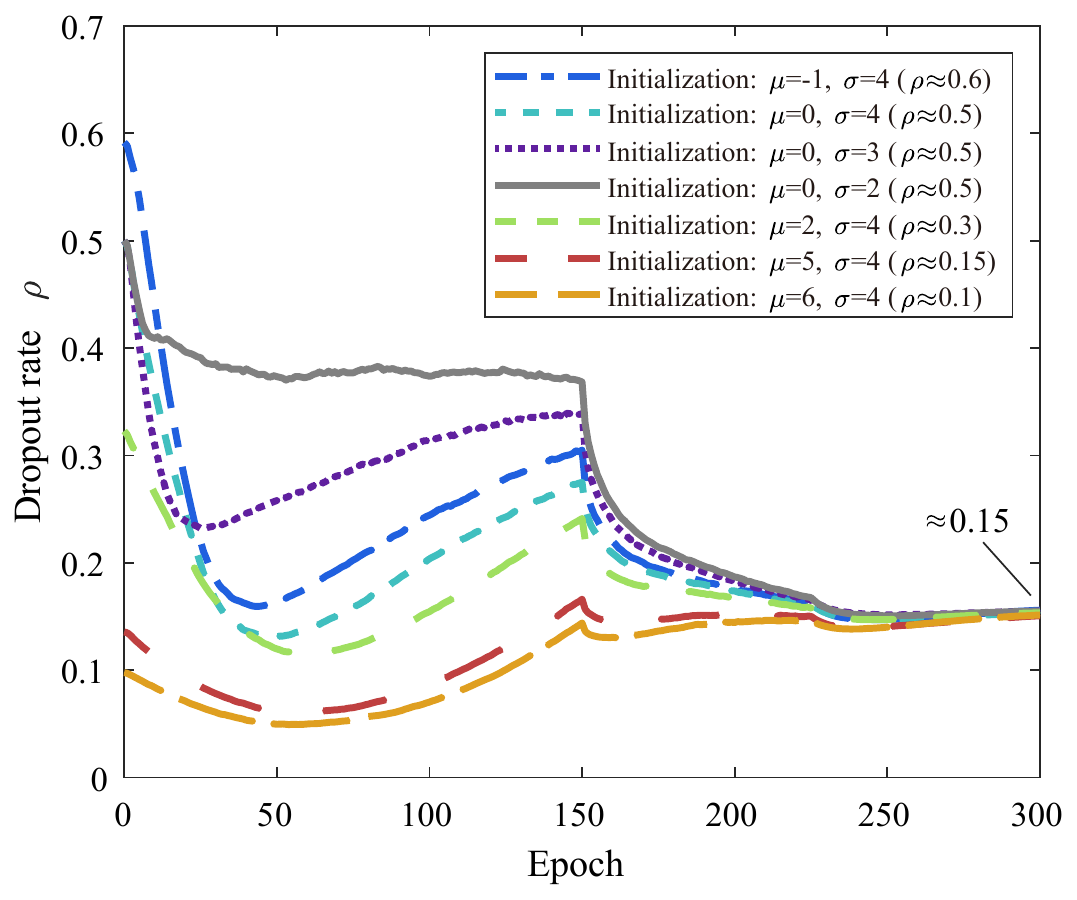}
    \vspace{-6mm}
    \subcaption{\footnotesize }
  \end{subfigure}
  \begin{subfigure}[t]{0.245\linewidth}
    \centering
    \includegraphics[width=\linewidth]{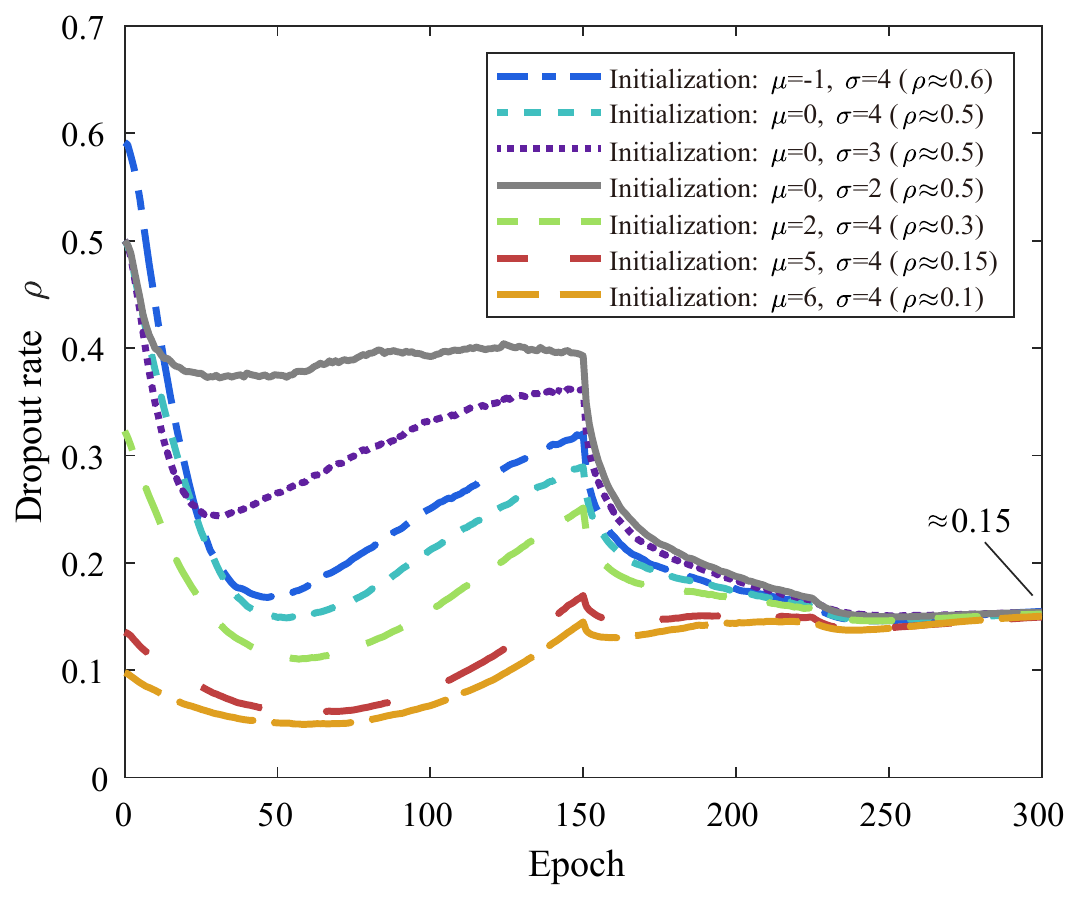}
    \vspace{-6mm}
    \subcaption{\footnotesize }
  \end{subfigure}
  \begin{subfigure}[t]{0.245\linewidth}
    \centering
    \vspace{-31mm}
    \resizebox{\linewidth}{!}{
        \begin{tabular}{lc}
            \hline
            \multicolumn{1}{c}{ Initialization} &  Test acc. (\%) \\
            \hline
             $\mu=-1$, $\sigma=4$ ($\rho\approx0.6$) &  $94.24$ \\
             $\mu=0$, $\sigma=4$ ($\rho\approx0.5$) &  $94.28$ \\
             $\mu=0$, $\sigma=3$ ($\rho\approx0.5$) &  $94.27$ \\
             $\mu=0$, $\sigma=2$ ($\rho\approx0.5$) &  $94.36$ \\
             $\mu=2$, $\sigma=4$ ($\rho\approx0.3$) &  $93.92$ \\
             $\mu=5$, $\sigma=4$ ($\rho\approx0.15$) &  $94.08$ \\
             $\mu=6$, $\sigma=4$ ($\rho\approx0.1$) &  $93.94$ \\
            \hline
        \end{tabular}}
    \vspace{+6mm}
    \subcaption{\footnotesize }
  \end{subfigure}
  \vspace{-4mm}
  \caption{\footnotesize {Illustration of the adaptive dropout rates in the advanced dropout technique on the CIFAR-$10$ dataset. The subfigures mean the dropout rate curves of the advanced dropout of (a) the convolutional features of the last convolutional layer, (b) the first hidden layer, and (c) the second hidden layer. Differently initialized dropout rates $\rho$ (see legends) are set for initialization of the parameters $\mu_l$ and $\sigma_l$.} { Corresponding test accuracies (acc.) with different initializations are also reported in (d).}}\label{fig:paramcifar}
  \vspace{-6mm}
\end{figure*}

\vspace{-3mm}
\subsection{Ablation Studies}\label{ssec:ablation}

To investigate the effectiveness of the components of the advanced dropout technique, including the parametric prior and the SGVB inference, we conducted quantitative comparisons on the MNIST~\cite{mnist98}, the CIFAR-$10$ and -$100$~\cite{krizhevsky09cifar}, the~\emph{mini}ImageNet~\cite{vinyals2016matching}, and the Caltech-$256$~\cite{griffin2006caltech} datasets. The experimental results are shown in Table~\ref{tab:ablation}. We compare the full advanced dropout technique with the techniques whose the parametric prior component is removed (``advanced dropout w/o prior'' in Table~\ref{tab:ablation}) and whose parameters $\mu_l$ and $\sigma_l$ are fixed at $0.5$ (``advanced dropout w/o optimization'' in Table~\ref{tab:ablation}). The model we applied is the two-hidden-layer FC neural network with $800$ hidden nodes in each layer on the MNIST dataset, while the VGG$16$ model is used on the other datasets as the base models.

It can be observed that, the full advanced dropout technique outperforms the base model without any dropout training by about $0.7\%$, $0.5\%$, $1.3\%$, $1\%$, and $1.2\%$ on the five datasets, respectively. When removing the parametric prior from the full technique, the classification accuracies decrease slightly on all the datasets, even though the technique performs better than the base model as well. When we continue removing the SGVB inference and setting the technique with fixed parameters $\mu_l$ and $\sigma_l$, classification accuracies further decrease. Therefore, the parametric prior and the SGVB inference play their own positive roles in the advanced dropout, which are both essential.

\subsection{Analysis of Dropout Rate Characteristics}\label{ssec:analysisofdropoutrate}

In this section, we extensively study the characteristics of the advanced dropout technique, especially its dropout rate learning process. Firstly, we analyze the convergence of the adaptive dropout rate and the learned distributions of the dropout masks. Then, to compare the advanced dropout technique with the one generating dropout rate directly (without an explicit distribution), we design a dropout variant in this kind and conduct several experiments.

\vspace{-2mm}
\subsubsection{Convergence of Adaptive Dropout Rate}\label{sssec:convergence}

In this section, we discuss the convergence of the advanced dropout technique. Due to the importance of the distribution parameters $\mu_l$ and $\sigma_l$ of each layer in DNNs, we investigate their convergence via a quantitative indicator, called dropout rate which is essential in dropout techniques. In standard dropout~\cite{hinton12}, the dropout rate is considered as one minus the parameter $p$ of the Bernoulli distribution. Given that the parameter $p$ represents the mean of the Bernoulli distribution, we can generalize the calculation of the dropout rate by one minus the mean of the distribution of the dropout masks. In particular, the dropout rate $\rho_l$ of the advanced dropout technique for all the $K_l$ nodes $m_1^{(l)},\cdots,m_{K_l}^{(l)}$ can be explicitly calculated by $\mu_l$ and $\sigma_l$ as

\begin{footnotesize}
\begin{equation}\label{eq:dropoutrate}
  \rho_l=1-\text{E}[m_j^{(l)}]\approx1-\text{Sigmoid}\left(\frac{\mu_l}{\sqrt{1+\frac{\pi}{8}\sigma_l^2}}\right).
\end{equation}
\vspace{-4mm}
\end{footnotesize}

In this case, we conducted groups of experiments and illustrate their dropout rate curves in Figure~\ref{fig:parammnist} and~\ref{fig:paramcifar}. The MNIST and the CIFAR-$10$ datasets were used for evaluations. For the MNIST dataset, a FC neural network with two hidden layers and $800$ hidden nodes of each was trained for $200$ epochs with the fixed learning rate of $0.01$, while we applied a VGG$16$ model as the base model and trained it for $200$ epochs on the CIFAR-$10$ dataset but with the dynamic learning rate same as that in Section~\ref{sssec:implement}. 

\begin{figure*}[!t]
    \centering
    \resizebox{0.9\linewidth}{!}{
    \begin{tabular}{cc}
    \multirow{2}[0]{*}{
    \begin{subfigure}[t]{0.8\linewidth}
    \vspace{-19mm}
        \centering
        \includegraphics[width=1\linewidth]{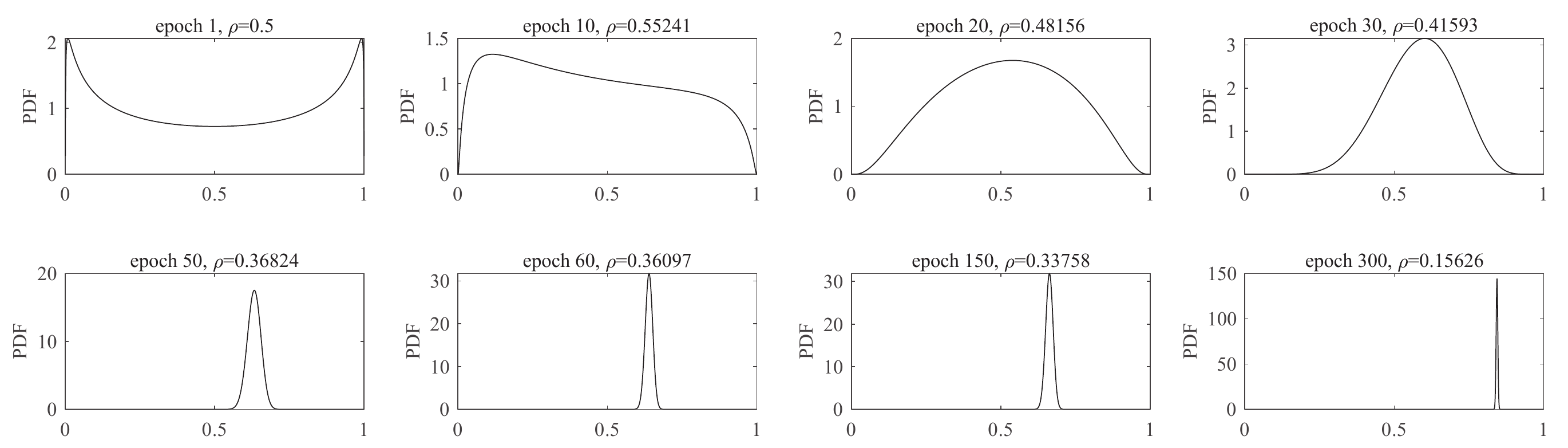}
        \vspace{-6mm}
        \subcaption{\footnotesize Learned distributions of different epochs}
    \end{subfigure}
    } & \begin{subfigure}[t]{0.18\linewidth}
        \centering
        \includegraphics[width=1\linewidth]{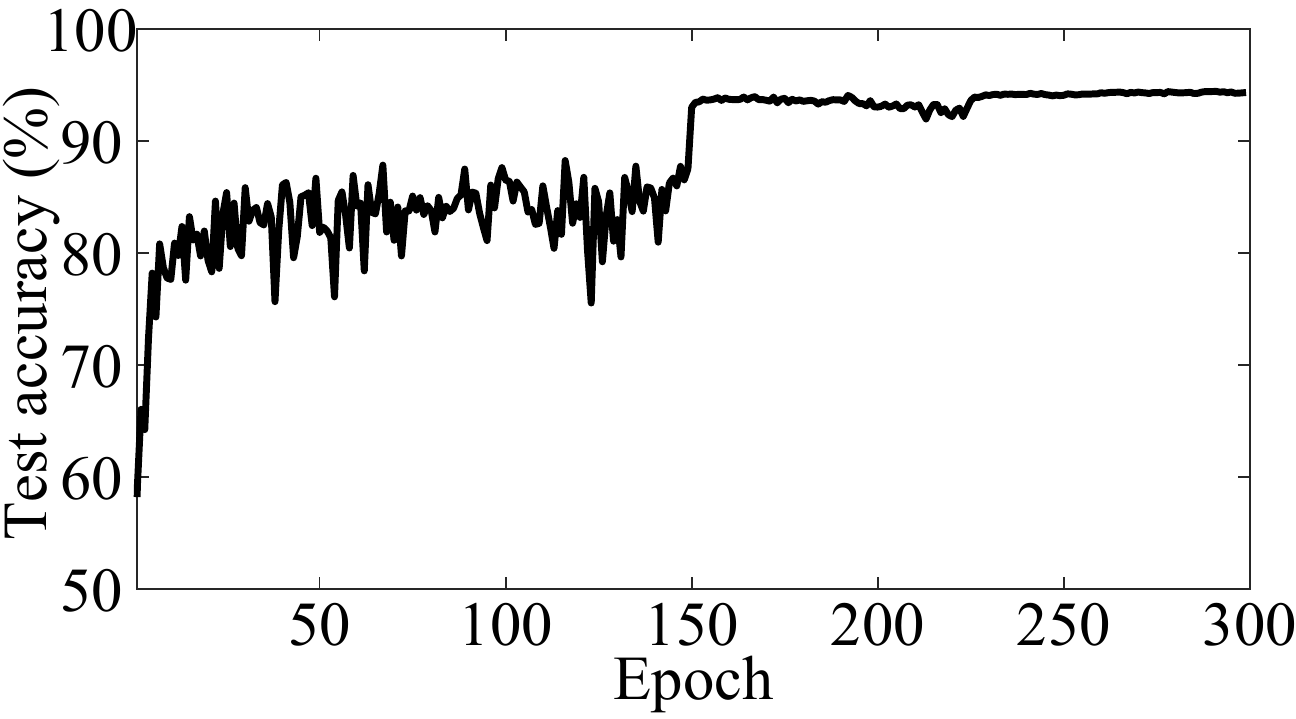}
        \vspace{-6mm}
        \subcaption{\footnotesize Test accuracies}
    \end{subfigure} \\
    & \begin{subfigure}[t]{0.18\linewidth}
        \centering
        \includegraphics[width=1\linewidth]{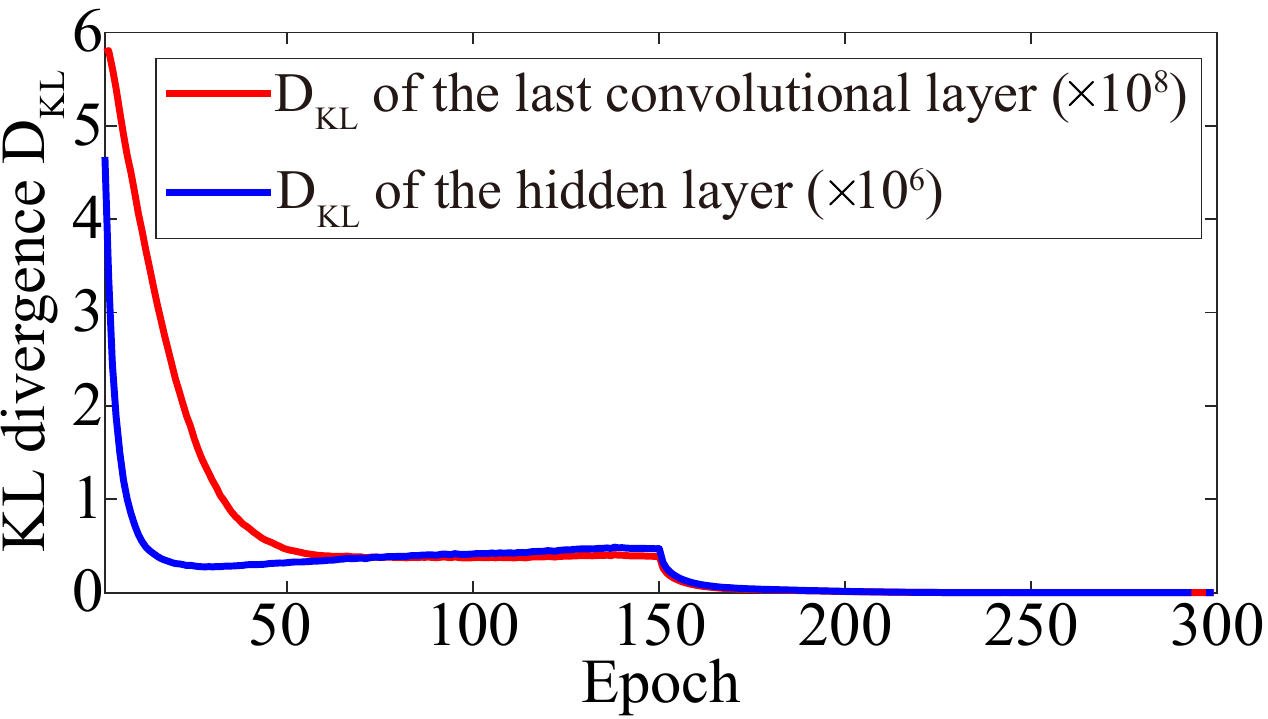}
        \vspace{-6mm}
        \subcaption{\footnotesize KL divergences}
    \end{subfigure} \\
    \end{tabular}}
    \vspace{-2mm}
    \caption{\footnotesize Learned distributions of the dropout masks with the VGG$16$ model on the CIFAR-$10$ dataset. Learned distributions of eight selected epochs are shown in (a). Note that for other datasets and base models, the distributions are similar with the shown ones. To further discuss the relationship between the convergence of the model and the distributions, we illustrate the curve of test accuracies in (b) and KL divergences from the distributions of different epochs to that of the last epoch in (c).}
    \label{fig:learneddistribution}
    \vspace{-2mm}
\end{figure*}

\begin{table*}[!t]
  \centering
  \caption{\footnotesize Comparison of the advanced dropout and a dropout variant that directly computes the dropout rate by the MLP. The test accuracies ($\%$) and the $p$-values of Student's \emph{t}-tests on CIFAR-$10$ and -$100$ datasets are shown, respectively. Note that the best results are marked in~\textbf{bold}.}
  \vspace{-1mm}
  \footnotesize
  \resizebox{0.8\linewidth}{!}{
    \begin{tabular}{lcccc}
    \toprule
    \multicolumn{1}{c}{ Dataset} & \multicolumn{2}{c}{ CIFAR-$10$} & \multicolumn{2}{c}{ CIFAR-$100$}\\
    \multicolumn{1}{c}{ Base model} &  VGG$16$/$p$-value &  ResNet$18$/$p$-value &  VGG$16$/$p$-value &  ResNet$18$/$p$-value \\
    \midrule
     Concrete+MLP &  $94.10\pm0.08$/$3.87\times10^{-3}$ &  $95.08\pm0.04$/$4.13\times10^{-5}$ &  $73.77\pm0.19$/$1.76\times10^{-5}$ &  $76.45\pm0.19$/$1.77\times10^{-5}$ \\
     Advanced dropout &  $\boldsymbol{94.28\pm0.03}$/ N/A &  $\boldsymbol{95.52\pm0.09}$/ N/A &  $\boldsymbol{74.94\pm0.24}$/ N/A &  $\boldsymbol{77.78\pm0.08}$/ N/A \\
    \bottomrule
    \end{tabular}}
  \label{tab:ablation2}
  \vspace{-6mm}
\end{table*}

\begin{figure}[!t]
  \centering
  \includegraphics[width=0.8\linewidth]{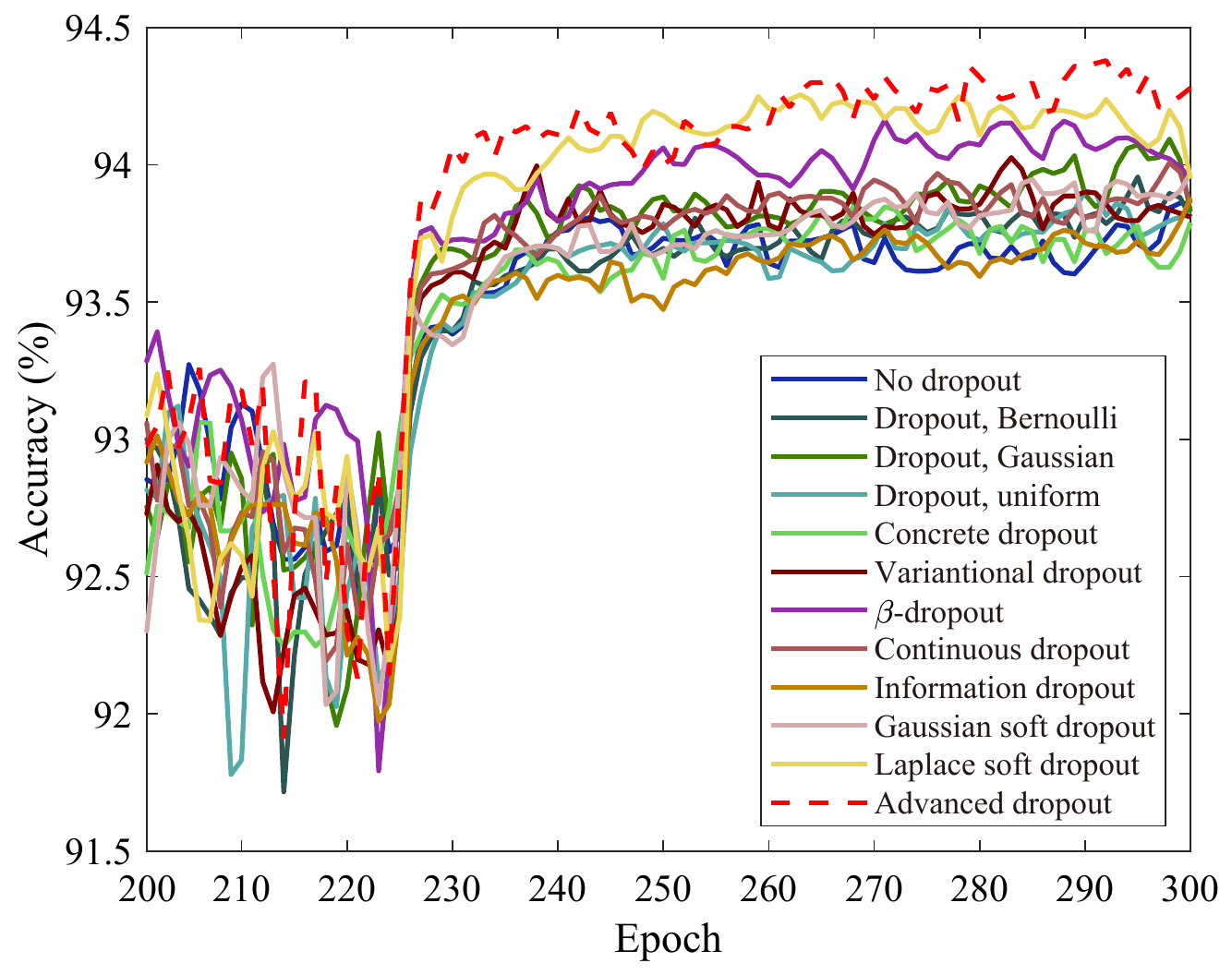}
  \vspace{-4mm}
  \caption{\footnotesize { Test accuracy curves with VGG$16$ on the CIFAR-$10$ dataset as an example. We only illustrate the last $100$ epochs for better comparisons of the accuracies of different dropout variants and the proposed advanced dropout technique.}}\label{fig:testaccepoch}
  \vspace{-6mm}
\end{figure}

{In Figure~\ref{fig:parammnist} and~\ref{fig:paramcifar}, with different initial dropout rates, it can be observed that the dropout rate of the proposed advanced dropout technique converged to similar values on the two datasets, respectively (\emph{i.e.}, $\approx 0.06$ on the MNIST dataset and $\approx 0.15$ on the CIFAR-$10$ dataset). This indicates that different initial dropout rates do not influence the optimization results and can converge to the similar last values. All the models with different initializations converge to the same values gradually after the $150^{th}$ epoch. Therefore, we can conclude that the advanced dropout technique has good convergence performance. We should note that the initial dropout rate $0.5$ (a commonly used value in~\cite{hinton12,shen2018continuous}) is an okay choice and the other values are also suitable. Meanwhile, the dropout rates for each layer are optimized and vary during training, which means that different epochs need different values of dropout rate. The low dropout rates at the last epochs are just suitable for the last epochs only.}

An interesting phenomenon is observed in Figure~\ref{fig:paramcifar}. In the different layers, the dropout rates with distinct initializations fall immediately until reaching a local minimum of the dropout rate curves and then increase continuously. Before reaching the local minimum, the model fits the data rapidly and the dropout rates of different layers are adaptively reduced to a low level in a quick way for fully optimizing model parameters. Then after the local minimum, failing of convergence of the optimization (because of too large learning rate) occurs and a smaller learning rate is required to decrease optimization steps (\emph{i.e.}, the learning rate multiplying gradients in SGD). As the decrease of the optimization steps can be also led to by multiplying smaller dropout masks onto the outputs of each layer (\emph{i.e.}, increasing the dropout rates), the dropout rates are driven to rise automatically. In addition, at the $150^{th}$ and the $225^{th}$ epochs, the dropout rates fall off again due to the decay of the learning rate.

{In addition, different initializations of $\mu$ and $\sigma$ only slightly affect the test accuracies of the models, as demonstrated in Figure~\ref{fig:parammnist}(d) and~\ref{fig:paramcifar}(d).}

\vspace{-3mm}
\subsubsection{Learned Distributions of the Dropout Masks}\label{sssec:learneddistribution}

We discuss the shapes of the learned  distributions of the dropout masks, as shown in Figure \ref{fig:learneddistribution}(a). We only illustrate the distributions optimized with the VGG$16$ model on the CIFAR-$10$ dataset, since the distributions for other datasets and base models are similar with the shown ones. As shown in Figure~\ref{fig:learneddistribution}(a), the distributions change dramatically between epoch $1$ and epoch $60$, and then converged till the end of the training process. Based on the view of Bayesian inference, the variance of an estimated posterior distribution will converge to a small value and the estimated posterior distribution will be concentrated in a compact area, if the data amount is large enough~\cite{bishop06,ma2011bayesian}.

To further discuss the relationship between the convergence of the model and the distributions, we illustrate the curve of test accuracies in Figure~\ref{fig:learneddistribution}(b) and the curve of the KL divergences from the distributions on different epochs to that of the last epoch in Figure~\ref{fig:learneddistribution}(c). The KL divergences decrease significantly at the beginning of training and then converge. During the whole training process, the KL divergences continuously decrease until the last epoch as the test accuracies increase. This means that the distributions are optimized gradually with model training.

\begin{figure*}[!t]
  \centering
  \includegraphics[width=0.95\linewidth]{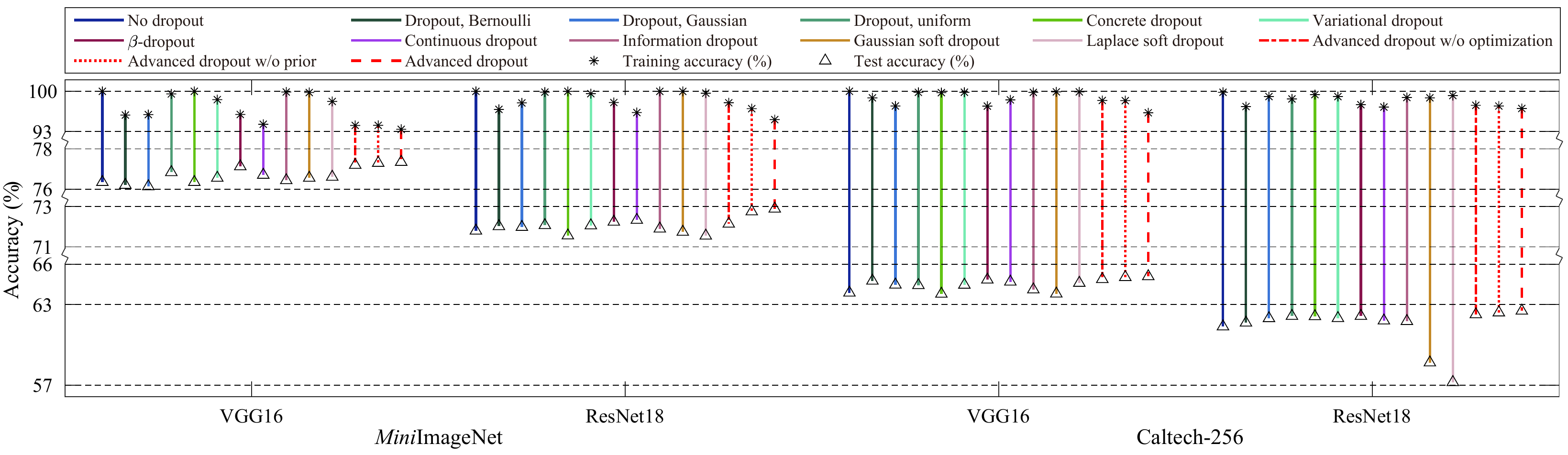}
  \vspace{-4mm}
  \caption{\footnotesize {Training and test accuracy comparisons of the last epoch with VGG$16$ and ResNet$18$ models on~\emph{mini}ImageNet and Caltech-$256$ datasets, respectively. The vertical lines in red are for the proposed advanced dropout, the advanced dropout w/o prior, and the advanced dropout w/o optimization, respectively. It can be observed that all the essential components in the proposed advanced dropout contribute to overfitting prevention. The other solid lines (with different colors) are for the referred methods.}}\label{fig:trainingacc}
  \vspace{-2mm}
\end{figure*}

\begin{table*}[!t]
  \centering
  \caption{\footnotesize Top-$1$ and top-$5$ test accuracies (acc., $\%$) with different training set sizes on Caltech-$256$ dataset. $\xi$ means the training sample number per class. Note that the best results are marked in~\textbf{bold}, respectively.}
  \vspace{-3mm}
  \resizebox{0.8\linewidth}{!}{
    \begin{tabular}{lcccccc}
    \toprule
    \multicolumn{1}{c}{\multirow{2}[0]{*}{Model}} & \multicolumn{2}{c}{$\xi=60$} & \multicolumn{2}{c}{$\xi=30$} & \multicolumn{2}{c}{$\xi=15$} \\
          & Top-$1$ acc. & Top-$5$ acc. & Top-$1$ acc. & Top-$5$ acc. & Top-$1$ acc. & Top-$5$ acc. \\
    \midrule
    No dropout & $63.87\pm0.07$ & $80.41\pm0.07$ & $51.12\pm0.11$ & $70.77\pm0.08$ & $36.19\pm0.10$ & $55.66\pm0.15$ \\
    Dropout, Bernoulli & $64.76\pm0.14$ & $81.34\pm0.14$ & $51.88\pm0.14$ & $71.21\pm0.14$ & $36.31\pm0.27$ & $56.17\pm0.24$ \\
    Dropout, Gaussian & $64.47\pm0.22$ & $81.05\pm0.19$ & $51.97\pm0.14$ & $71.36\pm0.13$ & $36.35\pm0.10$ & $56.21\pm0.12$ \\
    Dropout, uniform & $64.43\pm0.10$ & $81.01\pm0.10$ & $51.78\pm0.23$ & $71.11\pm0.11$ & $35.82\pm0.07$ & $56.02\pm0.06$ \\
    Concrete dropout & $63.79\pm0.07$ & $80.37\pm0.07$ & $50.72\pm0.21$ & $70.55\pm0.11$ & $35.08\pm0.10$ & $55.30\pm0.09$ \\
    Variational dropout & $64.46\pm0.11$ & $81.04\pm0.04$ & $52.08\pm0.08$ & $70.91\pm0.14$ & $36.42\pm0.14$ & $56.14\pm0.26$ \\
    $\beta$-dropout & $64.84\pm0.07$ & $81.07\pm0.09$ & $52.13\pm0.13$ & $71.33\pm0.04$ & $36.33\pm0.04$ & $56.28\pm0.14$ \\
    Continuous dropout & $64.69\pm0.09$ & $81.06\pm0.09$ & $52.12\pm0.18$ & $71.31\pm0.19$ & $36.24\pm0.22$ & $56.16\pm0.22$ \\
    Information dropout & $64.11\pm0.13$ & $81.55\pm0.06$ & $51.61\pm0.11$ & $70.92\pm0.17$ & $35.83\pm0.13$ & $55.97\pm0.09$ \\
    Gaussian soft dropout & $63.80\pm0.10$ & $80.38\pm0.09$ & $51.01\pm0.12$ & $70.73\pm0.14$ & $34.65\pm0.22$ & $55.26\pm0.15$ \\
    Laplace soft dropout & $64.60\pm0.05$ & $81.18\pm0.04$ & $51.61\pm0.14$ & $70.80\pm0.06$ & $35.49\pm0.21$ & $55.09\pm0.18$ \\
    Advanced dropout & $\boldsymbol{65.09\pm0.03}$ & $\boldsymbol{81.64\pm0.04}$ & $\boldsymbol{52.57\pm0.05}$ & $\boldsymbol{71.62\pm0.13}$ & $\boldsymbol{36.95\pm0.12}$ & $\boldsymbol{56.58\pm0.16}$ \\
    \bottomrule
    \end{tabular}}
  \label{tab:trainnum}
  \vspace{-4mm}
\end{table*}

\begin{table}[!t]
  \scriptsize
  \centering
  \caption{\footnotesize Test accuracies ($\%$) of fully connected (FC) neural networks with different depth on MNIST dataset. The model structures mean ``hidden layer number $\times$ hidden node number per layer''. Note that the best results are marked in~\textbf{bold}, respectively.}
  \vspace{-3mm}
  \resizebox{\linewidth}{!}{
    \begin{tabular}{lccc}
    \toprule
    \multicolumn{1}{c}{Model} & FC ($2\!\times\!800$) & FC ($4\!\times\!800$) & FC ($8\!\times\!800$) \\
    \midrule
    No dropout & $98.23\pm0.11$ & $98.16\pm0.30$ & $98.06\pm0.47$ \\
    Dropout, Bernoulli & $98.46\pm0.06$ & $98.39\pm0.10$ & $98.18\pm0.03$ \\
    Dropout, Gaussian & $98.45\pm0.05$ & $98.41\pm0.15$ & $98.05\pm0.11$ \\
    Dropout, uniform & $98.50\pm0.12$ & $98.68\pm0.06$ & $98.40\pm0.16$ \\
    Concrete dropout & $98.45\pm0.04$ & $98.54\pm0.03$ & $98.16\pm0.21$ \\
    Variational dropout & $98.46\pm0.14$ & $98.34\pm0.17$ & $98.20\pm0.12$ \\
    $\beta$-dropout & $98.62\pm0.09$ & $98.68\pm0.06$ & $98.46\pm0.07$ \\
    Continuous dropout & $98.45\pm0.20$ & $98.42\pm0.23$ & $98.28\pm0.07$ \\
    Information dropout & $98.22\pm0.25$ & $98.18\pm0.22$ & $97.65\pm0.14$ \\
    Gaussian soft dropout & $98.64\pm0.04$ & $98.63\pm0.05$ & $98.67\pm0.05$ \\
    Laplace soft dropout & $98.70\pm0.10$ & $98.53\pm0.04$ & $98.68\pm0.06$ \\
    Advanced dropout & $\boldsymbol{98.89\pm0.04}$ & $\boldsymbol{98.99\pm0.04}$ & $\boldsymbol{98.86\pm0.06}$ \\
    \bottomrule
    \end{tabular}}
  \label{tab:depth}
  \vspace{-6mm}
\end{table}

\vspace{-2mm}
\subsubsection{Comparison with Dropout Rate Generation w/o Explicit Distribution}\label{sssec:compdrwoexpldis}

In this section, we compare the proposed advanced dropout (with an explicit distribution assumption) with the cases that compute the dropout rates directly (without explicit distribution assumption). To this end, we design a dropout variant with the concrete distribution~\cite{maddison2017the}, in which the parameter $p$,~\emph{i.e.}, one minus the dropout rate, is directly computed by an MLP. We name the designed dropout variant as ``concrete+MLP''. Table~\ref{tab:ablation2} lists the experimental results and the $p$-values of the Student's~\emph{t}-tests on the CIFAR-$10$ and -$100$ datasets, respectively. The advanced dropout achieves statistically significantly better performance than the ``concrete+MLP'' method on two datasets.

\vspace{-4mm}
\subsection{Capability of Overfitting Prevention}\label{ssec:overfitting}

{We illustrate the test accuracy curves of the last $100$ epochs with VGG$16$ on the CIFAR-$10$ dataset as an example in Figure~\ref{fig:testaccepoch}. For the proposed advanced dropout technique, an increase trend can be clearly found and the curve varies slightly, although the best accuracy may be obtained in earlier epochs. Meanwhile, for some referred techniques, for example, the Laplace soft dropout, overfitting can be found between the $280^{\text{th}}$ and $300^{\text{th}}$ epochs. This means that the proposed advanced dropout technique can actually reduce overfitting. Similar results can be also found with other base models and on other datasets.}

{One way for overfitting prevention is to trade off the bias and the variance of a model to achieve small gap between training and test performance,~\emph{i.e.}, improved generalization. Here, we show the training and the corresponding test accuracies of the last epoch when applying different dropout techniques to the base models in Figure~\ref{fig:trainingacc}. It can be observed that the proposed advanced dropout technique reduces the gap between training and test accuracies (\emph{i.e.}, reducing overfitting) and achieves the highest test accuracies at the same time. The models after removing the essential components (as discussed in Section~\ref{ssec:ablation}) are still able to reduce the training-test accuracy gap compared with the referred methods, although they obtain suboptimal results than that with the ``full'' advanced dropout technique. This means all the essential components contribute to overfitting prevention.}

To quantitatively investigate the capability of overfitting prevention of the advanced dropout technique, we designed two groups of experiments. Given the actual situation that overfitting occurs when the model size is too huge or the training dataset is too small, two groups of experiments are as follows: one exponentially decreases the training sample number in each class on the Caltech-$256$ dataset and the other exponentially increases the hidden layer number in a FC neural network on the MNIST dataset. The experimental results are shown in Table~\ref{tab:trainnum} and~\ref{tab:depth}.

\begin{figure*}[!t]
  \centering
  \includegraphics[width=0.9\linewidth]{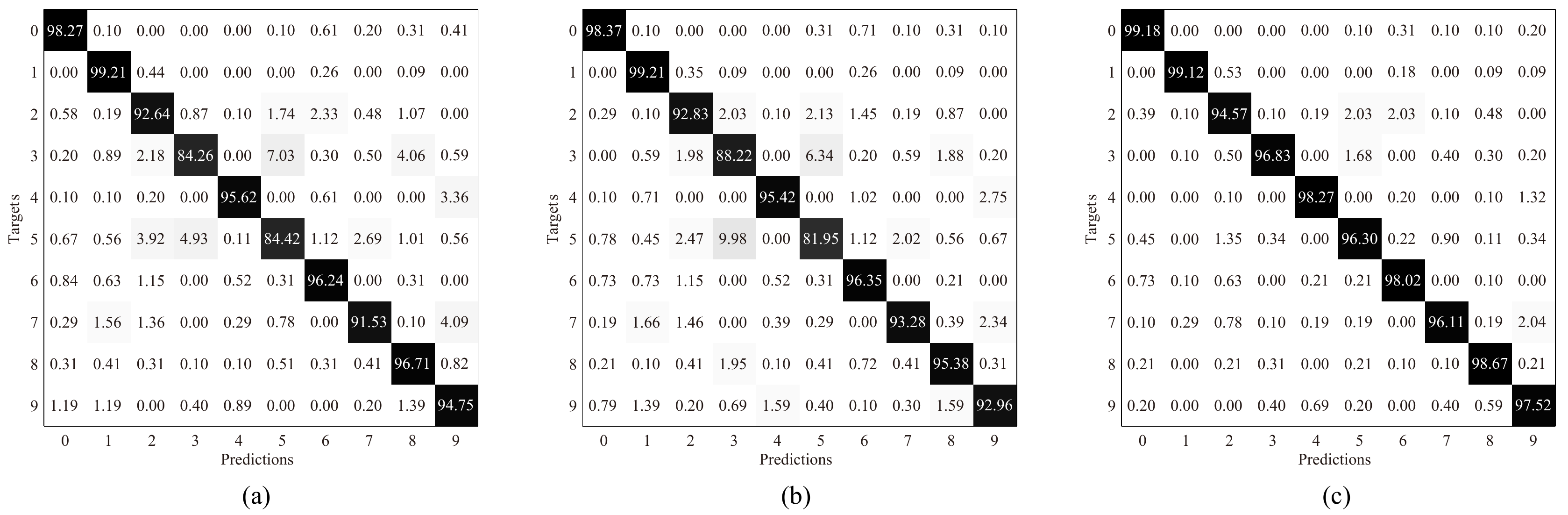}
  \vspace{-4mm}
  \caption{\footnotesize Confusion matrices of different techniques for model uncertainty inference on MNIST dataset: (a) MC dropout~\cite{gal16mcdropout}, (b) ApDeepSense~\cite{yao2018apdeepsense}, and (c) advanced dropout.}\label{fig:confusion}
  \vspace{-6mm}
\end{figure*}

On the Caltech-$256$ dataset, a coefficient $\xi$ is introduced to represent the training sample number per class and measure the size of the training set. In the experiments, $\xi$ is set as $\{60,30,15\}$ with the test set fixed as the remaining part of the whole dataset by removing $60$ samples from each class ($\xi=60$). Moreover, we applied the VGG$16$ model as the base model for each case. The top-$1$ and the top-$5$ test accuracies are reported in Table~\ref{tab:trainnum}. With the decrease of $\xi$, the classification accuracies of all the dropout techniques drop dramatically, which is expected. However, the advanced dropout technique achieves the best performance on the three cases with different $\xi$. When $\xi$ is equal to $60$, it obtains the top-$1$ and the top-$5$ accuracies of $65.09\%$ and $81.64\%$, outperforming the second best technique by about $0.2\%$ and $0.1\%$, respectively. At the mean time, it achieves the top-$1$ and top-$5$ accuracies of $52.57\%$ and $71.62\%$, which are $0.44\%$ and $0.28\%$ larger than the second best technique, respectively, with the training set reduced by half ($\xi=30$). Furthermore, For $\xi=15$, it surpasses the second best technique by $0.53\%$ and $0.30\%$ in terms of classification accuracy, where the advantage is clearer than the other two cases. It can be obviously found that, with the shrinkage of the training set, the advanced dropout technique maintains the first place on classification accuracy and enlarges the gap with the corresponding second best technique.

\begin{table}[!t]
  \centering
  \caption{\footnotesize AUROC of max probability (Max.P) and entropy (Ent.), and test accuracies (acc., $\%$) of model certainty inference on the MNIST dataset. Note that the best results are marked in~\textbf{bold}.}
  \vspace{-3mm}
    \begin{tabular}{lccc}
    \toprule
    \multicolumn{1}{c}{\multirow{2}[0]{*}{Method}} & \multicolumn{2}{c}{AUROC} & \multirow{2}[0]{*}{Acc.} \\
          & Max.P & Ent.  &  \\
    \midrule
    MC dropout & $0.9210$  & $0.8892$  & $93.49$ \\
    ApDeepSense & $0.9185$  & $0.8951$  & $93.56$ \\
    Advanced dropout & $\boldsymbol{0.9552}$  & $\boldsymbol{0.9357}$  & $\boldsymbol{97.47}$ \\
    \bottomrule
    \end{tabular}
  \label{tab:uncertainty}
  \vspace{-6mm}
\end{table}

In Table~\ref{tab:depth}, the numbers of hidden layers of the FC neural networks are set as $2$, $4$, and $8$. With the increase of the model depth, most of the referred techniques, except for the soft dropouts, performs worse and the maximum drop in classification accuracy is more than $0.5\%$. Meanwhile, the performance of the base model without any dropout techniques also decreases. However, the advanced dropout technique maintains its classification accuracies of more than $98.8\%$. In addition, it achieves the best performance among all techniques in each model depth.

{In summary, the proposed advanced dropout technique reduces the gap between training and test accuracies (\emph{i.e.}, reducing overfitting) and achieves the highest test accuracies at the same time.} Moreover, the advanced dropout technique performs best among the referred methods, when we increase the model depth or decrease the training sample number. It also performs well in the extreme case, for example, $\xi=15$ or the eight-hidden-layer FC neural network. The experimental results verify the superior capability of overfitting prevention of the advanced dropout technique.

\vspace{-4mm}
\section{Extension of Applications of Advanced Dropout}\label{sec:applications}
\vspace{-1mm}

In this section, we will generalize the advanced dropout as a practical technique in four other application areas, rather than merely a regularization technique for DNN training in image classification. Hence, we apply the advanced dropout technique in uncertainty inference and network pruning in computer vision, text classification, and regression.

\begin{figure}[!t]
    \centering
    \begin{subfigure}[t]{0.8\linewidth}
        \centering
        \includegraphics[width=1\linewidth]{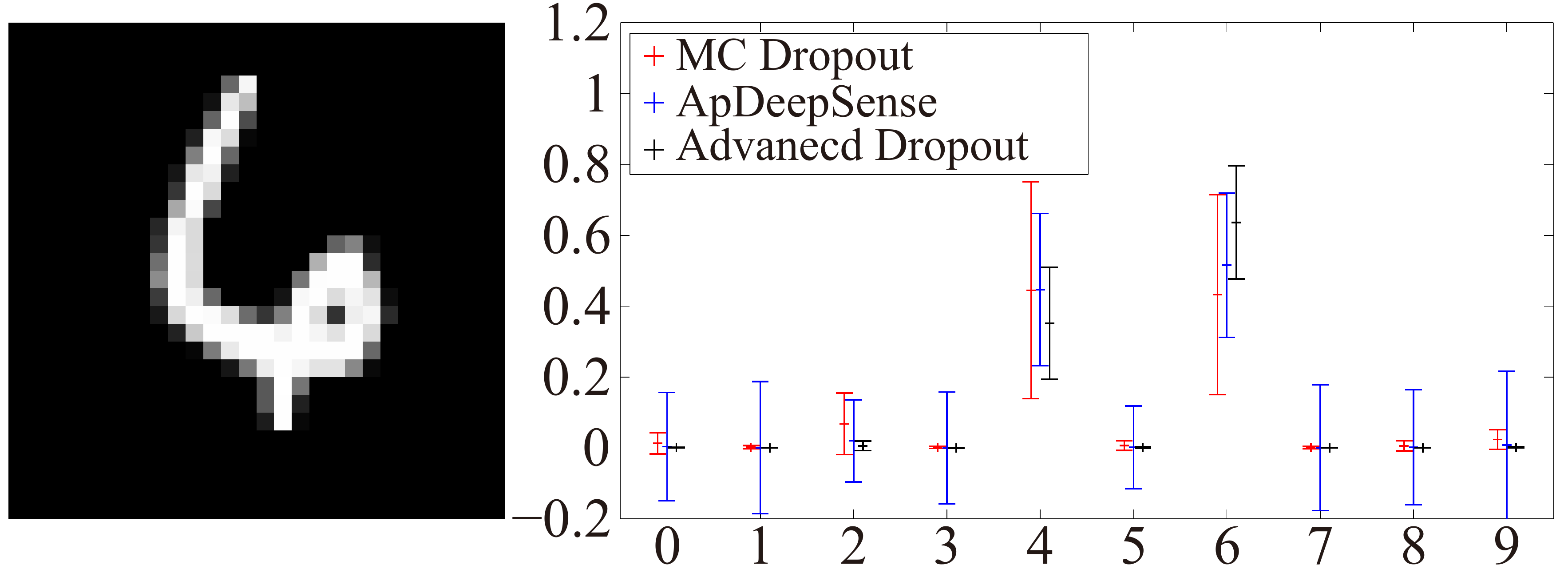}
        \vspace{-6mm}
        \subcaption{\footnotesize }
    \end{subfigure}
    \begin{subfigure}[t]{0.8\linewidth}
        \centering
        \includegraphics[width=1\linewidth]{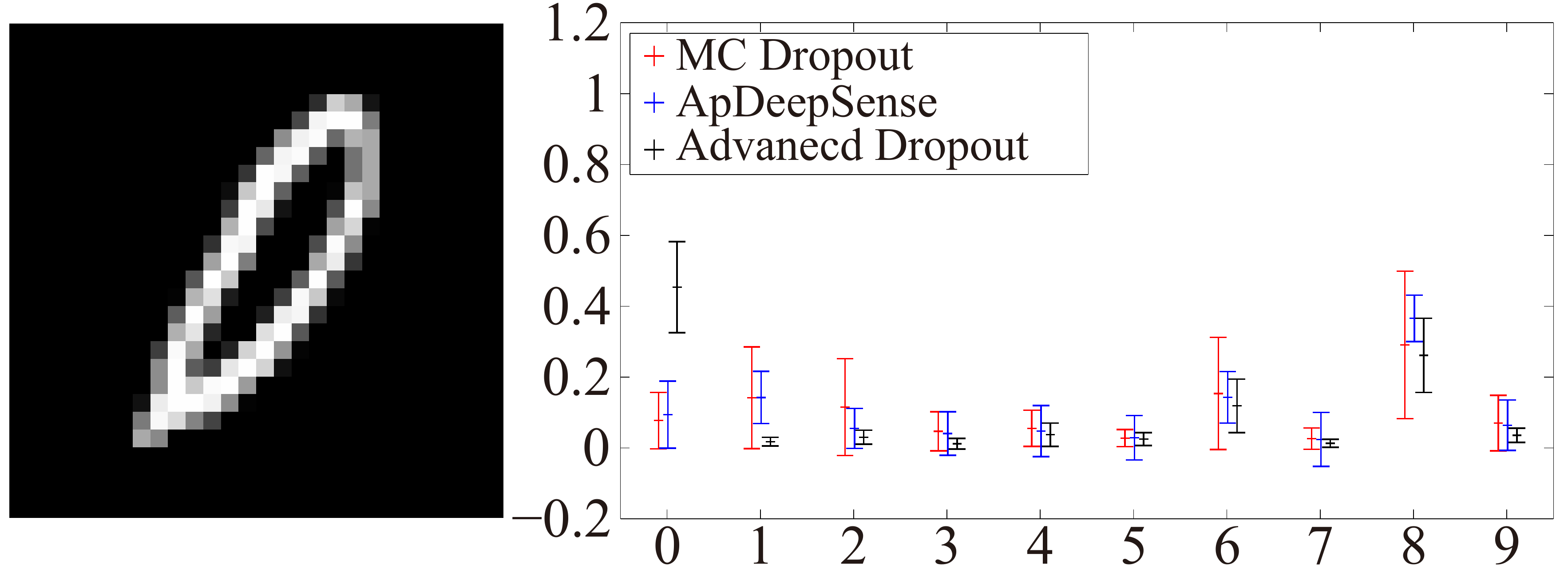}
        \vspace{-6mm}
        \subcaption{\footnotesize }
    \end{subfigure}
    \begin{subfigure}[t]{0.8\linewidth}
        \centering
        \includegraphics[width=1\linewidth]{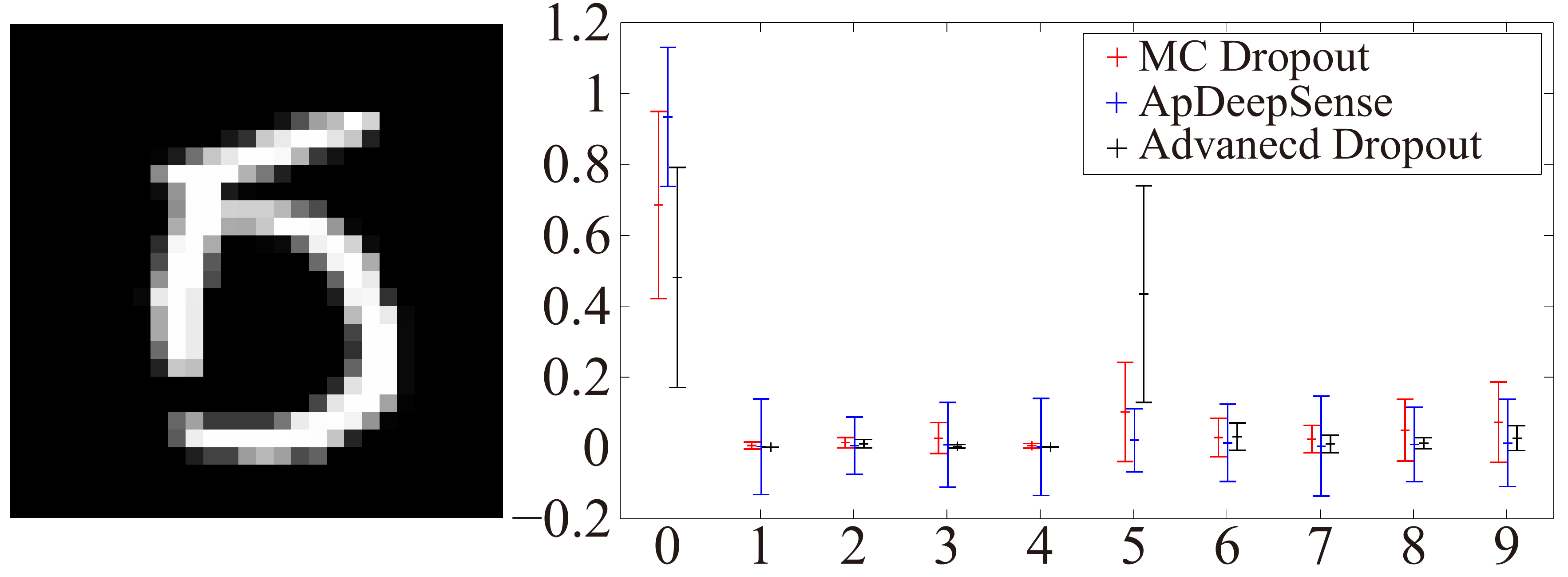}
        \vspace{-6mm}
        \subcaption{\footnotesize }
    \end{subfigure}
    \vspace{-4mm}
    \caption{\footnotesize Examples of uncertainty inference on MNIST dataset. Each column in the subfigures means a class. (a) A sample $6$ which is easily confused with $4$; (b) a sample $0$ which is easily considered as a part of $8$; and (c) a sample $5$ which is predicted as $0$ by all the techniques.}
    \label{fig:uncertainty}
    \vspace{-6mm}
\end{figure}

\begin{figure*}[!t]
    \centering
    \begin{subfigure}[t]{0.35\linewidth}
        \centering
        \includegraphics[width=1\linewidth]{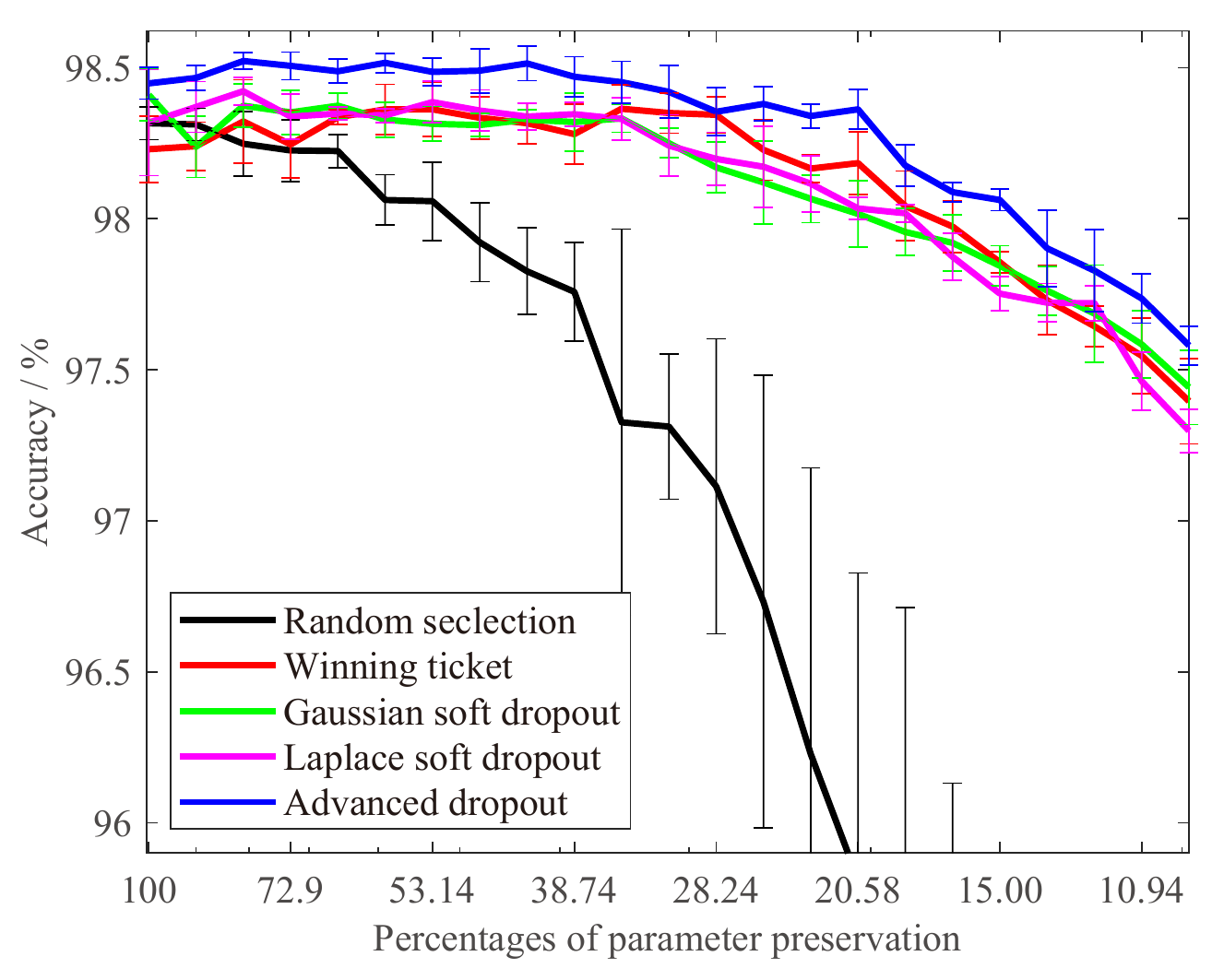}
        \vspace{-6mm}
        \subcaption{\footnotesize  Node pruning}
    \end{subfigure}
    \begin{subfigure}[t]{0.35\linewidth}
        \centering
        \includegraphics[width=1\linewidth]{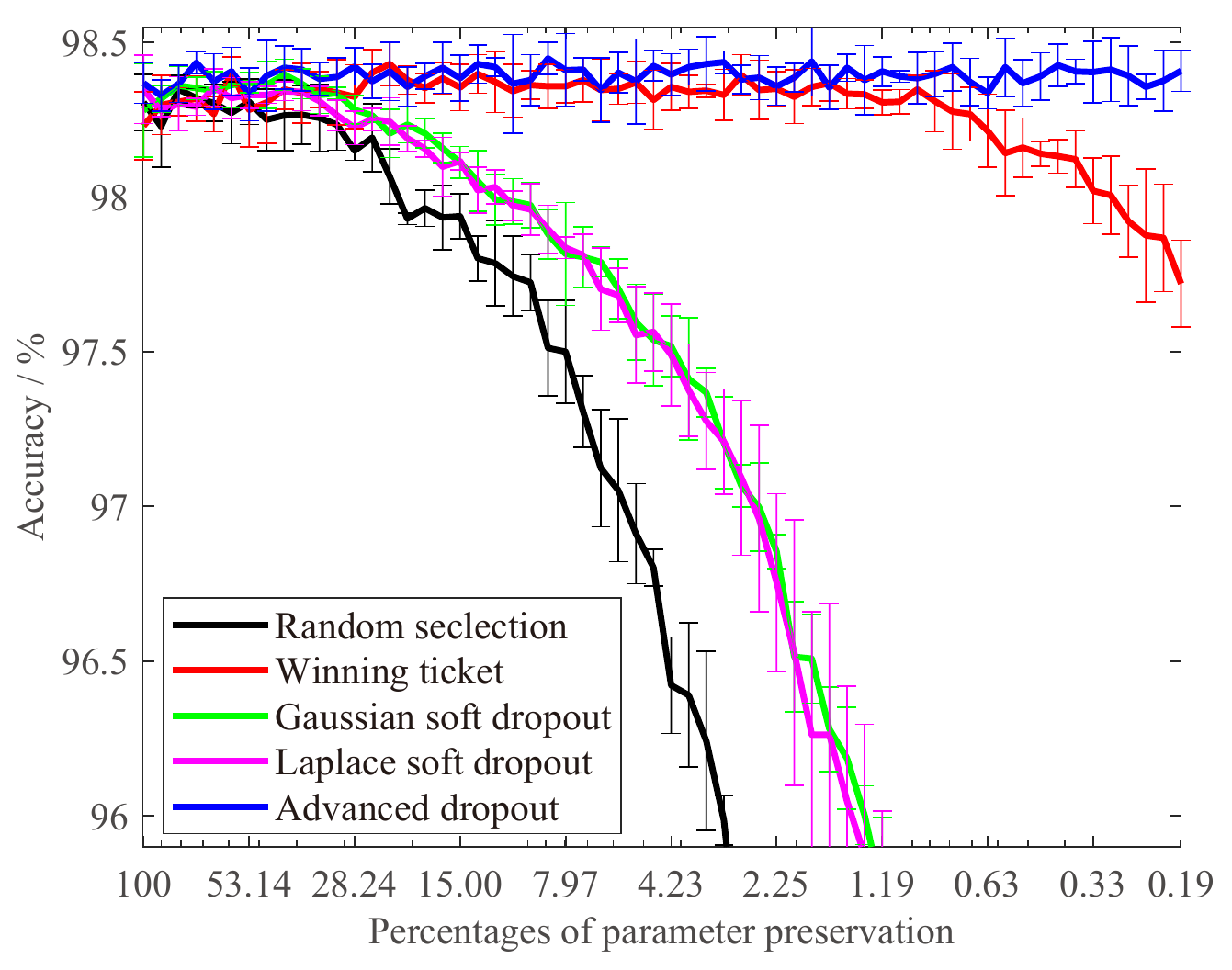}
        \vspace{-6mm}
        \subcaption{\footnotesize  Parameter pruning}
    \end{subfigure}
    \vspace{-3mm}
    \caption{\footnotesize The performance of the advanced dropout technique for (a) node pruning and (b) parameter pruning compared with the random selection, the winning ticket~\cite{frankle2019the}, and Gaussian soft dropout and Laplace soft dropout~\cite{xie2019soft} on the MNIST dataset. Note that the percentages of parameter preservation are displayed on a logarithmic scale.}
    \label{fig:pruning}
    \vspace{-4mm}
\end{figure*}

\vspace{-2mm}
\subsection{Uncertainty Inference}\label{ssec:uncertainty}

Although deep learning has attracted significant attention on research in various fields, such as classification and regression, standard deep learning tools including CNNs and RNNs are not able to capture model uncertainty and confidence through the point estimation in model training by gradient descent-based algorithms~\cite{gal16mcdropout}. Therefore, we need effective methods, for example, Bayesian learning-based methods, to estimate model uncertainty. In recent years, various techniques have been proposed for this purpose, such as MC dropout~\cite{gal16mcdropout} and ApDeepSense~\cite{yao2018apdeepsense}.

In this paper, we propose to implement the advanced dropout technique in uncertainty inference to validate its effectiveness for this propose.

Here, we modify the advanced dropout as a Monte Carlo (MC) dropout technique, perform moment matching in the test process, and estimate the mean $\text{E}[\boldsymbol{y}^*|\boldsymbol{x}^*]$ and the variance $\text{Var}[\boldsymbol{y}^*|\boldsymbol{x}^*]$ of the test output $\boldsymbol{y}^*$ given the corresponding test sample $\boldsymbol{x}^*$ as

\begin{footnotesize}
\begin{align}\label{eq:mc}
  \text{E}[\boldsymbol{y}^*|\boldsymbol{x}^*]&=\int\boldsymbol{y}^*\int q_{\boldsymbol{\Phi}}(\boldsymbol{W}|\boldsymbol{Z})p(\boldsymbol{y}^*|\boldsymbol{x}^*,\boldsymbol{W})d\,\boldsymbol{W}d\,\boldsymbol{y}^* \nonumber\\
  &\approx\frac{1}{T}\sum_{t=1}^T \boldsymbol{\hat{y}}^*(\boldsymbol{x}^*,\boldsymbol{W}^t),\\
  \text{Var}[\boldsymbol{y}^*|\boldsymbol{x}^*]&=\int(\boldsymbol{y}^*)^2\int q_{\boldsymbol{\Phi}}(\boldsymbol{W}|\boldsymbol{Z}) p(\boldsymbol{y}^*|\boldsymbol{x}^*,\boldsymbol{W})d\,\boldsymbol{W}d\,\boldsymbol{y}^*-\text{E}^2[\boldsymbol{y}^*|\boldsymbol{x}^*]\nonumber\\
  &\approx\frac{1}{T}\sum_{t=1}^T\left(\boldsymbol{\hat{y}}^*(\boldsymbol{x}^*,\boldsymbol{W}^t)-\text{E}[\boldsymbol{y}^*|\boldsymbol{x}^*]\right)^2,
\end{align}
\vspace{-4mm}
\end{footnotesize}

\noindent where $T$ is number of times of the stochastic forward passes through the DNN, and the mean $\text{E}[\boldsymbol{y}^*|\boldsymbol{x}^*]$ and the variance $\text{Var}[\boldsymbol{y}^*|\boldsymbol{x}^*]$ are considered as the expected output and the model uncertainty.

We conducted experiments via a FC neural network with two hidden layers and $800$ hidden nodes of each on the MNIST dataset and compared it with two referred methods,~\emph{i.e.}, MC dropout~\cite{gal16mcdropout} and ApDeepSense~\cite{yao2018apdeepsense}.

\begin{algorithm}[!t]
    \caption{Parameter pruning with the advanced dropout technique}
    \begin{algorithmic}[1]
        \Require Randomly initialized parameter sets $\boldsymbol{\Theta}_0$ and $\boldsymbol{\Lambda}_0$
        \Ensure $\boldsymbol{\Theta}_j$: estimated model parameters in the $j^{th}$ round
        \State Initialize a network with parameter sets $\boldsymbol{\Theta}_0$ and $\boldsymbol{\Lambda}_0$.
        \State Train the network with the advanced dropout technique and obtain parameter sets $\boldsymbol{\Theta}_j$ and $\boldsymbol{\Lambda}_j$.
        \State Prune $q\%$ parameters in $\boldsymbol{\Theta}_j$ which correspond to the $q\%$ largest dropout rates calculated by \eqref{eq:dropoutrate}.
        \State Reset the remaining parameters in $\boldsymbol{\Theta}_j$ to their values in $\boldsymbol{\Theta}_0$ and parameters in $\boldsymbol{\Lambda}_j$ to their values in $\boldsymbol{\Lambda}_0$.
        \State Return to the step $2$.
    \end{algorithmic}
    \label{alg:netpruning}
\end{algorithm}

The performance of each technique is assessed by area under the receiver operating characteristic curve (AUROC) of two measurements of uncertainty,~\emph{i.e.}, max probability $\mathcal{P}$ and entropy $\mathcal{H}$, which are introduced in the works~\cite{hendrycks2017a,malinin2018predictive}. $\mathcal{P}$ and $\mathcal{H}$ of test output $\boldsymbol{y}^*$ given the corresponding test sample $\boldsymbol{x}^*$ are respectively defined as

\begin{footnotesize}
\vspace{-4mm}
\begin{align}\label{eq:maxprob}
  \mathcal{P}&=\max_{c=1,\cdots,C}p(y_c^*|\boldsymbol{x}^*,\boldsymbol{W}),\\
  \mathcal{H}&=-\sum_{c=1}^C p(y_c^*|\boldsymbol{x}^*,\boldsymbol{W})\ln(p(y_c^*|\boldsymbol{x}^*,\boldsymbol{W})),
\end{align}
\vspace{-4mm}
\end{footnotesize}

\noindent where $C$ is the number of classes. From Table~\ref{tab:uncertainty}, it is observed that the advanced dropout technique surpasses both referred techniques by more than $0.03$ and $0.04$ on the AUC of max probability and entropy, respectively. Meanwhile, the accuracies of these techniques are shown in Table~\ref{tab:uncertainty}, while the advanced dropout outperforms both referred techniques with a significant improvement around $4\%$. Furthermore, Figure~\ref{fig:confusion} shows the confusion matrices of these techniques, one subfigure each. The advanced dropout outperforms the other two techniques in eight classes (class $2$-$9$) with the largest increase by about $15\%$ and the smallest one by about $2\%$, while it keeps the same level as the others in the other two classes.

Figure~\ref{fig:uncertainty} shows four examples of visualization of the uncertainty inference results predicted by all three techniques. In Figure~\ref{fig:uncertainty}(a), all the three techniques are confused between $4$ and $6$, even though the ground truth is $6$. While the two referred techniques predict the wrong class, the advanced dropout technique makes a correct prediction with the highest mean and the smallest variance. This indicates that it is more confident in the prediction. Similarly in Figure~\ref{fig:uncertainty}(b), the advanced dropout technique predicts the correct class, while the referred techniques are all wrong. Meanwhile, in Figure~\ref{fig:uncertainty}(c), all the three techniques predict to the wrong class $0$, while the true label is $5$. However, the advanced dropout technique hesitates between the classes $0$ and $5$ with similar means and larger variances. On the other hand, the referred techniques have their largest prediction values in class $0$ with smaller variances than the advanced dropout technique.

\begin{table*}[!t]
  \centering
  \caption{\footnotesize {Test accuracies (\%) and $p$-value of Student's~\emph{t}-tests on the Reuters-$21578$ dataset. Note that the best results are marked in~\textbf{bold} and the second best results are marked by~\underline{underline}, respectively. The significance level was set as $0.05$.}}
  \vspace{-3mm}
  \resizebox{0.5\linewidth}{!}{
    \begin{tabular}{lccc}
    \toprule
    \multicolumn{1}{c}{Method} & $10$-$1\times800$-$10$/$p$-value & $10$-$2\times800$-$10$/$p$-value & $10$-$4\times800$-$10$/$p$-value \\
    \midrule
    No dropout & $87.68\pm0.18$/$8.12\times10^{-7}$ & $87.58\pm0.24$/$8.12\times10^{-7}$ & $87.63\pm0.07$/$8.12\times10^{-7}$ \\
    Dropout, Bernoulli & $87.75\pm0.21$/$1.45\times10^{-6}$ & $87.09\pm0.72$/$1.45\times10^{-6}$ & $87.50\pm0.31$/$1.45\times10^{-6}$ \\
    Dropout, Gaussian & $87.51\pm0.38$/$6.79\times10^{-6}$ & $87.45\pm0.28$/$6.79\times10^{-6}$ & $87.95\pm0.20$/$6.79\times10^{-6}$ \\
    Dropout, uniform & $88.08\pm0.19$/$4.94\times10^{-6}$ & $87.83\pm0.12$/$4.94\times10^{-6}$ & $88.22\pm0.30$/$4.94\times10^{-6}$ \\
    Concrete dropout & $88.73\pm0.18$/$2.45\times10^{-4}$ & $\underline{88.82\pm0.30}$/$2.45\times10^{-4}$ & $88.73\pm0.19$/$2.45\times10^{-4}$ \\
    Variational dropout & $\underline{88.99\pm0.25}$/$4.32\times10^{-3}$ & $\underline{88.82\pm0.32}$/$4.32\times10^{-3}$ & $\underline{88.83\pm0.13}$/$4.32\times10^{-3}$ \\
    $\beta$-dropout & $88.50\pm0.25$/$1.14\times10^{-4}$ & $88.58\pm0.31$/$1.14\times10^{-4}$ & $88.67\pm0.18$/$1.14\times10^{-4}$ \\
    Continuous dropout & $88.11\pm0.27$/$1.96\times10^{-5}$ & $88.13\pm0.18$/$1.96\times10^{-5}$ & $88.13\pm0.22$/$1.96\times10^{-5}$ \\
    Information dropout & $88.77\pm0.17$/$2.63\times10^{-4}$ & $88.31\pm0.28$/$2.63\times10^{-4}$ & $88.73\pm0.15$/$2.63\times10^{-4}$ \\
    Gaussian soft dropout & $88.67\pm0.26$/$4.36\times10^{-4}$ & $88.63\pm0.21$/$4.36\times10^{-4}$ & $88.58\pm0.21$/$4.36\times10^{-4}$ \\
    Laplace soft dropout & $88.69\pm0.14$/$1.05\times10^{-4}$ & $88.50\pm0.21$/$1.05\times10^{-4}$ & $88.55\pm0.08$/$1.05\times10^{-4}$ \\
    Advanced dropout & $\boldsymbol{89.62\pm0.26}$/ N/A & $\boldsymbol{89.31\pm0.11}$/ N/A & $\boldsymbol{89.35\pm0.08}$/ N/A \\
    \bottomrule
    \end{tabular}}
  \label{tab:reuters}
  \vspace{-2mm}
\end{table*}

\begin{table*}[!t]
  \centering
  \caption{\footnotesize Test RMSE and $p$-value of Student's~\emph{t}-tests on the four UCI datasets. Note that the best results are marked in~\textbf{bold} and the second best results are marked by~\underline{underline}, respectively. The significance level was set as $0.05$.}\vspace{-3mm}
  \scriptsize
  \resizebox{0.8\linewidth}{!}{
  \begin{tabular}{lcccc}
    \toprule
    \multicolumn{1}{c}{ Dataset} &  Boston Housing &  Concrete Strength &  Wine Quality Red &  Yacht Hydrodynamics \\
    \multicolumn{1}{c}{ Base model} &  $13$-$2\times50$-$1$ &  $8$-$2\times50$-$1$ &  $11$-$2\times50$-$1$ &  $6$-$2\times50$-$1$ \\
    \multicolumn{1}{c}{ Method} &  RMSE/$p$-value &  RMSE/$p$-value &  RMSE/$p$-value &  RMSE/$p$-value \\
    \midrule
     No dropout &  $8.5535\pm0.0129$/$5.67\times10^{-5}$ &  $15.8219\pm0.0275$/$5.92\times10^{-4}$ &  $0.8154\pm0.0014$/$3.51\times10^{-3}$ &  $13.1689\pm0.1765$/$2.05\times10^{-4}$ \\
     Dropout, Bernoulli &  $8.5065\pm0.0071$/$6.88\times10^{-5}$ &  $15.7828\pm0.0131$/$1.69\times10^{-4}$ &  $0.8130\pm0.0005$/$3.86\times10^{-3}$ &  $13.0169\pm0.0722$/$1.11\times10^{-5}$ \\
     Dropout, Gaussian &  $8.5062\pm0.0092$/$2.14\times10^{-4}$ &  $15.7775\pm0.0123$/$1.79\times10^{-4}$ &  $\underline{0.8128\pm0.0003}$/$1.59\times10^{-3}$ &  $12.9740\pm0.1621$/$3.57\times10^{-4}$ \\
     Dropout, uniform &  $8.5446\pm0.0143$/$1.25\times10^{-4}$ &  $15.8134\pm0.0244$/$5.00\times10^{-4}$ &  $0.8135\pm0.0004$/$9.25\times10^{-4}$ &  $13.0386\pm0.0791$/$1.46\times10^{-5}$ \\
     Concrete dropout &  $8.7723\pm0.0909$/$1.14\times10^{-3}$ &  $16.0284\pm0.0283$/$1.14\times10^{-5}$ &  $0.8273\pm0.0049$/$1.69\times10^{-3}$ &  $13.2463\pm0.3342$/$1.74\times10^{-3}$ \\
     Variational dropout &  $8.5078\pm0.0090$/$1.68\times10^{-4}$ &  $15.7920\pm0.0162$/$2.46\times10^{-4}$ &  $0.8133\pm0.0004$/$1.11\times10^{-3}$ &  $13.0626\pm0.0592$/$3.79\times10^{-6}$ \\
     Gaussian soft dropout &  $8.4894\pm0.0119$/$2.93\times10^{-3}$ &  $15.7604\pm0.0139$/$8.43\times10^{-4}$ &  $\underline{0.8128\pm0.0003}$/$2.57\times10^{-3}$ &  $12.3218\pm0.0419$/$3.20\times10^{-3}$ \\
     Laplace soft dropout &  $\underline{8.4728\pm0.0074}$/$7.36\times10^{-3}$ &  $\underline{15.7488\pm0.0214}$/$9.78\times10^{-3}$ &  $0.8133\pm0.0005$/$2.46\times10^{-3}$ &  $\underline{12.2844\pm0.0246}$/$1.88\times10^{-3}$ \\
     Advanced dropout &  $\boldsymbol{8.4577\pm0.0009}$/ N/A &  $\boldsymbol{15.7085\pm0.0003}$/ N/A &  $\boldsymbol{0.8119\pm0.0001}$/ N/A &  $\boldsymbol{12.2142\pm0.0057}$/ N/A \\
    \bottomrule
  \end{tabular}}
  \label{tab:uci}
  \vspace{-6mm}
\end{table*}

In summary, the advanced dropout technique is suitable for model uncertainty inference and performs better than the referred methods. It can effectively infer the uncertainty and improve the classification accuracies simultaneously.

\vspace{-3mm}
\subsection{Network Pruning}\label{ssec:pruing}

{In this section, we extend the application of the advanced dropout technique to the field of network pruning, which is a popular and important topic in deep learning. Network pruning strategies can be divided into two categories,~\emph{i.e.}, node pruning and parameter pruning. The algorithm for parameter pruning with the advanced dropout technique is summarized in Algorithm~\ref{alg:netpruning}. The node pruning was conducted in a similar way, which replaces pruning $q\%$ parameters by pruning $q\%$ nodes.}

{We conducted experiments on the MNIST dataset in an FC neural network with two hidden layers and $800$ hidden nodes of each. We compared the advanced dropout technique with a random selection method as a baseline, a state-of-the-art technique (winning ticket~\cite{frankle2019the}), and two dropout variants (the Gaussian soft dropout and the Laplace soft dropout~\cite{xie2019soft}). The random selection method stochastically chooses nodes or parameters in each round. We adopted the SGD algorithm for model training and set $q$ as $10$.}

{Figure~\ref{fig:pruning}(a) shows the performance of node pruning on the MNIST dataset. The advanced dropout technique achieves higher accuracy than the referred methods on each percentage of node preservation. In addition, Figure~\ref{fig:pruning}(b) shows the performance of parameter pruning on the MNIST dataset, which can also demonstrate the better performance of the advanced dropout. Therefore, the advanced dropout technique can be effectively utilized in network pruning and is better than the referred techniques.}

\vspace{-3mm}
\subsection{{Text Classification}}\label{ssec:textcls}

{In this section, in addition to the aforementioned computer vision tasks, we further evaluated the performance of the proposed advanced dropout technique in text classification on Reuter-$21578$ dataset\footnote{{\url{http://kdd.ics.uci.edu/databases/reuters21578/}}}. According to the official data preprocessing procedure, each document in the dataset was expressed by a $10$-dimensional vector. The dropout techniques were implemented by using the FC neural networks with one, two, and four hidden layer(s) and $800$ hidden nodes each. All the models were trained for $50$ epochs using the SGD algorithm with the learning rate as $0.01$ and the weight decay as $5\times10^{-4}$. We evaluated each model for $5$ runs with random initializations.}

{Experimental results and the $p$-value of the Student's~\emph{t}-tests between the advanced dropout and other referred dropout variants are shown in Table~\ref{tab:reuters}. The proposed advanced dropout achieves the best accuracies in the models with different depths. As the $p$-value are all smaller than the significance level (\emph{i.e.}, $\alpha=0.05$), the advanced dropout obtains statistically significant improvement.}

\vspace{-3mm}
\subsection{Regression}\label{ssec:regression}

{In this section, except for the aforementioned classification tasks, we evaluated the proposed advanced dropout for regression tasks on the UCI datasets~\cite{dua17uci}. The dropout techniques were implemented by using the FC neural networks with $2$ hidden layers and $50$ hidden nodes each. All the models were trained for $50$ epochs using the SGD algorithm with the learning rate as $0.01$ and the weight decay as $5\times10^{-4}$. We evaluated each model for $5$ runs with random initializations. The root mean squared error (RMSE) was chosen as the evaluation metric.}

{Experimental results and the $p$-value of the Student's~\emph{t}-tests between the advanced dropout and other referred dropout variants are listed in Table~\ref{tab:uci}. On each dataset, the proposed advanced dropout achieves the smallest RMSEs. As the $p$-value are all smaller than the significance level (\emph{i.e.}, $\alpha=0.05$), statistically significant improvement was obtained by the advanced dropout.}

\vspace{-3mm}
\section{Conclusions}\label{sec:conclusions}

In this paper, we proposed the advanced dropout technique, a model-free methodology, to improve the ability of overfitting prevention and the classification performance of DNNs. The advanced dropout uses a model-free and easily implemented distribution with a parametric prior to adaptively adjust the dropout rate. Furthermore, the prior parameters are optimized by the SGVB inference to perform an end-to-end training procedure of DNNs. In the experiments, we evaluated the effectiveness of the advanced dropout technique with nine referred techniques in different base models on {seven} widely used datasets (including five small-scale datasets and {two} large-scale datasets). The advanced dropout statistically significantly outperformed all the referred techniques. We compared training time and effectiveness ratios between the techniques and found that the advanced dropout achieves highest effectiveness ratios on most datasets. Ablation studies were conducted to analyze the effectiveness of each component. We further compared training time and effectiveness ratios between the techniques and found that the advanced dropout achieves highest effectiveness ratios on most datasets. Next, we conducted a series of analysis of dropout rate characteristics, including convergence of the adaptive dropout rate, the learned distributions of dropout masks, and a comparison with dropout rate generation without using an explicit distribution. In addition, the ability of overfitting prevention of the advanced dropout was evaluated and confirmed. Finally, we extended the application of the advanced dropout to uncertainty inference, network pruning, {text classification}, and regression, and we found that the advanced dropout is superior to the corresponding referred methods.

\vspace{-4mm}
\section*{Acknowledgments}
\vspace{-1mm}

This work was supported in part by the National Key R\&D Program of China under Grant $2019$YFF$0303300$ and under Subject II No. $2019$YFF$0303302$, in part by National Natural Science Foundation of China (NSFC) No. $61922015$, $61773071$, U$19$B$2036$, in part by Beijing Natural Science Foundation Project No. Z$200002$, in part by the Beijing Academy of Artificial Intelligence (BAAI) under Grant BAAI$2020$ZJ$0204$, and in part by the Beijing Nova Programme Interdisciplinary Cooperation Project under Grant Z$191100001119140$.

\vspace{-4mm}
\bibliographystyle{IEEEbib}
\bibliography{bare_jrnl_compsoc}

\begin{thebibliography}{10}

\bibitem{simonyan2015very}
K.~Simonyan and A.~Zisserman,
\newblock ``Very deep convolutional networks for large-scale image
  recognition,''
\newblock in {\em International Conference on Learning Representations}, 2015.

\bibitem{he2016deep}
K.~He, X.~Zhang, S.~Ren, and J.~Sun,
\newblock ``Deep residual learning for image recognition,''
\newblock in {\em Computer Vision and Pattern Recognition}, 2016, pp. 770--778.

\bibitem{huang2017densely}
G.~Huang, Z.~Liu, L.~V. Der~Maaten, and K.~Q. Weinberger,
\newblock ``Densely connected convolutional networks,''
\newblock in {\em Computer Vision and Pattern Recognition}, 2017, pp.
  2261--2269.

\bibitem{li2019dual}
X.~Li, L.~Yu, D.~Chang, Z.~Ma, and J.~Cao,
\newblock ``Dual cross-entropy loss for small-sample fine-grained vehicle
  classification,''
\newblock {\em IEEE Transactions on Vehicular Technology}, vol. 68, no. 5, pp.
  4204--4212, 2019.

\bibitem{li2019large}
X.~Li, D.~Chang, T.~Tian, and J.~Cao,
\newblock ``Large-margin regularized softmax cross-entropy loss,''
\newblock {\em IEEE Access}, vol. 7, pp. 19572--19578, 2019.

\bibitem{xu2016instance}
P.~Xu, Q.~Yin, Y.~Qi, Y.-Z. Song, Z.~Ma, L.~Wang, and J.~Guo,
\newblock ``Instance-level coupled subspace learning for fine-grained
  sketch-based image retrieval,''
\newblock in {\em European Conference on Computer Vision Workshops}. Springer,
  2016, pp. 19--34.

\bibitem{xu2018sketchmate}
P.~Xu, Y.~Huang, T.~Yuan, K.~Pang, Y.~Z. Song, T.~Xiang, T.~M. Hospedales,
  Z.~Ma, and J.~Guo,
\newblock ``{SketchMate}: Deep hashing for million-scale human sketch
  retrieval,''
\newblock in {\em Computer Vision and Pattern Recognition}, 2018, pp.
  8090--8098.

\bibitem{xu2018cross}
P.~Xu, Q.~Yin, Y.~Huang, Y.~Z. Song, Z.~Ma, L.~Wang, T.~Xiang, W.~B. Kleijn,
  and J.~Guo,
\newblock ``Cross-modal subspace learning for fine-grained sketch-based image
  retrieval,''
\newblock {\em Neurocomputing}, vol. 278, pp. 75--86, 2018.

\bibitem{ma2019shoe}
Z.~Ma, Y.~Ding, S.~Wen, J.~Xie, Y.~Jin, Z.~Si, and H.~Wang,
\newblock ``Shoe-print image retrieval with multi-part weighted {CNN},''
\newblock {\em IEEE ACCESS}, vol. 7, pp. 59728 -- 59736, 2019.

\bibitem{bai2021unsupervised}
C.~Bai, H.~Li, J.~Zhang, L.~Huang, and L.~Zhang,
\newblock ``Unsupervised adversarial instance-level image retrieval,''
\newblock {\em IEEE Transactions on Multimedia}, 2021.

\bibitem{xu2018webly-supervised}
Z.~{Xu}, S.~{Huang}, Y.~{Zhang}, and D.~{Tao},
\newblock ``Webly-supervised fine-grained visual categorization via deep domain
  adaptation,''
\newblock {\em IEEE Transactions on Pattern Analysis and Machine Intelligence},
  vol. 40, no. 5, pp. 1100--1113, 2018.

\bibitem{adeli2019semi-supervised}
E.~{Adeli}, K.~{Thung}, L.~{An}, G.~{Wu}, F.~{Shi}, T.~{Wang}, and D.~{Shen},
\newblock ``Semi-supervised discriminative classification robust to
  sample-outliers and feature-noises,''
\newblock {\em IEEE Transactions on Pattern Analysis and Machine Intelligence},
  vol. 41, no. 2, pp. 515--522, 2019.

\bibitem{ma2018group}
K.~{Ma}, Z.~{Duanmu}, Z.~{Wang}, Q.~{Wu}, W.~{Liu}, H.~{Yong}, H.~{Li}, and
  L.~{Zhang},
\newblock ``Group maximum differentiation competition: Model comparison with
  few samples,''
\newblock {\em IEEE Transactions on Pattern Analysis and Machine Intelligence},
  2018.

\bibitem{zhu2019image-text}
F.~Zhu, Z.~Ma, X.~Li, G.~Chen, J.-T. Chien, J.-H. Xue, and J.~Guo,
\newblock ``Image-text dual neural network with decision strategy for
  small-sample image classification,''
\newblock {\em Neurocomputing}, vol. 328, pp. 182--188, 2019.

\bibitem{hinton12}
G.~E. Hinton, N.~Srivastava, A.~Krizhevsky, I.~Sutskever, and R.~Salakhutdinov,
\newblock ``Improving neural networks by preventing co-adaptation of feature
  detectors,''
\newblock {\em arXiv}, 2012.

\bibitem{xie2019soft}
J.~Xie, Z.~Ma, G.~Zhang, J.-H. Xue, Z.-H. Tan, and J.~Guo,
\newblock ``Soft dropout and its variational {B}ayes approximation,''
\newblock in {\em IEEE International Workshop on Machine Learning for Signal
  Processing}, 2019.

\bibitem{achille2018information}
A.~{Achille} and S.~{Soatto},
\newblock ``Information dropout: Learning optimal representations through noisy
  computation,''
\newblock {\em IEEE Transactions on Pattern Analysis and Machine Intelligence},
  vol. 40, no. 12, pp. 2897--2905, 2018.

\bibitem{labach2019survey}
A.~Labach and H.~Salehinejad,
\newblock ``Survey of dropout methods for deep neural networks,''
\newblock {\em arXiv}, pp. 1--11, 2019.

\bibitem{shen2018continuous}
X.~{Shen}, X.~{Tian}, T.~{Liu}, F.~{Xu}, and D.~{Tao},
\newblock ``Continuous dropout,''
\newblock {\em IEEE Transactions on Neural Networks and Learning Systems}, vol.
  29, no. 9, pp. 3926--3937, 2018.

\bibitem{kingma2015variational}
D.~P. Kingma, T.~Salimans, and M.~Welling,
\newblock ``Variational dropout and the local reparameterization trick,''
\newblock in {\em Neural Information Processing Systems}, 2015, vol.~28, pp.
  2575--2583.

\bibitem{wan2013regularization}
L.~Wan, M.~D. Zeiler, S.~Zhang, Y.~L. Cun, and R.~Fergus,
\newblock ``Regularization of neural networks using {DropConnect},''
\newblock in {\em International Conference on Machine Learning}, 2013, pp.
  1058--1066.

\bibitem{wang2019jumpout}
S.~Wang, T.~Zhou, and J.~A. Bilmes,
\newblock ``Jumpout: Improved dropout for deep neural networks with {ReLUs},''
\newblock in {\em International Conference on Machine Learning}, 2019, pp.
  6668--6676.

\bibitem{ba2013adaptive}
J.~Ba and B.~J. Frey,
\newblock ``Adaptive dropout for training deep neural networks,''
\newblock in {\em Neural Information Processing Systems}, 2013, pp. 3084--3092.

\bibitem{maeda2015a}
S.~Maeda,
\newblock ``A bayesian encourages dropout,''
\newblock in {\em International Conference on Learning Representation}, 2015.

\bibitem{gal16mcdropout}
Y.~Gal and Z.~Ghahramani,
\newblock ``Dropout as a {B}ayesian approximation: representing model
  uncertainty in deep learning,''
\newblock in {\em International Conference on Machine Learning}, 2016, pp.
  1050--1059.

\bibitem{gal2016a}
Y.~Gal and Z.~Ghahramani,
\newblock ``A theoretically grounded application of dropout in recurrent neural
  networks,''
\newblock in {\em Neural Information Processing Systems}, 2016, pp. 1027--1035.

\bibitem{khan2019regularization}
S.~H. Khan, M.~Hayat, and F.~Porikli,
\newblock ``Regularization of deep neural networks with spectral dropout,''
\newblock {\em Neural Networks}, vol. 110, pp. 82--90, 2019.

\bibitem{ko2017controlled}
B.~Ko, H.~Kim, K.~Oh, and H.~Choi,
\newblock ``Controlled dropout: A different approach to using dropout on deep
  neural network,''
\newblock in {\em International Conference on Big Data and Smart Computing},
  2017, pp. 358--362.

\bibitem{li2017dropout}
Y.~Li and Y.~Gal,
\newblock ``Dropout inference in bayesian neural networks with
  alpha-divergences,''
\newblock in {\em International Conference on Machine Learning}, 2017, pp.
  2052--2061.

\bibitem{salehinejad2019ising}
H.~{Salehinejad} and S.~{Valaee},
\newblock ``Ising-dropout: A regularization method for training and compression
  of deep neural networks,''
\newblock in {\em ICASSP 2019 - 2019 IEEE International Conference on
  Acoustics, Speech and Signal Processing (ICASSP)}, 2019, pp. 3602--3606.

\bibitem{chen2019mutual}
J.~Chen, Z.~Wu, J.~Zhang, and F.~Li,
\newblock ``Mutual information-based dropout: Learning deep relevant feature
  representation architectures,''
\newblock {\em Neurocomputing}, vol. 361, pp. 173--184, 2019.

\bibitem{wang2013fast}
S.~I. Wang and C.~D. Manning,
\newblock ``Fast dropout training,''
\newblock in {\em International Conference on Machine Learning}, 2013, pp.
  118--126.

\bibitem{Srivastava14}
N.~Srivastava, G.~E. Hinton, A.~Krizhevsky, I.~Sutskever, and R.~Salakhutdinov,
\newblock ``Dropout: a simple way to prevent neural networks from
  overfitting,''
\newblock {\em Journal of Machine Learning Research}, vol. 15, no. 1, pp.
  1929--1958, 2014.

\bibitem{gal17concrete}
Y.~Gal, J.~Hron, and A.~Kendall,
\newblock ``Concrete dropout,''
\newblock in {\em Neural Information Processing Systems}, 2017, pp. 3581--3590.

\bibitem{liu2019beta}
L.~{Liu}, Y.~{Luo}, X.~{Shen}, M.~{Sun}, and B.~{Li},
\newblock ``$\beta$-dropout: A unified dropout,''
\newblock {\em IEEE Access}, vol. 7, pp. 36140--36153, 2019.

\bibitem{bulo2016dropout}
S.~R. Bulò, L.~Porzi, and P.~Kontschieder,
\newblock ``Dropout distillation,''
\newblock in {\em International Conference on Machine Learning}, 2016.

\bibitem{ma2017dropout}
X.~Ma, Y.~Gao, Z.~Hu, Y.~Yu, Y.~Deng, and E.~Hovy,
\newblock ``Dropout with expectation-linear regularization,''
\newblock in {\em International Conference on Learning Representations}, 2017.

\bibitem{gao2019demystifying}
H.~Gao, J.~Pei, and H.~Huang,
\newblock ``Demystifying dropout,''
\newblock in {\em International Conference on Machine Learning}, 2019.

\bibitem{li2016improved}
Z.~Li, B.~Gong, and T.~Yang,
\newblock ``Improved dropout for shallow and deep learning,''
\newblock in {\em Advances in Neural Information Processing Systems}, 2016.

\bibitem{wang2019rademacher}
H.~Wang, W.~Yang, Z.~Zhao, T.~Luo, J.~Wang, and Y.~Tang,
\newblock ``Rademacher dropout: {A}n adaptive dropout for deep neural network
  via optimizing generalization gap,''
\newblock {\em Neurocomputing}, vol. 357, pp. 177--187, 2019.

\bibitem{liu2019variational}
Y.~Liu, W.~Dong, L.~Zhang, D.~Gong, and Q.~Shi,
\newblock ``Variational {B}ayesian dropout with a hierarchical prior,''
\newblock in {\em Computer Vision and Pattern Recognition}, 2019, pp.
  7124--7133.

\bibitem{kingma2014auto-encoding}
D.~P. Kingma and M.~Welling,
\newblock ``Auto-encoding variational {B}ayes,''
\newblock in {\em International Conference on Learning Representations}, 2014.

\bibitem{bishop06}
C.~M. Bishop,
\newblock {\em Pattern Recognition and Machine Learning},
\newblock Springer Science+Business Media LLC., 2006.

\bibitem{mnist98}
Y.~LeCun and C.~Cortes,
\newblock ``The {MNIST} database of handwritten digits,''
  http://yann.lecun.com/exdb/mnist/, 1998.

\bibitem{krizhevsky09cifar}
A.~Krizhevsky,
\newblock ``Learning multiple layers of features from tiny images,''
\newblock techreport, CIFAR, 2009.

\bibitem{vinyals2016matching}
O.~Vinyals, C.~Blundell, T.~P. Lillicrap, K.~Kavukcuoglu, and D.~Wierstra,
\newblock ``Matching networks for one shot learning,''
\newblock {\em arXiv}, 2016.

\bibitem{griffin2006caltech}
G.~Griffin, A.~Holub, and P.~Perona,
\newblock ``Caltech256 image dataset,''
  \url{http://www.vision.caltech.edu/Image_Datasets/Caltech256/}, 2006.

\bibitem{denoord2016pixel}
A.~V. Den~Oord, N.~Kalchbrenner, and K.~Kavukcuoglu,
\newblock ``Pixel recurrent neural networks,''
\newblock {\em arXiv}, 2016.

\bibitem{russakovsky2015imagenet}
O.~Russakovsky, J.~Deng, H.~Su, J.~Krause, S.~Satheesh, S.~Ma, Z.~Huang,
  A.~Karpathy, A.~Khosla, M.~S. Bernstein, A.~C. Berg, and L.~Feifei,
\newblock ``{ImageNet} large scale visual recognition challenge,''
\newblock {\em International Journal of Computer Vision}, vol. 115, no. 3, pp.
  211--252, 2015.

\bibitem{li2016efficient}
S.~Li, S.~Y. Siu, T.~Fang, and L.~Quan,
\newblock ``Efficient multi-view surface refinement with adaptive resolution
  control,''
\newblock in {\em European Conference on Computer Vision}, 2016, pp. 349--364.

\bibitem{howard2017mobilenets}
A.~G. Howard, M.~Zhu, B.~Chen, D.~Kalenichenko, W.~Wang, T.~Weyand,
  M.~Andreetto, and H.~Adam,
\newblock ``{MobileNets}: {E}fficient convolutional neural networks for mobile
  vision applications,''
\newblock {\em arXiv}, 2017.

\bibitem{zagoruyko2016wide}
S.~Zagoruyko and N.~Komodakis,
\newblock ``Wide residual networks,''
\newblock in {\em British Machine Vision Conference}, 2016.

\bibitem{ma2011bayesian}
Z.~Ma and A.~Leijon,
\newblock ``{B}ayesian estimation of {B}eta mixture models with variational
  inference,''
\newblock {\em IEEE Transactions on Pattern Analysis and Machine Intelligence},
  vol. 33, no. 11, pp. 2160--2173, 2011.

\bibitem{maddison2017the}
C.~J. Maddison, A.~Mnih, and Y.~W. Teh,
\newblock ``The concrete distribution: {A} continuous relaxation of discrete
  random variables,''
\newblock in {\em International Conference on Learning Representations}, 2017.

\bibitem{yao2018apdeepsense}
S.~{Yao}, Y.~{Zhao}, H.~{Shao}, C.~{Zhang}, A.~{Zhang}, D.~{Liu}, S.~{Liu},
  L.~{Su}, and T.~{Abdelzaher},
\newblock ``{ApDeepSense}: Deep learning uncertainty estimation without the
  pain for {IoT} applications,''
\newblock in {\em 2018 IEEE 38th International Conference on Distributed
  Computing Systems (ICDCS)}, 2018, pp. 334--343.

\bibitem{frankle2019the}
J.~Frankle and M.~Carbin,
\newblock ``The lottery ticket hypothesis: Finding sparse, trainable neural
  networks,''
\newblock in {\em International Conference on Learning Representations}, 2019.

\bibitem{hendrycks2017a}
D.~Hendrycks and K.~Gimpel,
\newblock ``A baseline for detecting misclassified and out-of-distribution
  examples in neural networks,''
\newblock in {\em International Conference on Learning Representations}, 2017.

\bibitem{malinin2018predictive}
A.~Malinin and M.~Gales,
\newblock ``Predictive uncertainty estimation via prior networks,''
\newblock in {\em Neural Information Processing Systems}, 2018, pp. 7047--7058.

\bibitem{dua17uci}
D.~Dua and C.~Graff,
\newblock ``{UCI} machine learning repository,'' http://archive.ics.uci.edu/ml,
  2017.

\end{thebibliography}

\end{document}